\documentclass{article} 
\usepackage[T1]{fontenc}
\usepackage{iclr2027_conference,times}

\usepackage{amsmath,amsfonts,bm}

\def\eqref#1{equation~\ref{#1}}

\def\1{\bm{1}}

\DeclareMathAlphabet{\mathsfit}{\encodingdefault}{\sfdefault}{m}{sl}
\SetMathAlphabet{\mathsfit}{bold}{\encodingdefault}{\sfdefault}{bx}{n}

\usepackage{hyperref}
\usepackage{url}
\usepackage{booktabs}
\usepackage{enumitem}
\usepackage{multirow}
\usepackage{graphicx}
\usepackage[table]{xcolor}
\usepackage{subfigure}
\usepackage[most]{tcolorbox}
\usepackage{bbm}
\usepackage{colortbl,xcolor} 
\usepackage{wrapfig}
\usepackage{lipsum}
\usepackage{pifont}
\usepackage{dashrule}
\usepackage{amsmath,amsfonts,amssymb}
\usepackage{footmisc}
\usepackage{float}
\usepackage{soul}
\usepackage{lscape}

\usepackage{CJKutf8}
\usepackage{booktabs}
\usepackage{tabularx}
\usepackage{array}

\usepackage{twemojis}
\usepackage{inconsolata}   
\tcbuselibrary{listings}
\newcommand{\sig}[1]{\textsuperscript{#1}}   

\definecolor{citeblue}{HTML}{1F5AA6}

\hypersetup{
    colorlinks=true,
    citecolor=citeblue,
    linkcolor=black,
    urlcolor=blue
}

\usepackage[utf8]{inputenc} 
\usepackage{CJKutf8}

\usepackage[most]{tcolorbox}
\usepackage{xcolor}
\definecolor{hypblue}{HTML}{2F6F9F}       
\definecolor{hypdarkblue}{HTML}{24577D}   
\definecolor{hyplightblue}{HTML}{F1F7FC}  

\newtcolorbox{hypothesisbox}[1]{
  enhanced,
  breakable,
  colback=hyplightblue,
  colframe=hyplightblue,
  boxrule=0pt,
  borderline west={2.2pt}{0pt}{hypblue},
  arc=0pt,
  outer arc=0pt,
  boxsep=0pt,
  left=5mm,
  right=5mm,
  top=3.5mm,
  bottom=3.5mm,
  before skip=10pt,
  after skip=10pt,
  fontupper=\normalsize,
  coltitle=hypblue,
  attach title to upper,
  after title={\par\smallskip},
}
\usepackage{siunitx}
\usepackage[%
    framemethod=tikz,
    skipbelow=\topskip,
    skipabove=\topskip
]{mdframed}
\mdfsetup{%
    leftmargin=0pt,
    rightmargin=0pt,
    backgroundcolor=bggray,
    middlelinecolor=black,
    roundcorner=3
}
\newtcolorbox[list inside=prompt,auto counter,number within=section]{prompt}[1][]{
    colbacktitle=black!60,
    fonttitle=\small,
    coltitle=white,
    fontupper=\footnotesize,
    boxsep=4pt,
    left=0pt,
    top=0pt,
    bottom=0pt,
    boxrule=1pt,
    #1,
}

\usepackage[most]{tcolorbox}

\newtcolorbox{definitionbox}{
  enhanced,
  breakable,
  colback=yellow!12!white,
  colframe=yellow!12!white,
  boxrule=0pt,
  borderline west={2.2pt}{0pt}{yellow!55!orange},
  arc=0pt,
  outer arc=0pt,
  boxsep=0pt,
  left=7pt,
  right=4pt,
  top=4pt,
  bottom=4pt,
  before skip=8pt,
  after skip=8pt,
  width=\linewidth,
  box align=base,
}

\definecolor{first}{HTML}{159DEB}
\definecolor{second}{HTML}{E50C1C}
\newcommand{\na}{\multicolumn{1}{c}{--}}

\title{
  \textcolor{first}{\textbf{Where's}}
  \textcolor{second}{\textbf{Waldo?}} \\
  Query-language Preference under \\ 
  Cross-lingual Knowledge Disparities 
}

\author{Dayeon Ki$^{1}$\thanks{Work done during an internship at Microsoft.} \quad
Ruochen Zhang$^{2}$ \quad
Silviu Cucerzan$^{2}$ \quad
Ryen W. White$^{2}$ \quad
Ning Gao$^{2}$ \\
$^{1}$University of Maryland \quad
$^{2}$Microsoft \\
\texttt{dayeonki@umd.edu}
}

\iclrfinalcopy 
\begin{document}

\maketitle

\begin{abstract}
Large Language Models increasingly serve as interfaces for knowledge-intensive information seeking tasks across languages by synthesizing multilingual evidence. 
Prior work has shown that they often exhibit query-language preference\textemdash the tendency to favor sources written in the language of the query\textemdash but has largely examined this behavior in settings where equivalent knowledge is available across languages.
However, this bias becomes consequential when sources in different languages provide incomplete or inconsistent accounts of the same fact, since the information users receive then depends on the sources a model selects to use.
To characterize query-language preference under such cross-lingual knowledge disparities, we introduce \textsc{Waldo},\footnote{\textsc{Waldo} stands for ``\textbf{W}ikipedi\textbf{A} \textbf{L}ost in \textbf{D}isparity of kn\textbf{O}wledge,'' named after the character from the \textcolor{first}{Where's} \textcolor{second}{Waldo?} series (in~\autoref{fig:main_fig}), where readers search crowded scenes to find him; similarly, \textsc{Waldo} tests whether models can use relevant facts across multilingual documents with missing or conflicting knowledge.} a multilingual Question-Answering (QA) benchmark constructed from Wikipedia.
\textsc{Waldo} contains 12K QA pairs targeting knowledge \emph{gaps}, where a fact is available in one language but absent in another, and knowledge \emph{conflicts}, where language editions provide conflicting versions of the same fact.
Evaluating eight models across five languages, we find that when one language edition merely lacks the relevant fact, models generally use evidence from the other language regardless of the query language.
Under conflicting accounts, however, model responses strongly align with the document in the query language, causing semantically equivalent queries to elicit different accounts depending on the user's language.
Finally, we explore two different approaches that could mitigate this preference under knowledge conflicts: a mechanistic intervention that ablates attention heads associated with query-language preference, and LoRA-based training, which reduces the preference gap by up to 61.5\%.\footnote{Code, data, and fine-tuned models will be released upon publication.}

\end{abstract}

\vspace{-1em}
\section{Introduction}
\vspace{-0.1em}

\begin{figure*}
    \centering
    \includegraphics[width=\linewidth]{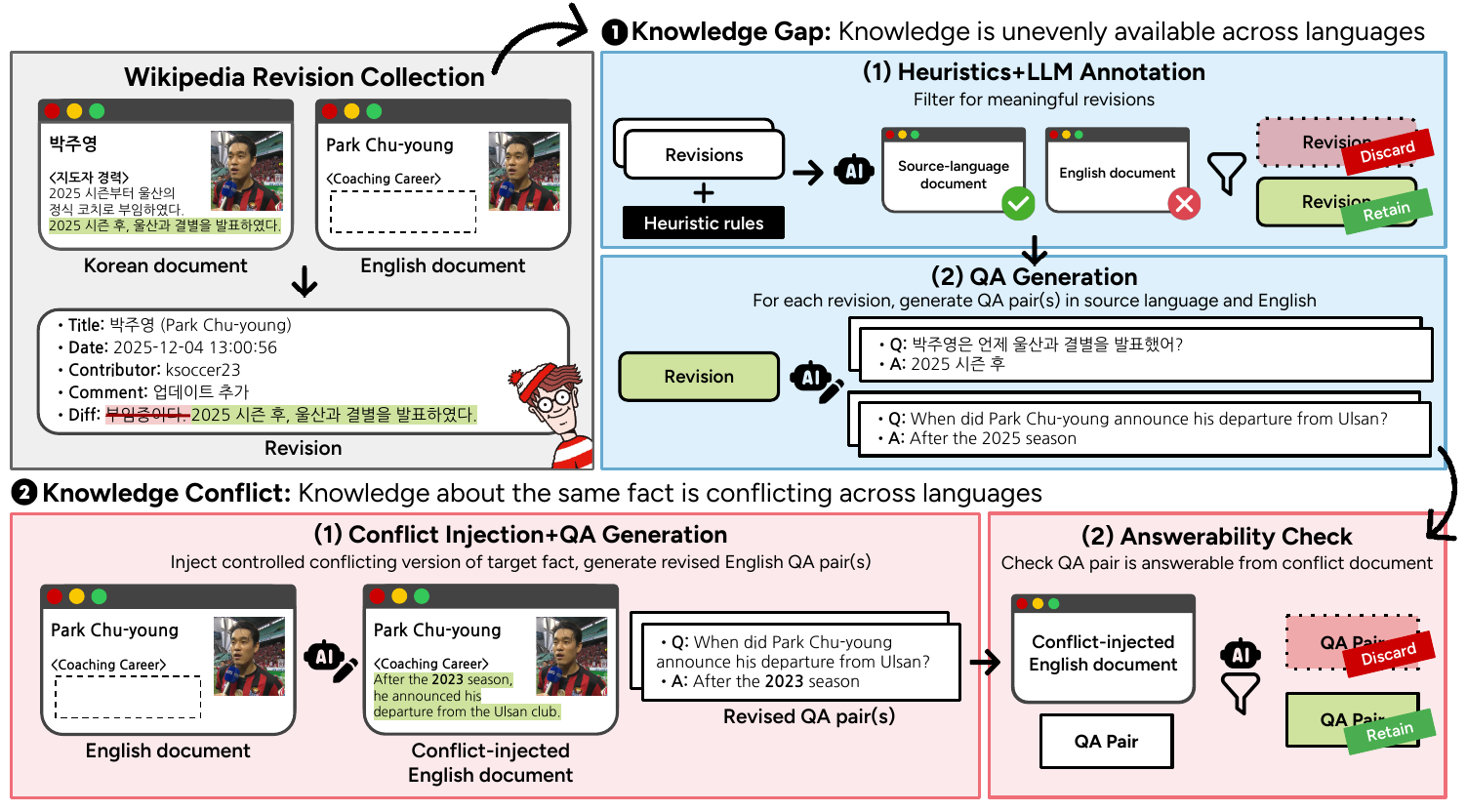}
    \vspace{-1.2em}
    \caption{\textbf{Overview of \textsc{Waldo}.}
    For each source language (e.g., Korean), we collect its Wikipedia revisions and pair them with the corresponding English documents.
    For \raisebox{-0.2em}{\includegraphics[height=1em]{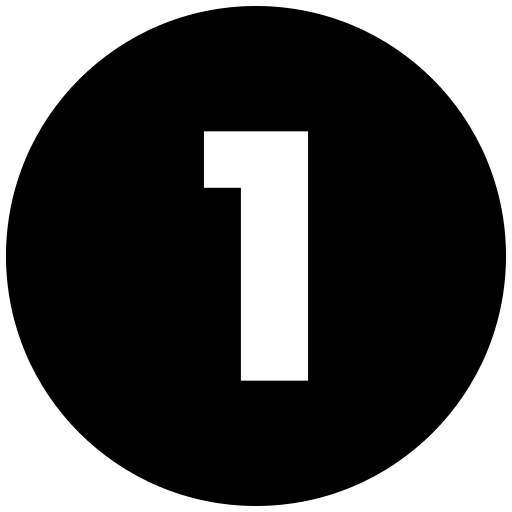}} \textbf{Knowledge Gap}, we identify revisions containing facts present in the source-language document but absent from English, generating aligned source-language and English QA pairs for each.
    For \raisebox{-0.2em}{\includegraphics[height=1em]{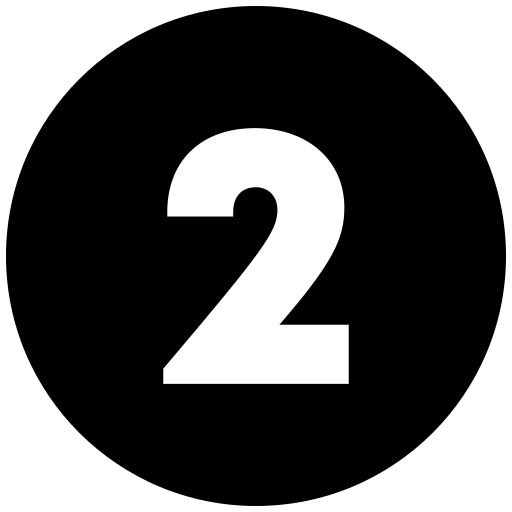}} \textbf{Knowledge Conflict}, we inject a controlled conflicting fact into the English document and generate revised English QA pairs.
    }
    \label{fig:main_fig}
     \vspace{-1em}
\end{figure*}

Large Language Models (LLMs) increasingly mediate knowledge-intensive information seeking tasks across diverse languages \citep{longpre-etal-2021-mkqa}. 
When answering a user's query, they may draw on information from sources in multiple languages and select which sources inform their final responses \citep{wu2024languagesequalinsightsmultilingual, chirkova-etal-2024-retrieval}.
As LLMs become interfaces between users and multilingual information ecosystems \citep{information_2026}, understanding \textit{how} they select among competing knowledge sources becomes increasingly important.

A growing body of evidence suggests that multilingual systems often exhibit query-language preference during generation: a tendency to favor knowledge sources written in the same language as the user query \citep{ki2026linguistic}.
Consequently, the language in which a user poses a question can influence which sources inform the response, and therefore what information the user ultimately receives.
This is not necessarily problematic when equivalent information is available across languages\textemdash an assumption underlying much prior work \citep{park-lee-2025-investigating, qi-etal-2025-consistency}. 
In practice, however, multilingual sources are rarely perfectly redundant or mutually consistent \citep{filatova2009directions}.

Query-language preference becomes consequential when knowledge is \textit{heterogeneous} across languages, which can take two forms.
First, knowledge may be unevenly available: a fact documented in one language may be absent from sources in another, creating a \emph{knowledge gap} \citep{hecht2010tower}.
Second, sources in different languages may provide incompatible accounts of the same fact, resulting in a \emph{knowledge conflict} \citep{li-etal-2024-land, cattan2025draggedconflictsdetectingaddressing}.
When neither account can be resolved using available signals, semantically equivalent queries in different languages should surface the gap or conflict rather than elicit different responses depending on the query language.
At scale, these differences can create asymmetries in answer quality \citep{jin2024better}, produce uneven downstream performance \citep{ranaldi2026multilingual}, deepen opinion polarization \citep{echochamber_2024}, reinforce the marginalization of underrepresented narratives \citep{phillipson1992linguistic}, and ultimately undermine user trust in multilingual information systems \citep{arguedas2023news}.
These risks motivate the study of query-language preference under cross-lingual knowledge disparities.

To this end, we introduce \textbf{\textsc{Waldo}}, a multilingual Question-Answering (QA) benchmark for evaluating query-language preference under knowledge gaps and knowledge conflicts (\S\ref{sec:dataset}).
As illustrated in~\autoref{fig:main_fig}, \textsc{Waldo} constructs QA instances around target facts that satisfy two criteria: they appear in Wikipedia documents in languages other than English but are absent from the corresponding English documents (\S\ref{sec:gap}), and they postdate the reported knowledge cutoffs of the evaluated model. 
This design helps minimize the possibility that models answer from their parametric knowledge, enabling observed response differences to be more directly attributed to their use of the context \citep{jin-etal-2024-tug}.
We additionally introduce factually inconsistent versions of the target facts by manually perturbing the corresponding English Wikipedia documents to simulate real-world knowledge conflict scenarios.
By holding the available multilingual context fixed while varying \textit{only} the query language, \textsc{Waldo} enables a counterfactual test of whether query language alone changes the knowledge source on which a model bases its answer.

Using \textsc{Waldo}, we evaluate eight models across five languages (\S\ref{sec:setup}) and diagnose the extent to which they exhibit query-language preference. 
Under knowledge gaps, models largely overcome this preference, producing consistent responses across query languages.
Under knowledge conflicts, however, their response distributions differ substantially by query language (\S\ref{sec:diagnosis}).
These findings motivate our exploration of potential approaches for mitigating query-language preference in the presence of conflicting evidence.
We introduce Query-language Preference Head Ablation (\textbf{\textsc{QPHA}}) and adapt Quantized Low-Rank Adapter (\textbf{\textsc{QLoRA}}) training \citep{qlora} as two targeted methods (\S\ref{sec:mitigation}), and show both help reduce the preference gap (\S\ref{sec:mitigation_res}).
More broadly, \textsc{Waldo} provides an evaluation testbed for studying how language bias interacts with answer quality in multilingual systems. Our initial analyses and mitigation approaches open new directions for understanding the underlying sources of these biases and improving the robustness of cross-lingual knowledge transfer.

We make three main contributions in this work:
\begin{itemize}[leftmargin=15pt, itemsep=2pt, parsep=-1pt]
    \item \textbf{Benchmark:} We present \textsc{Waldo}, a multilingual QA benchmark comprising 12K queries targeting cross-lingually heterogeneous facts drawn from Wikipedia. 
    It supports the study of query-language preference under knowledge gaps, where facts are unevenly available across languages, and knowledge conflicts, where languages provide incompatible accounts of the same fact.
    
    \item \textbf{Diagnosis:} We introduce two metrics for quantifying query-language preference and systematically evaluate eight models across five languages. We find little preference under knowledge gaps, but show that it becomes pronounced under knowledge conflicts.

    \item \textbf{Mitigation:} We evaluate two approaches targeting query-language preference under conflicting evidence: \textsc{QPHA} and \textsc{QLoRA} training with two customized objectives. Both methods effectively reduce preference, with \textsc{QLoRA} reducing the gap by up to 61.5\%.

\end{itemize}

\section{Background: Definition, Evaluation, and Mitigation}
\label{sec:related}

\noindent\textbf{Definition.}
Language preference refers to a model's systematic tendency to favor knowledge sources in some languages over others.
Such preference may arise from unequal language representation in pre-training data \citep{joshi2020state} or from cross-lingual tokenization differences \citep{petrov2023token_unfairness}.
Query-language preference is a specific form of language preference in which a model favors sources written in the same language as the query \citep{ki2026linguistic}.
Prior work has studied how query-language preference manifests in parametric knowledge \citep{goldman2025eclekticnovelchallengeset}, multilingual retrieval \citep{yang2024language, park-lee-2025-investigating}, and generation \citep{sharma-etal-2025-faux, li2025multilingual, ki2026linguistic}. 
We focus on \textit{generation}, holding the multilingual retrieved context fixed to isolate how query language affects the knowledge source on which a model bases its response.

\noindent\textbf{Datasets and Evaluation.}
Existing work studies query-language preference in two broad settings: settings that assume semantically equivalent knowledge is available across languages, and simulated settings with limited ecological validity \citep{devries2020ecologicallyvalidresearchlanguage}.
One line of work retrieves naturally occurring evidence from real-world sources \citep{qi-etal-2025-consistency, park-etal-2026-enhancing, huo-etal-2026-breaking, wang-etal-2026-languages}. However, they cannot isolate query-language preference when language editions differ in the availability or content of relevant facts.
Conversely, \citet{sharma-etal-2025-faux} introduces cross-lingual knowledge conflicts, but uses fictional documents. 
A separate body of work documents cross-lingual knowledge gaps \citep{massa2012manypedia, bao2012omnipedia, miquel2018wikipedia, miquel2020wikipedia, hupkes2025multilokomultilinguallocalknowledge} as well as factual and narrative inconsistencies across Wikipedia language editions \citep{samir-etal-2024-locating, li-etal-2024-land, zhao2026abundanceconcealsweaknessknowledge}. Yet, this work generally neither provides a reusable benchmark nor connects such disparities to query-language preference.
\textsc{Waldo} addresses these gaps by targeting naturally occurring, cross-lingually heterogeneous facts from Wikipedia, covering knowledge gaps and controlled knowledge conflicts.


\noindent\textbf{Mitigation.}
Prior efforts to mitigate language preference have primarily targeted the knowledge-source retrieval stage, including calibrated relevance estimation \citep{park-etal-2026-enhancing, wang-etal-2026-languages, jang2026mlairemultilinguallanguageawareinformation}, query translation \citep{amiraz-etal-2025-cross}, and language-controlled retrieval \citep{huo-etal-2026-breaking, qi-etal-2026-language, lee-etal-2026-clear, hong2026lamaropenlanguageawaremultilingual}.
At the generation stage, existing approaches translate retrieved documents into English \citep{ranaldi2026multilingual} or into the query language \citep{park-lee-2025-investigating} to unify the evidence in a single pivot language.
To our knowledge, no prior work has specifically studied \textit{query}-language preference at generation time or explored mechanistic interventions or fine-tuning approaches to mitigate it; we take a first step toward doing so.



\section{Creating \textsc{Waldo}}
\label{sec:dataset}

Wikipedia is a valuable source of cross-lingual knowledge disparities: documents about the same entity in different language editions are not merely translations of one another \citep{hecht2013mining}, but contain localized knowledge \citep{haraway1988situated, kumar2021digital} shaped by the distinct contributor communities that produce and maintain them \citep{hale2014multilinguals, sen2015barriers, johnson2022considerations}. 
For each source language, we collect Wikipedia documents and their contributor-authored revision histories from publicly available Wikipedia dumps available as of June 1, 2026.\footnote{\url{https://dumps.wikimedia.org/}}
We retain revisions made on or after January 1, 2020, representing each revision as a diff to its parent revision.

\noindent\textbf{Source Languages.}
We collect documents in five source languages spanning diverse resource levels, language families, scripts, and Wikipedia sizes: French (fr), Simplified Chinese (zh), Japanese (ja), Korean (ko), and Bengali (bn). Per-language characteristics are detailed in Appendix~\ref{appendix:lang_stats}.

We refer to the complete collection of revisions per language as the \textit{full} set, which we plan to release. For our experiments, we sample five dump files per language to form a balanced \textit{evaluation} set.

\subsection{\textcolor{first}{Knowledge Gap} [\raisebox{-0.2em}{\includegraphics[height=1em]{figures/logo/one.png}}]}
\label{sec:gap}

\noindent\textbf{Heuristics \& LLM Annotation.}
To retain \textit{meaningful} revisions for evaluation, we filter candidate revisions using a combination of heuristic rules and LLM-based annotation with \textsc{GPT-5.6-Terra} \citep{openai2026gpt56}.
After manual auditing, we define a suite of rules that a revision must (1) not be a bot edit or maintenance-related change (e.g., a revert); (2) have a length within a predefined range; (3) introduce at least one factual statement without removing information; and (4) contain an introduced fact that remains in the latest version of the source-language Wikipedia document but is absent from the latest linked English document, identified via the same Wikidata QID.

During LLM annotation, we also record a \texttt{timestamp}, formatted as \textsc{YYYY-MM}, indicating the month in which the introduced fact is asserted to have occurred.
This procedure retains, on average, 0.3\% of candidate revisions across languages.
See Appendix~\ref{appendix:heuristics} for filtering details, examples of excluded revisions, and per-language retain rates.

\noindent\textbf{QA Generation.}
For each retained revision, we provide \textsc{GPT-5.6-Terra} with the revision, its source-language document, and the corresponding English document.
The model generates \textit{aligned} QA pairs in the source language ($\mathcal{Q}_\mathrm{src},\mathcal{A}_\mathrm{src}$) and English ($\mathcal{Q}_\mathrm{en},\mathcal{A}_\mathrm{en}$), which are translations of one another and target the same fact introduced by the revision.
We retain a revision only if the source-language and English sets contain the same number of QA pairs and every aligned QA pair achieves an \textsc{xCOMET} translation-quality score of at least 80.0 \citep{guerreiro2023xcomettransparentmachinetranslation}. 
Further details and per-language \textsc{xCOMET} scores are provided in Appendix~\ref{appendix:qa_translation}.


\subsection{\textcolor{second}{Knowledge Conflict} [\raisebox{-0.2em}{\includegraphics[height=1em]{figures/logo/two.png}}]}
\label{sec:conflict}

\noindent\textbf{Conflict Injection \& QA Generation.}
For each revision retained in \S\ref{sec:gap}, we introduce a controlled knowledge conflict by prompting an LLM (\textsc{GPT-5.6-Terra}) to inject an inconsistent version of the target fact into the corresponding English document and update the answers to the same questions \citep{chen-etal-2022-rich, mo-etal-2026-faithfulness, zhao2026abundanceconcealsweaknessknowledge, conflict_2026}.
To simulate real-world knowledge conflicts, we follow prior conflict-injection practices \citep{longpre-etal-2021-entity, wang2024resolving}. Each injected conflict must satisfy two criteria: (1) it modifies a core attribute of the target fact, such as a name, quantity, date, or location, rather than merely negating it (e.g., ``is'' $\rightarrow$ ``is not''); and (2) it is applied consistently throughout the document while preserving the original tone. \autoref{tab:conflict_type} reports the distribution of conflict types per language, with numerical and date changes being the most common on average. Appendix~\ref{appendix:conflict_examples} provides qualitative examples of each conflict  type.


\noindent\textbf{Answerability Check.}
To verify that each conflict-injected English document supports its revised answer, we use \textsc{GPT-5.6-Sol}, a different model from the generator, to answer each original English question using the conflict-injected document.
We retain only QA pairs that pass this answerability check \citep{sun-etal-2026-task}; on average, 98.8\% of pairs are retained across languages.
Our final evaluation set contains approximately 12K QA pairs across five source languages, each of which passes the procedures for both knowledge gaps (\S\ref{sec:gap}) and conflicts (\S\ref{sec:conflict}). More details are in Appendix~\ref{appendix:answerability}.

For each evaluated model, we retain only QA pairs whose (1) \texttt{timestamp} postdates the model's reported knowledge cutoff and (2) when given only the question, the model answers incorrectly in both the source language and English, giving an average retention rate of 23.8\%.
This reduces the likelihood that a model can answer from parametric knowledge alone \citep{yeh-li-2026-retrieved} or preferentially follows evidence consistent with its prior beliefs (i.e., confirmation bias; \citet{jin2024better}).
It therefore allows observed response differences to be attributed more directly to the provided context.

\begin{wraptable}{r}{0.48\textwidth}
   \vspace{-1.2em}
    \centering
    \resizebox{\linewidth}{!}{%
        \begin{tabular}{lrrrrrr}
        \toprule
        \textbf{Language} & 
        \textbf{Date} & 
        \textbf{Num} &
        \textbf{Per} &
        \textbf{Loc} &
        \textbf{Org} &
        \textbf{Misc} \\
        \midrule

        \textbf{French} & 11.3 & 20.0 & 15.9 & 7.6 & 10.6 & 34.6 \\
        \textbf{Chinese} & 17.5 & 26.3 & 8.6 & 9.7 & 9.6 & 28.3 \\
        \textbf{Japanese} & 17.9 & 18.4 & 9.9 & 8.5 & 12.0 & 33.3 \\
        \textbf{Korean} & 27.9 & 12.6 & 14.8 & 5.1 & 8.2 & 31.4 \\
        \textbf{Bengali} & 14.8 & 25.4 & 10.1 & 6.9 & 8.8 & 34.0 \\
        \rowcolor{gray!15}
        \textbf{Average} & 17.9 & 20.5 & 11.9 & 7.6 & 9.8 & 32.3 \\

        \bottomrule
        \end{tabular}
    }
\vspace{-0.8em}
\caption{\textbf{Distribution of injected conflict types per language (\%).}
\textbf{Num:} Number; \textbf{Per:} Person; \textbf{Loc:} Location; \textbf{Org:} Organization.}
\vspace{-1.5em}
\label{tab:conflict_type}
\end{wraptable}


\section{Experiment Setup}
\label{sec:setup}

\subsection{Study Formulation}
\label{sec:formulation}
Each \textsc{Waldo} instance contains an aligned QA pair, a source-language document, its corresponding English document, a conflict-injected English document, and the revised English answers.

\noindent\textbf{Context.}
For each instance, the model receives the source-language document together with either the original English document (in knowledge gap) or its conflict-injected version (in knowledge conflict), presented in randomized order to control for position bias \citep{liu-etal-2024-lost}.
We hold this multilingual context fixed while varying \textit{only} the query language. Thus, any systematic difference in model responses across the two queries reflects the effect of the query language.


\noindent\textbf{Response.}
For each knowledge gap and conflict setting $s$, given the fixed context $\mathcal{C}_s$ and a query $\mathcal{Q}_{\ell}$ in language $\ell \in \{\mathrm{src}, \mathrm{en}\}$, model $\mathcal{M}$ produces $\hat{\mathcal{A}}_{\ell,s}$.
For each model, since we retain only target facts that postdate its knowledge cutoff, the outputs are less likely to be determined by its parametric knowledge than by the provided context.
Thus, if a model is invariant to query language, changing the query language should \textit{not} systematically alter which language-specific account it follows.

\begin{definitionbox}
We say that a model exhibits \textbf{query-language preference} when it follows the source-language account in response to a source-language query but does not provide the equivalent English answer in response to the English query. 
Under knowledge gaps, this may manifest as abstention or an unsupported answer to the English query; under knowledge conflicts, it may appear as instead selecting the conflict-injected English answer.
\end{definitionbox}


\noindent\textbf{Models.}
We evaluate eight models that satisfy two criteria:  they have publicly reported knowledge cutoffs and context windows of at least 128K tokens to accommodate Wikipedia-length documents.
To assess the generality of query-language preference, the suite spans model sizes, families, and knowledge cutoffs: \textsc{LLaMA-3.1} 8B \citep{grattafiori2024llama3herdmodels}, \textsc{Qwen-3} 8B \citep{yang2025qwen3technicalreport}, and \textsc{Aya-Expanse} 8B \citep{dang2024ayaexpansecombiningresearch} are open-weight models; \textsc{GPT-4o} \citep{openai2024gpt4ocard}, \textsc{GPT-4.1} \citep{openai2025gpt41}, \textsc{DeepSeek-V4-Pro} \citep{deepseekai2026deepseekv4highlyefficientmilliontoken}, \textsc{Grok-4.3},\footnote{\url{https://docs.x.ai/developers/models/grok-4.3}} and \textsc{GPT-5.5} \citep{openai2026gpt55} are closed-source models.
Per-model details are in Appendix Table~\ref{tab:model_statistics}.

\subsection{Evaluation Metrics}
\label{sec:metrics}
We use two complementary metrics for quantifying whether models recover the original target fact as the query language changes.
Answer agreement measures whether a model response matches the reference answer expressed in the query language.
Answer probability is the length-normalized likelihood assigned to the answer supported by the source-language document.



\noindent\textbf{Answer Agreement.}
For each QA pair, we use \textsc{GPT-5.6-Terra} to determine whether the model response matches the reference answer in the query language.
We do not directly compare responses across languages to avoid potentially less reliable cross-lingual model judgments \citep{fu2026languagesdisagreeselfevolvingmultilingual}.

For each knowledge gap and conflict setting \(s\), we compute the percentage of QA pairs in \(\mathcal{P}_s\) whose model response \(\hat{\mathcal{A}}_{\ell,s}\) matches the reference answer \(\mathcal{A}_\ell\) under query language \(\ell\):
\begin{equation}
\mathrm{Agreement}({\ell},s)
=
\frac{1}{|\mathcal{P}_s|}
\sum_{i\in\mathcal{P}_s}
\mathrm{Agree}\!\left(
\hat{\mathcal{A}}_{\ell,s}^{(i)},
\mathcal{A}_{\ell}^{(i)}
\right)
\times 100,
\end{equation}
where \(\mathrm{Agree}(\cdot,\cdot)\) is 1 if they match and 0 otherwise. We define the query-language agreement gap:
\begin{equation}
\Delta_\mathrm{agree}(s)
=
\mathrm{Agreement}(\mathcal{Q}_{\mathrm{src}}, s)
-
\mathrm{Agreement}(\mathcal{Q}_{\mathrm{en}}, s).
\end{equation}
A large positive \(\Delta_\mathrm{agree}(s)\) indicates that the model more often recovers the original target fact when queried in the source language than in English, consistent with query-language preference.



\noindent\textbf{Answer Probability.}
Answer agreement captures whether a decoded response matches the source-supported reference answer, but not how \textit{strongly} the model favors that answer.
Thus, we compute the length-normalized conditional likelihood of the source-supported answer \(\mathcal{A}_{\mathrm{src}}=(t_1,\ldots,t_k)\) under each query language $\ell$:
\begin{equation}
P_{\theta}(\mathcal{A}_{\mathrm{src}} \mid \mathcal{Q}_{\ell}, \mathcal{C}_s)
=
\left[
\prod_{m=1}^{k}
p_{\theta}(t_m \mid \mathcal{Q}_{\ell}, \mathcal{C}_s, t_{<m})
\right]^{1/k},
\end{equation}
where \(p_{\theta}\) is the model's next-token distribution, and \(t_{<m}=(t_1,\ldots,t_{m-1})\). Length normalization makes scores comparable across answers of different lengths.
We average this score across evaluated QA pairs (yielding $\mathrm{Prob}$) and report the source-answer likelihood shift:
\begin{equation}
\Delta_{\mathrm{prob}}(s)
=
\mathrm{Prob}(\mathcal{Q}_\mathrm{src},s)
-
\mathrm{Prob}(\mathcal{Q}_\mathrm{en},s).
\end{equation}
A positive \(\Delta_{\mathrm{prob}}(s)\) means that the model assigns greater likelihood to the source-supported answer when queried in the source language than in English. 
Conversely, more comparable likelihoods across query languages could help systems surface cross-lingual disparities.
\section{Diagnosis Results}
\label{sec:diagnosis}

\begin{figure*}
    \centering
    \includegraphics[width=\linewidth]{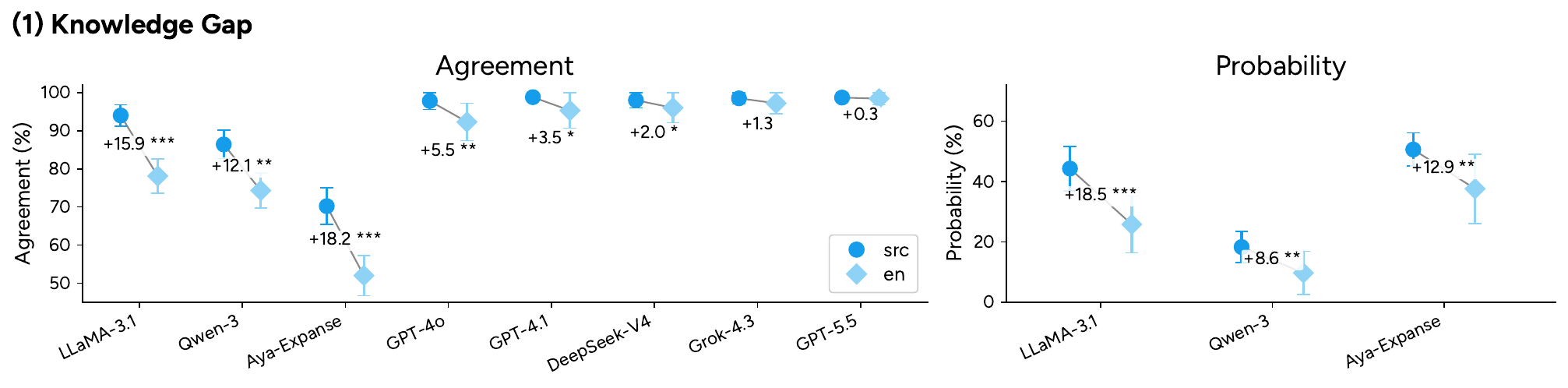}
    \includegraphics[width=\linewidth]{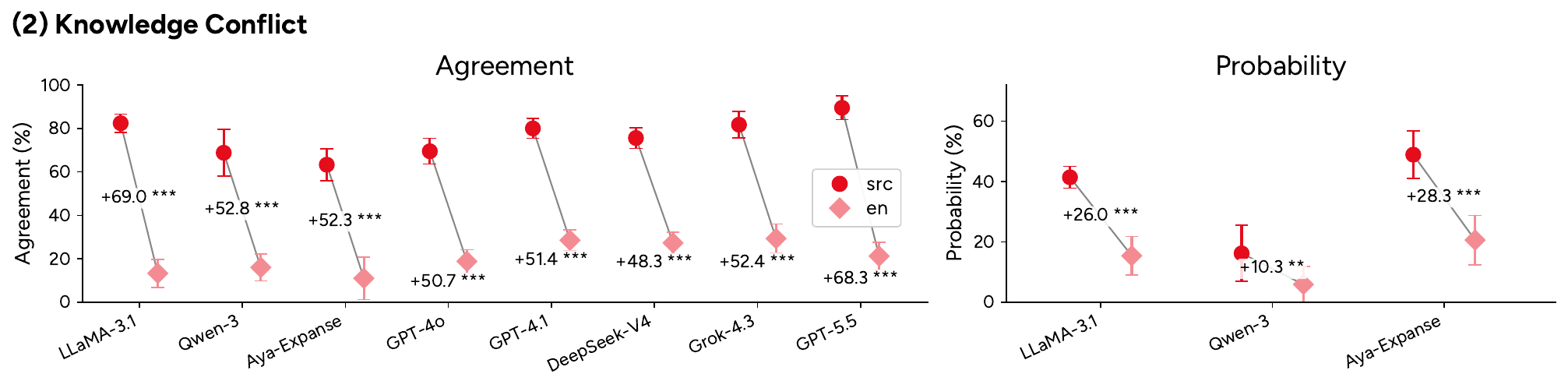}
    \vspace{-1em}
    \caption{\textbf{Answer agreement and probability under \textcolor{first}{knowledge gap} and \textcolor{second}{conflict} settings.}
    A larger positive $\Delta$(src-en) denotes stronger query-language preference. 
    Error bars are 95\% bootstrap CIs.
    We use McNemar's test for agreement differences and the Wilcoxon signed-rank test for probability differences, with Bonferroni correction for multiple comparisons. *: significant with $p$ $<$ .05; **: $p$ $<$ .01; ***: $p$ $<$ .001; unmarked: not significant.
    See Appendix~\ref{appendix:per_lang} for per-language results.}
    \vspace{-1em}
    \label{fig:main_results}
\end{figure*}

\autoref{fig:main_results} presents our diagnosis for query-language preference.\footnote{While our primary experiments use a controlled two-document context (\S\ref{sec:formulation}) to isolate the effect of query language, Appendix~\ref{appendix:endtoend} shows that the results generalize when an external retrieval component is added.} We highlight four findings below.

\noindent\textbf{Closed-source models largely overcome query-language preference under knowledge gaps.}
As shown in~\autoref{fig:main_results} (\textit{top}), closed-source models show little to no statistically significant difference in answer agreement across query languages.
In contrast, open-weight models exhibit significant agreement gaps: 15.9\% for \textsc{LLaMA-3.1}, 12.1\% for \textsc{Qwen-3}, and 18.2\% for \textsc{Aya-Expanse}. 
Notably, among the three, \textsc{Aya-Expanse} shows the largest gap despite explicit training on diverse languages and multilingual tasks \citep{dang2024ayaexpansecombiningresearch}.
Answer probability shows a similar pattern: all three open-weight models assign significantly different likelihoods to the source-supported answer across query languages.
These findings suggest that, when the required knowledge to correctly answer a query is available only in one language, closed-source models generally recover it regardless of the query language, whereas the smaller open-weight models often fail to.

\noindent\textbf{Knowledge conflicts amplify query-language preference.}
Under knowledge conflicts (shown in \autoref{fig:main_results}, \textit{bottom}), query-language preference becomes substantially more pronounced.
All models exhibit statistically significant answer-agreement gaps between source-language and English queries, with an average gap of 55.7\% across models.
The answer-probability gap likewise increases relative to the knowledge gap setting, by 7.5 \textit{pp} for \textsc{LLaMA-3.1}, 1.7 \textit{pp} for \textsc{Qwen-3}, and 15.4 \textit{pp} for \textsc{Aya-Expanse}.
When disaggregated by source language (Appendix~\ref{appendix:per_lang}), probability gaps are largest for lower-resource languages such as Bengali and generally decline with language-resource availability, reaching their lowest values for French.
Overall, even closed-source models that show weak preference under knowledge gaps preferentially follow the query-language-matched document when evidence conflicts across languages.\footnote{Consistent with this account, models assign higher likelihood to the conflicting English-supported answer under English queries than source-language queries; see Appendix~\ref{appendix:conflict}. Also, as a control, adding the English fact to the English document reduces the preference gap to near zero (Appendix~\ref{appendix:truth}).}

\noindent\textbf{Query-language preference persists when the document-language asymmetry is reversed.}
By design, \textsc{Waldo} is asymmetric: the original target fact is in the source-language document, while it is absent (gap) or contains an incompatible version (conflict) in the English document.
Is the query-language preference observed in \textsc{Waldo} driven merely by the asymmetric construction of its QA pairs?
To test this, we reverse the knowledge conflict condition: the English document now contains the original target fact, whereas the source-language document contains the conflicting account.
We use \textsc{GPT-5.6-Terra} to translate and incorporate each fact into the appropriate document language (see Prompt A.9 of Appendix~\ref{appendix:prompts}).
As shown in~\autoref{tab:swapped}, all models continue to exhibit strong query-language preference after the document roles are swapped, with the direction of the effect correspondingly reversed.
This result indicates that the strong preference observed under knowledge conflicts is not an artifact of \textsc{Waldo}'s particular asymmetry but generalizes across both directions.

\begin{wraptable}{r}{0.6\textwidth}
   \vspace{-1.2em}
    \centering
    \resizebox{\linewidth}{!}{%
    \begin{tabular}{l
                    S[table-format=2.1]
                    S[table-format=2.1]
                    S[table-format=-2.1, table-space-text-post={\sig{***}}]
                    S[table-format=2.1]
                    S[table-format=2.1]
                    S[table-format=2.1, table-space-text-post={\sig{***}}]}
        \toprule
        \multirow{2}{*}{\textbf{Model}}
          & \multicolumn{3}{c}{\textbf{Agreement (\%)}}
          & \multicolumn{3}{c}{\textbf{Probability (\%)}} \\
        \cmidrule(lr){2-4} \cmidrule(lr){5-7}
        & $\mathrm{src}$ & $\mathrm{en}$ & {$\Delta_\mathrm{agree}$} & $\mathrm{src}$ & $\mathrm{en}$ & {$\Delta_\mathrm{prob}$} \\
        \midrule
        \textsc{LLaMA-3.1}       & 21.0 & 86.3 & -65.3{\sig{***}} & 41.4 & 15.4 & 26.0{\sig{***}} \\
        \textsc{Qwen-3}          & 28.7 & 81.6 & -52.9{\sig{***}} &  7.9 &  4.9 &  2.9{\sig{*}}   \\
        \textsc{Aya-Expanse}     & 24.3 & 69.4 & -45.1{\sig{***}} & 17.9 & 15.4 &  2.4{\sig{*}}   \\
        \midrule
        \textsc{GPT-4o}          & 15.6 & 67.7 & -52.1{\sig{***}} & \na & \na & \na \\
        \textsc{GPT-4.1}         & 25.3 & 77.9 & -52.6{\sig{***}} & \na & \na & \na \\
        \textsc{DeepSeek-V4-Pro} & 22.5 & 72.6 & -50.1{\sig{***}} & \na & \na & \na \\
        \textsc{Grok-4.3}        & 27.1 & 80.9 & -53.8{\sig{***}} & \na & \na & \na \\
        \textsc{GPT-5.5}         & 20.4 & 76.5 & -56.1{\sig{***}} & \na & \na & \na \\
        \bottomrule
    \end{tabular}%
    }
\vspace{-0.5em}
\caption{\textbf{Answer agreement and probability gaps under knowledge conflict for swapped roles.} Same statistical test performed as in~\autoref{fig:main_fig}. *: significant with $p$ $<$ .05; ***: $p$ $<$ .001. No probabilities for closed-source models (--).}
\vspace{-1.5em}
\label{tab:swapped}
\end{wraptable}

\noindent\textbf{Query-language preference is not explained by measured alternative factors.}
Could the observed preference reflect other properties of the data rather than the query language itself?
We investigate this by testing four potential confounders\textemdash pre-training effect, embedding similarity, surface cues, and query-topic locality\textemdash but find no significant associations with query-language preference metrics (Appendix~\ref{appendix:hypotheses}).
Together, the results suggest that the measured effect of query-language preference is not accounted for by the examined confounders.\footnote{Appendix~\ref{appendix:more_metrics} reports additional evidence from F1 score gap, citation attribution, and response-language analyses, which corroborates the answer agreement and probability results.}


\section{Mitigation for Knowledge Conflicts}

Under conflicting evidence across languages, models \textit{strongly} favor the document in the query language (\S\ref{sec:diagnosis}).
As an initial exploration of how this behavior may be mitigated, we study two targeted approaches (\S\ref{sec:mitigation}) and evaluate their effectiveness in reducing query-language preference (\S\ref{sec:mitigation_res}).

\subsection{Mitigation Approaches}
\label{sec:mitigation}

\noindent\textbf{Query-language Preference Head Ablation (\textsc{QPHA}).}
Prior work suggests that individual attention heads can mediate distinct information sources (e.g., context versus parametric knowledge \citep{jin-etal-2024-tug, zhu2026focus}), and that head ablation can mitigate conflict-driven behavior \citep{jin-etal-2024-cutting, li2025taming}.
Building on this, we investigate whether query-language preference is associated with a localized set of attention heads and whether ablating these heads mitigates the behavior.
We identify candidate heads by adapting the comparison underlying our metrics (\S\ref{sec:metrics}).

For each conflict instance in \textsc{Waldo}, we score the source-supported answer $\mathcal{A}_\mathrm{src}$ and the conflicting English-supported answer $\mathcal{A}^{\prime}_\mathrm{en}$ by their length-normalized log-likelihood under source-language and English queries.
We then use gradient--activation attribution \citep{sixteen_2019, jiang-etal-2025-hicd, focus_2026} to measure each attention head's contribution to these likelihoods $\mathcal{S}_{\ell}(\mathcal{A})$:
\begin{equation}
\mathcal{S}_{\ell}(\mathcal{A})
=
\frac{1}{k}
\sum_{i=1}^{k}
\log p_{\theta}\!\left(
t_i
\mid
\mathcal{Q}_{\ell},
\mathcal{C}_{\mathrm{conflict}},
t_{<i}
\right),
\qquad
\mathrm{Attr}_{l,h}(\mathcal{A}, \ell)
=
\sum \mathcal{O}^{(l,h)}
\odot
\frac{\partial \mathcal{S}_{\ell}(\mathcal{A})}
{\partial \mathcal{O}^{(l,h)}},
\label{eq:qpha_attr}
\end{equation}
where \(\mathcal{O}^{(l,h)}\) is the output of head \(h\) in layer
\(l\), and the summation aggregates over its activation dimensions.
Here, \(\mathcal{A}=(t_1,\ldots,t_k)\) and \(t_{<i}=(t_1,\ldots,t_{i-1})\) denotes the preceding answer tokens.

We identify query-language preference heads using the following contrast:
\begin{equation}
\mathrm{QPHA}_{l,h}
=
\left[
\mathrm{Attr}_{l,h}(\mathcal{A}_{\mathrm{src}}, \mathrm{src})
-
\mathrm{Attr}_{l,h}(\mathcal{A}_{\mathrm{en}}', \mathrm{src})
\right]
-
\left[
\mathrm{Attr}_{l,h}(\mathcal{A}_{\mathrm{src}}, \mathrm{en})
-
\mathrm{Attr}_{l,h}(\mathcal{A}_{\mathrm{en}}', \mathrm{en})
\right].
\label{eq:qpha}
\end{equation}
A large positive score indicates a head's attribution shifts toward the source-supported answer under source-language queries and the conflicting English-supported answer under English queries.
We average this score across conflict instances and select the \(k\) heads with the largest positive, statistically significant mean scores (under a two-sided Wilcoxon signed-rank test).

To examine the trade-off between intervention sparsity and strength \citep{kang2025largevisionlanguagemodelneeds}, we evaluate $k \in \{2,4,8\}$ with two ablation modes: \textbf{Zero}, which sets the selected head outputs to zero \citep{sixteen_2019}, and \textbf{Mean}, which replaces each output with its sequence-wise mean activation, reducing potential out-of-distribution effects of zero ablation \citep{wang2023interpretability}. 
For each, we compare against a \textbf{Random} control that ablates $k$ randomly selected heads \citep{crosbie-shutova-2025-induction}.

\begin{figure*}
    \centering
    \includegraphics[width=\linewidth]{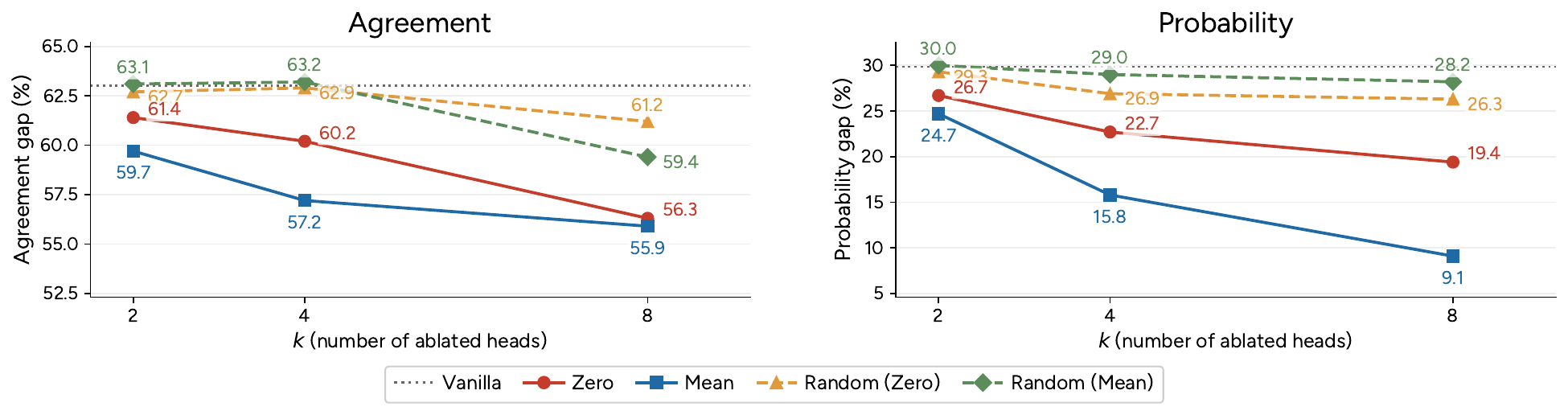}
    \vspace{-1.5em}
    \caption{\textbf{Answer agreement and probability gaps with varying $k$.} Both gaps shrink as more attention heads are ablated (i.e., increasing $k$), and \textsc{QPHA} yields a larger reduction than the random control. Shown for \textsc{LLaMA-3.1}; see Appendix~\ref{appendix:other_models} for other models.}
    \label{fig:increasing_k}
\end{figure*}


\noindent\textbf{\textsc{QLoRA}.}
Whereas \textsc{QPHA} is a localized inference-time intervention, we ask whether a training-time adaptation can more effectively mitigate query-language preference by encouraging models to recover the original target fact regardless of query language. 
We fine-tune models using Quantized Low-Rank Adapters (\textsc{QLoRA}; \citealp{qlora}) with two customized objectives, both based on the length-normalized log-likelihood \(\mathcal{S}_{\ell}(\mathcal{A})\) defined in Eq.~\ref{eq:qpha_attr}:
\begin{itemize}[leftmargin=15pt, itemsep=2pt, parsep=-1pt]
    \item \textbf{Promote:} We promote the likelihood of the original English answers under English queries:
    \begin{equation}
    \mathcal{L}_{\mathrm{promote}} = -\,\mathcal{S}_{\mathrm{en}}(\mathcal{A}_{\mathrm{en}}).
    \end{equation}
    \item \textbf{Contrastive:} We additionally set models to forget the conflicting English-supported answer and retain the source-supported answer under source-language queries \citep{nam2025causal}:
    \begin{equation}
    \mathcal{L}_{\mathrm{contrastive}}
    = \underbrace{\mathcal{S}_{\mathrm{en}}(\mathcal{A}_{\mathrm{en}}^{\prime})}_{\text{Forget}}
    - \underbrace{\mathcal{S}_{\mathrm{src}}(\mathcal{A}_{\mathrm{src}})}_{\text{Retain}}
    - \lambda\,\underbrace{\mathcal{S}_{\mathrm{en}}(\mathcal{A}_{\mathrm{en}})}_{\text{Promote}},
    \end{equation}
    with the Forget$-$Retain term clipped to $[-20,20]$ for gradient stability.
\end{itemize}


As both methods require access to model internals, we evaluate them only on the three open-weight models. 
We further restrict to knowledge-conflict instances that exhibit query-language preference, yielding 5.1K QA pairs in total across models (1.9K revisions; 52.3\% retain rate). 
For a fair comparison, we allocate 20\% of revisions to training and development: \textsc{QLoRA} uses it for training and model selection while \textsc{QPHA} uses it for head identification.  We use the remaining 80\% for evaluating both methods. See Appendix~\ref{appendix:implementation} for implementation details.


\vspace{-0.5em}
\subsection{Results \& Analyses}
\label{sec:mitigation_res}
\vspace{-0.3em}
\noindent\textbf{\textsc{QPHA} reduces query-language preference.}
In~\autoref{tab:mitigation}, we observe that \textsc{QPHA} ($k=8$) reduces both agreement and probability gaps across models.
Zero and mean ablation each reduce the agreement gap by 4.8 \textit{pp} on average, exceeding their respective random controls (1.3 \textit{pp} for zero and 2.1 \textit{pp} for mean). 
Similarly for the probability gap, mean ablation of \textsc{QPHA}-selected heads yields the largest reduction (9.2 \textit{pp}), followed by zero ablation (5.3 \textit{pp}), again both exceeding their respective random controls (1.2 \textit{pp} for mean and 2.1 \textit{pp} for zero).
These improvements over random ablation suggest that \textsc{QPHA} identifies heads more strongly associated with query-language preference.
Across most model--ablation-mode settings, ablating these heads reduces the preference gap, providing an initial proof-of-concept for a mechanistic head-level intervention.


\begin{wraptable}{r}{0.59\textwidth}
\vspace{-1.2em}
\centering
\resizebox{\linewidth}{!}{%
\begin{tabular}{l *{6}{S[table-format=2.1, table-space-text-post={\sig{***}}]}}
    \toprule
    \multirow{2}{*}{\textbf{Approach}}
      & \multicolumn{2}{c}{\textbf{\textsc{LLaMA-3.1}}}
      & \multicolumn{2}{c}{\textbf{\textsc{Qwen-3}}}
      & \multicolumn{2}{c}{\textbf{\textsc{Aya-Expanse}}} \\
    \cmidrule(lr){2-3} \cmidrule(lr){4-5} \cmidrule(lr){6-7}
      & {$\Delta_\mathrm{agree}$} & {$\Delta_\mathrm{prob}$}
      & {$\Delta_\mathrm{agree}$} & {$\Delta_\mathrm{prob}$}
      & {$\Delta_\mathrm{agree}$} & {$\Delta_\mathrm{prob}$} \\
    \midrule
    Vanilla         & 63.0 & 29.9 & 59.0 & 12.1 & 58.8 & 33.5 \\
    \midrule
    \rowcolor{gray!20}
    \multicolumn{7}{c}{\textsc{\textbf{QPHA}} ($k=$ 8)} \\
    \midrule
    Random (Zero)   & 61.2 & 26.3{\sig{*}} & 57.0 & 12.1 & 58.6 & 30.8{\sig{*}} \\
    Random (Mean)   & 59.4{\sig{*}} & 28.2 & 57.5 & 11.8 & 57.5 & 31.8 \\
    Zero            & 56.3{\sig{**}} & 19.4{\sig{***}} & 54.3{\sig{*}} & 10.5 & 55.7{\sig{*}} & 29.7{\sig{*}} \\
    Mean            & 55.9{\sig{**}} &  9.1{\sig{***}} & 55.4{\sig{*}} & 10.1{\sig{*}} & 55.0{\sig{*}} & 28.7{\sig{**}} \\
    \midrule
    \rowcolor{gray!20}
    \multicolumn{7}{c}{\textbf{\textsc{QLoRA}}} \\
    \midrule
    Promote         & 2.3{\sig{***}} & 4.3{\sig{***}} & 4.3{\sig{***}} & 7.7{\sig{**}}  & 0.3{\sig{***}} & 4.9{\sig{***}} \\
    Contrastive     & 1.5{\sig{***}} & 6.4{\sig{***}} & 1.4{\sig{***}} & 5.3{\sig{***}} & 0.9{\sig{***}} & 1.7{\sig{***}} \\
    \bottomrule
\end{tabular}%
}
\vspace{-0.8em}
\caption{\textbf{Mitigation results on answer agreement and probability gaps.} For both, lower is better (↓).
We perform Wilcoxon signed-rank test against Vanilla, with Bonferroni correction. 
*: significant with $p$ $<$ .05; **: $p$ $<$ .01; ***: $p$ $<$ .001; unmarked: not significant.
Appendix~\ref{appendix:mitigation_res} reports full results for \(k \in \{2,4,8\}\) and each query language.}
\vspace{-1.4em}
\label{tab:mitigation}
\end{wraptable}

\noindent\textbf{Small-scale \textsc{QLoRA} adaptation yields larger reductions.}
Fine-tuning \textsc{QLoRA} adapters with targeted objectives further reduces query-language preference more than inference-time \textsc{QPHA}. 
The promote objective closes 58.0\% of the agreement gap and 19.5\% of the probability gap, while the contrastive objective pushes this further to 59.0\% and 20.7\% each.
Overall, using only 10\% of the data for training, \textsc{QLoRA} substantially attenuates query-language preference under knowledge conflicts, which also further generalizes to the swapped-roles setting (Appendix~\ref{appendix:mitigation_swapped}).


\noindent\textbf{\textsc{QPHA}'s effect grows with the number of ablated heads.}
As illustrated in~\autoref{fig:increasing_k}, zero and mean ablation increasingly reduce agreement and probability gaps as \(k\) grows from 2 to 8.
This pattern supports that \textsc{QPHA} identifies heads associated with query-language preference: ablating more highly ranked heads yields larger reductions. 
Notably, ablating $k =$ 2 heads already significantly reduces the gap relative to the vanilla baseline ($p$ $<$ .05), showing that the preference can be reduced through a small head-level intervention. List of heads is in Appendix~\ref{appendix:ablated_heads}.

\noindent\textbf{Which one to use?}
\textsc{QPHA} provides a lightweight, training-free intervention that directly incorporates the mechanism underlying query-language preference, and is effective with only a few ablated heads. 
Yet, ablating more heads may affect model behavior beyond query-language preference.
In contrast, \textsc{QLoRA} achieves substantially larger reductions, but requires training and knowledge of the desired and conflicting answers across documents to define its objectives. 
Given these trade-offs, future work could develop more robust approaches that retain the advantages of both methods.

\vspace{-0.5em}
\section*{Conclusion \& Discussion}

We study query-language preference in multilingual systems under two forms of cross-lingual knowledge disparities. 
Using \textsc{Waldo}, we find that models largely overcome this preference under knowledge gaps, but exhibit it strongly under conflicts, which can be effectively mitigated through a mechanistic intervention or \textsc{LoRA} training.
More broadly, addressing query-language preference is important for \textit{equitable} multilingual information access. If the information a system surfaces depends on a user's query language, it may confine users to language-bounded evidence, reinforcing filter bubbles \citep{flaxman2016filter} and information cocoons \citep{sunstein2018republic}.
An important next step is to evaluate the real-world impact in deployment, including how source selection affects users' comprehension, trust, and information access. 
Future benchmarks should also extend beyond Wikipedia to multilingual sources with different authority structures and update dynamics, such as academic papers.
We hope that \textsc{Waldo} supports research on robust cross-lingual knowledge transfer and that our mitigation approaches serve as useful tools for studying and mitigating source-selection biases beyond query language, including biases related to formality \citep{li2024formalityfavoredunravelinglearning}, relevance \citep{wan-etal-2024-evidence}, recency \citep{recency_2025}, and source origin \citep{schuster-etal-2026-whose}.


\clearpage
\subsection*{AI use statement}

In this work, we used generative AI tools to generate synthetic datasets including translation, implement methods, propose or refine hypotheses, provide feedback on the methodology, and conduct qualitative analyses.
We have not used generative AI tools to develop theoretical models or conceptual frameworks, formulate or prove mathematical claims, design experiments, or interpret results.
Additionally, we used generative AI tools to formulate questions for surveys, modify figures, propose the name of our benchmark (\textsc{Waldo}), and assist with writing and presentation, including editing the manuscript to improve clarity and readability; making stylistic adjustments; and resolving layout issues such as widows and orphans.
The authors independently reviewed all AI-generated hypotheses, feedback, and qualitative analyses and verified the LLM-generated code for correctness.
We take responsibility for the final content of this work, including text, claims or artifacts produced with the aid of generative AI.

\subsection*{Ethics statement}

Research on query-language preference in multilingual systems raises fairness concerns, particularly for speakers of low-resource and underrepresented languages. When models provide different answers to semantically equivalent queries solely because they are expressed in different languages, they may reinforce existing disparities in access to reliable information. 
Our work aims to identify, measure, and mitigate such disparities rather than reinforce them. Nevertheless, the proposed mitigation methods should be evaluated carefully across languages and domains before deployment, as reducing one form of language preference may have uneven effects on answer quality, source faithfulness, or linguistic and cultural coverage.


\subsection*{Reproducibility statement}

We provide the information necessary to reproduce our data construction, measurement, and mitigation experiments.
We document the construction of \textsc{Waldo} in Section~\ref{sec:dataset} and provide additional implementation details in Appendix~\ref{appendix:waldo_details}.
We specify our evaluation metrics, statistical tests, and experimental settings in Section~\ref{sec:setup} and Appendix~\ref{appendix:implementation}.
We will release both the full and evaluation set of \textsc{Waldo}, code, experiment configurations, and model checkpoints required to reproduce our measurement pipeline, mitigation experiments, and reported analyses.

\subsection*{Acknowledgments}

We thank the Microsoft OneDrive and SharePoint (ODSP) Applied Sciences team for their constructive feedback and support throughout the summer, which helped shape this work.
We are also grateful to Nebojsa Jojic for his assistance in setting up a study for data collection.
Dayeon also extends special thanks to the friends who made the internship experience in Redmond memorable: Soyeong Jeong, Mouly Dewan, Kostas Triaridis, Teo Kitanovski, and Yoonseon Choi. 
She also thanks the interns with whom she discussed the project outside the ODSP team, including, but not limited to, Nikhil Sharma and Xinyan Velocity Yu.
Finally, we thank the members of the \textsc{CLIP} lab at the University of Maryland for their valuable feedback during paper review sessions, especially Kartik Ravisankar, Ramya Namuduri, and Minjoon Choi for their comments on an earlier draft.

\bibliography{iclr2027_conference}

@inproceedings{park-etal-2026-enhancing,
    title = "Enhancing Multilingual {RAG} Systems with Debiased Language Preference-Guided Query Fusion",
    author = "Park, Jeonghyun  and
      Kim, Byeongjeong  and
      Hwang, Seojin  and
      Lee, Hwanhee",
    editor = "Liakata, Maria  and
      Moreira, Viviane P.  and
      Zhang, Jiajun  and
      Jurgens, David",
    booktitle = "Findings of the {A}ssociation for {C}omputational {L}inguistics: {ACL} 2026",
    month = jul,
    year = "2026",
    address = "San Diego, California, United States",
    publisher = "Association for Computational Linguistics",
    url = "https://aclanthology.org/2026.findings-acl.1353/",
    doi = "10.18653/v1/2026.findings-acl.1353",
    pages = "27116--27136",
    ISBN = "979-8-89176-395-1",
}

@inproceedings{
ki2026linguistic,
title={Linguistic Nepotism: Trading-off Quality for Language Preference in Multilingual {RAG}},
author={Dayeon Ki and Marine Carpuat and Paul McNamee and Daniel Khashabi and Eugene Yang and Dawn Lawrie and Kevin Duh},
booktitle={Forty-third International Conference on Machine Learning},
year={2026},
url={https://openreview.net/forum?id=MDfCV0UHSb}
}

@inproceedings{park-lee-2025-investigating,
    title = "Investigating Language Preference of Multilingual {RAG} Systems",
    author = "Park, Jeonghyun  and
      Lee, Hwanhee",
    editor = "Che, Wanxiang  and
      Nabende, Joyce  and
      Shutova, Ekaterina  and
      Pilehvar, Mohammad Taher",
    booktitle = "Findings of the Association for Computational Linguistics: ACL 2025",
    month = jul,
    year = "2025",
    address = "Vienna, Austria",
    publisher = "Association for Computational Linguistics",
    url = "https://aclanthology.org/2025.findings-acl.295/",
    doi = "10.18653/v1/2025.findings-acl.295",
    pages = "5647--5675",
    ISBN = "979-8-89176-256-5",
}

@inproceedings{qi-etal-2026-language,
    title = "Language-Coupled Reinforcement Learning for Multilingual Retrieval-Augmented Generation",
    author = "Qi, Rui  and
      Mo, Fengran  and
      Chen, Yufeng  and
      Zhang, Xue  and
      Wang, Shuo  and
      Li, Hongliang  and
      Jinan, Xu  and
      Jiang, Meng  and
      Nie, Jian-Yun  and
      Huang, Kaiyu",
    editor = "Liakata, Maria  and
      Moreira, Viviane P.  and
      Zhang, Jiajun  and
      Jurgens, David",
    booktitle = "Findings of the {A}ssociation for {C}omputational {L}inguistics: {ACL} 2026",
    month = jul,
    year = "2026",
    address = "San Diego, California, United States",
    publisher = "Association for Computational Linguistics",
    url = "https://aclanthology.org/2026.findings-acl.700/",
    doi = "10.18653/v1/2026.findings-acl.700",
    pages = "14291--14310",
    ISBN = "979-8-89176-395-1",
}

@inproceedings{ranaldi2026multilingual,
  title={Multilingual retrieval-augmented generation for knowledge-intensive question answering task},
  author={Ranaldi, Leonardo and Haddow, Barry and Birch, Alexandra},
  booktitle={Findings of the Association for Computational Linguistics: EACL 2026},
  pages={697--716},
  year={2026}
}

@misc{hupkes2025multilokomultilinguallocalknowledge,
      title={MultiLoKo: a multilingual local knowledge benchmark for LLMs spanning 31 languages}, 
      author={Dieuwke Hupkes and Nikolay Bogoychev},
      year={2025},
      eprint={2504.10356},
      archivePrefix={arXiv},
      primaryClass={cs.CL},
      url={https://arxiv.org/abs/2504.10356}, 
}

@inproceedings{wan-etal-2024-evidence,
    title = "What Evidence Do Language Models Find Convincing?",
    author = "Wan, Alexander  and
      Wallace, Eric  and
      Klein, Dan",
    editor = "Ku, Lun-Wei  and
      Martins, Andre  and
      Srikumar, Vivek",
    booktitle = "Proceedings of the 62nd Annual Meeting of the Association for Computational Linguistics (Volume 1: Long Papers)",
    month = aug,
    year = "2024",
    address = "Bangkok, Thailand",
    publisher = "Association for Computational Linguistics",
    url = "https://aclanthology.org/2024.acl-long.403/",
    doi = "10.18653/v1/2024.acl-long.403",
    pages = "7468--7484",
}

@inproceedings{burchell-etal-2025-expanded,
    title = "An Expanded Massive Multilingual Dataset for High-Performance Language Technologies ({HPLT})",
    author = {Burchell, Laurie  and
      de Gibert, Ona  and
      Arefyev, Nikolay  and
      Aulamo, Mikko  and
      Ba{\~n}{\'o}n, Marta  and
      Chen, Pinzhen  and
      Fedorova, Mariia  and
      Guillou, Liane  and
      Haddow, Barry  and
      Haji{\v{c}}, Jan  and
      Helcl, Jind{\v{r}}ich  and
      Henriksson, Erik  and
      Klimaszewski, Mateusz  and
      Komulainen, Ville  and
      Kutuzov, Andrey  and
      Kyt{\"o}niemi, Joona  and
      Laippala, Veronika  and
      M{\ae}hlum, Petter  and
      Malik, Bhavitvya  and
      Mehryary, Farrokh  and
      Mikhailov, Vladislav  and
      Moghe, Nikita  and
      Myntti, Amanda  and
      O{'}Brien, Dayy{\'a}n  and
      Oepen, Stephan  and
      Pal, Proyag  and
      Piha, Jousia  and
      Pyysalo, Sampo  and
      Ram{\'i}rez-S{\'a}nchez, Gema  and
      Samuel, David  and
      Stepachev, Pavel  and
      Tiedemann, J{\"o}rg  and
      Vari{\v{s}}, Du{\v{s}}an  and
      Vojt{\v{e}}chov{\'a}, Tereza  and
      Zaragoza-Bernabeu, Jaume},
    editor = "Che, Wanxiang  and
      Nabende, Joyce  and
      Shutova, Ekaterina  and
      Pilehvar, Mohammad Taher",
    booktitle = "Proceedings of the 63rd Annual Meeting of the Association for Computational Linguistics (Volume 1: Long Papers)",
    month = jul,
    year = "2025",
    address = "Vienna, Austria",
    publisher = "Association for Computational Linguistics",
    url = "https://aclanthology.org/2025.acl-long.854/",
    doi = "10.18653/v1/2025.acl-long.854",
    pages = "17452--17485",
    ISBN = "979-8-89176-251-0",
}

@inproceedings{wang-etal-2026-languages,
    title = "All Languages Matter: Understanding and Mitigating Language Bias in Multilingual {RAG}",
    author = "Wang, Dan  and
      Mo, Guozhao  and
      Shi, Yafei  and
      Zhang, Cheng  and
      Zheng, Bo  and
      Cao, Boxi  and
      Chen, Xuanang  and
      Lu, Yaojie  and
      Lin, Hongyu  and
      He, Ben  and
      Han, Xianpei  and
      Sun, Le",
    editor = "Liakata, Maria  and
      Moreira, Viviane P.  and
      Zhang, Jiajun  and
      Jurgens, David",
    booktitle = "Proceedings of the 64th Annual Meeting of the {A}ssociation for {C}omputational {L}inguistics (Volume 1: Long Papers)",
    month = jul,
    year = "2026",
    address = "San Diego, California, United States",
    publisher = "Association for Computational Linguistics",
    url = "https://aclanthology.org/2026.acl-long.338/",
    doi = "10.18653/v1/2026.acl-long.338",
    pages = "7441--7455",
    ISBN = "979-8-89176-390-6",
}

@inproceedings{jin2024better,
  title={Better to ask in english: Cross-lingual evaluation of large language models for healthcare queries},
  author={Jin, Yiqiao and Chandra, Mohit and Verma, Gaurav and Hu, Yibo and De Choudhury, Munmun and Kumar, Srijan},
  booktitle={Proceedings of the ACM Web Conference 2024},
  pages={2627--2638},
  year={2024}
}

@article{arguedas2023news,
  title={News for the powerful and privileged: How misrepresentation and underrepresentation of disadvantaged communities undermine their trust in news},
  author={Arguedas, Amy Ross and Banerjee, Sayan and Mont'Alverne, Camila and Toff, Benjamin and Fletcher, Richard and Nielsen, Rasmus Kleis},
  year={2023},
  publisher={Reuters Institute for the Study of Journalism}
}

@book{phillipson1992linguistic,
  title={Linguistic imperialism},
  author={Phillipson, Robert},
  year={1992},
  publisher={oxford Un. Press}
}

@inproceedings{sharma-etal-2025-faux,
    title = "Faux Polyglot: A Study on Information Disparity in Multilingual Large Language Models",
    author = "Sharma, Nikhil  and
      Murray, Kenton  and
      Xiao, Ziang",
    editor = "Chiruzzo, Luis  and
      Ritter, Alan  and
      Wang, Lu",
    booktitle = "Proceedings of the 2025 Conference of the Nations of the Americas Chapter of the Association for Computational Linguistics: Human Language Technologies (Volume 1: Long Papers)",
    month = apr,
    year = "2025",
    address = "Albuquerque, New Mexico",
    publisher = "Association for Computational Linguistics",
    url = "https://aclanthology.org/2025.naacl-long.411/",
    doi = "10.18653/v1/2025.naacl-long.411",
    pages = "8090--8107",
    ISBN = "979-8-89176-189-6",
}

@misc{wu2024languagesequalinsightsmultilingual,
      title={Not All Languages are Equal: Insights into Multilingual Retrieval-Augmented Generation}, 
      author={Suhang Wu and Jialong Tang and Baosong Yang and Ante Wang and Kaidi Jia and Jiawei Yu and Junfeng Yao and Jinsong Su},
      year={2024},
      eprint={2410.21970},
      archivePrefix={arXiv},
      primaryClass={cs.CL},
      url={https://arxiv.org/abs/2410.21970}, 
}

@inproceedings{chirkova-etal-2024-retrieval,
    title = "Retrieval-augmented generation in multilingual settings",
    author = "Chirkova, Nadezhda  and
      Rau, David  and
      D{\'e}jean, Herv{\'e}  and
      Formal, Thibault  and
      Clinchant, St{\'e}phane  and
      Nikoulina, Vassilina",
    editor = "Li, Sha  and
      Li, Manling  and
      Zhang, Michael JQ  and
      Choi, Eunsol  and
      Geva, Mor  and
      Hase, Peter  and
      Ji, Heng",
    booktitle = "Proceedings of the 1st Workshop on Towards Knowledgeable Language Models (KnowLLM 2024)",
    month = aug,
    year = "2024",
    address = "Bangkok, Thailand",
    publisher = "Association for Computational Linguistics",
    url = "https://aclanthology.org/2024.knowllm-1.15/",
    doi = "10.18653/v1/2024.knowllm-1.15",
    pages = "177--188",
}

@inproceedings{huo-etal-2026-breaking,
    title = "Breaking Language Preference in Multilingual {RAG} via Language-Controllable Retrieval and Language-Agnostic Reasoning",
    author = "Huo, Wenshuai  and
      Feng, Xiaocheng  and
      Li, Baohang  and
      Fu, Chengpeng  and
      Huang, Yichong  and
      Wang, Hui  and
      Qin, Bing",
    editor = "Liakata, Maria  and
      Moreira, Viviane P.  and
      Zhang, Jiajun  and
      Jurgens, David",
    booktitle = "Findings of the {A}ssociation for {C}omputational {L}inguistics: {ACL} 2026",
    month = jul,
    year = "2026",
    address = "San Diego, California, United States",
    publisher = "Association for Computational Linguistics",
    url = "https://aclanthology.org/2026.findings-acl.374/",
    doi = "10.18653/v1/2026.findings-acl.374",
    pages = "7579--7589",
    ISBN = "979-8-89176-395-1",
}

@inproceedings{li-etal-2024-land,
    title = "This Land is {Your, My} Land: Evaluating Geopolitical Bias in Language Models through Territorial Disputes",
    author = "Li, Bryan  and
      Haider, Samar  and
      Callison-Burch, Chris",
    editor = "Duh, Kevin  and
      Gomez, Helena  and
      Bethard, Steven",
    booktitle = "Proceedings of the 2024 Conference of the North American Chapter of the Association for Computational Linguistics: Human Language Technologies (Volume 1: Long Papers)",
    month = jun,
    year = "2024",
    address = "Mexico City, Mexico",
    publisher = "Association for Computational Linguistics",
    url = "https://aclanthology.org/2024.naacl-long.213/",
    doi = "10.18653/v1/2024.naacl-long.213",
    pages = "3855--3871",
}

@article{longpre-etal-2021-mkqa,
    title = "{MKQA}: A Linguistically Diverse Benchmark for Multilingual Open Domain Question Answering",
    author = "Longpre, Shayne  and
      Lu, Yi  and
      Daiber, Joachim",
    editor = "Roark, Brian  and
      Nenkova, Ani",
    journal = "Transactions of the Association for Computational Linguistics",
    volume = "9",
    year = "2021",
    address = "Cambridge, MA",
    publisher = "MIT Press",
    url = "https://aclanthology.org/2021.tacl-1.82/",
    doi = "10.1162/tacl_a_00433",
    pages = "1389--1406",
}

@misc{grattafiori2024llama3herdmodels,
      title={{The Llama 3 Herd of Models}}, 
      author={Aaron Grattafiori and Abhimanyu Dubey and Abhinav Jauhri and Abhinav Pandey and Abhishek Kadian and Ahmad Al-Dahle and Aiesha Letman and Akhil Mathur and Alan Schelten and Alex Vaughan and Amy Yang and Angela Fan and Anirudh Goyal and Anthony Hartshorn and Aobo Yang and Archi Mitra and Archie Sravankumar and Artem Korenev and Arthur Hinsvark and Arun Rao and Aston Zhang and Aurelien Rodriguez and Austen Gregerson and Ava Spataru and Baptiste Roziere and Bethany Biron and Binh Tang and Bobbie Chern and Charlotte Caucheteux and Chaya Nayak and Chloe Bi and Chris Marra and Chris McConnell and Christian Keller and Christophe Touret and Chunyang Wu and Corinne Wong and Cristian Canton Ferrer and Cyrus Nikolaidis and Damien Allonsius and Daniel Song and Danielle Pintz and Danny Livshits and Danny Wyatt and David Esiobu and Dhruv Choudhary and Dhruv Mahajan and Diego Garcia-Olano and Diego Perino and Dieuwke Hupkes and Egor Lakomkin and Ehab AlBadawy and Elina Lobanova and Emily Dinan and Eric Michael Smith and Filip Radenovic and Francisco Guzmán and Frank Zhang and Gabriel Synnaeve and Gabrielle Lee and Georgia Lewis Anderson and Govind Thattai and Graeme Nail and Gregoire Mialon and Guan Pang and Guillem Cucurell and Hailey Nguyen and Hannah Korevaar and Hu Xu and Hugo Touvron and Iliyan Zarov and Imanol Arrieta Ibarra and Isabel Kloumann and Ishan Misra and Ivan Evtimov and Jack Zhang and Jade Copet and Jaewon Lee and Jan Geffert and Jana Vranes and Jason Park and Jay Mahadeokar and Jeet Shah and Jelmer van der Linde and Jennifer Billock and Jenny Hong and Jenya Lee and Jeremy Fu and Jianfeng Chi and Jianyu Huang and Jiawen Liu and Jie Wang and Jiecao Yu and Joanna Bitton and Joe Spisak and Jongsoo Park and Joseph Rocca and Joshua Johnstun and Joshua Saxe and Junteng Jia and Kalyan Vasuden Alwala and Karthik Prasad and Kartikeya Upasani and Kate Plawiak and Ke Li and Kenneth Heafield and Kevin Stone and Khalid El-Arini and Krithika Iyer and Kshitiz Malik and Kuenley Chiu and Kunal Bhalla and Kushal Lakhotia and Lauren Rantala-Yeary and Laurens van der Maaten and Lawrence Chen and Liang Tan and Liz Jenkins and Louis Martin and Lovish Madaan and Lubo Malo and Lukas Blecher and Lukas Landzaat and Luke de Oliveira and Madeline Muzzi and Mahesh Pasupuleti and Mannat Singh and Manohar Paluri and Marcin Kardas and Maria Tsimpoukelli and Mathew Oldham and Mathieu Rita and Maya Pavlova and Melanie Kambadur and Mike Lewis and Min Si and Mitesh Kumar Singh and Mona Hassan and Naman Goyal and Narjes Torabi and Nikolay Bashlykov and Nikolay Bogoychev and Niladri Chatterji and Ning Zhang and Olivier Duchenne and Onur Çelebi and Patrick Alrassy and Pengchuan Zhang and Pengwei Li and Petar Vasic and Peter Weng and Prajjwal Bhargava and Pratik Dubal and Praveen Krishnan and Punit Singh Koura and Puxin Xu and Qing He and Qingxiao Dong and Ragavan Srinivasan and Raj Ganapathy and Ramon Calderer and Ricardo Silveira Cabral and Robert Stojnic and Roberta Raileanu and Rohan Maheswari and Rohit Girdhar and Rohit Patel and Romain Sauvestre and Ronnie Polidoro and Roshan Sumbaly and Ross Taylor and Ruan Silva and Rui Hou and Rui Wang and Saghar Hosseini and Sahana Chennabasappa and Sanjay Singh and Sean Bell and Seohyun Sonia Kim and Sergey Edunov and Shaoliang Nie and Sharan Narang and Sharath Raparthy and Sheng Shen and Shengye Wan and Shruti Bhosale and Shun Zhang and Simon Vandenhende and Soumya Batra and Spencer Whitman and Sten Sootla and Stephane Collot and Suchin Gururangan and Sydney Borodinsky and Tamar Herman and Tara Fowler and Tarek Sheasha and Thomas Georgiou and Thomas Scialom and Tobias Speckbacher and Todor Mihaylov and Tong Xiao and Ujjwal Karn and Vedanuj Goswami and Vibhor Gupta and Vignesh Ramanathan and Viktor Kerkez and Vincent Gonguet and Virginie Do and Vish Vogeti and Vítor Albiero and Vladan Petrovic and Weiwei Chu and Wenhan Xiong and Wenyin Fu and Whitney Meers and Xavier Martinet and Xiaodong Wang and Xiaofang Wang and Xiaoqing Ellen Tan and Xide Xia and Xinfeng Xie and Xuchao Jia and Xuewei Wang and Yaelle Goldschlag and Yashesh Gaur and Yasmine Babaei and Yi Wen and Yiwen Song and Yuchen Zhang and Yue Li and Yuning Mao and Zacharie Delpierre Coudert and Zheng Yan and Zhengxing Chen and Zoe Papakipos and Aaditya Singh and Aayushi Srivastava and Abha Jain and Adam Kelsey and Adam Shajnfeld and Adithya Gangidi and Adolfo Victoria and Ahuva Goldstand and Ajay Menon and Ajay Sharma and Alex Boesenberg and Alexei Baevski and Allie Feinstein and Amanda Kallet and Amit Sangani and Amos Teo and Anam Yunus and Andrei Lupu and Andres Alvarado and Andrew Caples and Andrew Gu and Andrew Ho and Andrew Poulton and Andrew Ryan and Ankit Ramchandani and Annie Dong and Annie Franco and Anuj Goyal and Aparajita Saraf and Arkabandhu Chowdhury and Ashley Gabriel and Ashwin Bharambe and Assaf Eisenman and Azadeh Yazdan and Beau James and Ben Maurer and Benjamin Leonhardi and Bernie Huang and Beth Loyd and Beto De Paola and Bhargavi Paranjape and Bing Liu and Bo Wu and Boyu Ni and Braden Hancock and Bram Wasti and Brandon Spence and Brani Stojkovic and Brian Gamido and Britt Montalvo and Carl Parker and Carly Burton and Catalina Mejia and Ce Liu and Changhan Wang and Changkyu Kim and Chao Zhou and Chester Hu and Ching-Hsiang Chu and Chris Cai and Chris Tindal and Christoph Feichtenhofer and Cynthia Gao and Damon Civin and Dana Beaty and Daniel Kreymer and Daniel Li and David Adkins and David Xu and Davide Testuggine and Delia David and Devi Parikh and Diana Liskovich and Didem Foss and Dingkang Wang and Duc Le and Dustin Holland and Edward Dowling and Eissa Jamil and Elaine Montgomery and Eleonora Presani and Emily Hahn and Emily Wood and Eric-Tuan Le and Erik Brinkman and Esteban Arcaute and Evan Dunbar and Evan Smothers and Fei Sun and Felix Kreuk and Feng Tian and Filippos Kokkinos and Firat Ozgenel and Francesco Caggioni and Frank Kanayet and Frank Seide and Gabriela Medina Florez and Gabriella Schwarz and Gada Badeer and Georgia Swee and Gil Halpern and Grant Herman and Grigory Sizov and Guangyi and Zhang and Guna Lakshminarayanan and Hakan Inan and Hamid Shojanazeri and Han Zou and Hannah Wang and Hanwen Zha and Haroun Habeeb and Harrison Rudolph and Helen Suk and Henry Aspegren and Hunter Goldman and Hongyuan Zhan and Ibrahim Damlaj and Igor Molybog and Igor Tufanov and Ilias Leontiadis and Irina-Elena Veliche and Itai Gat and Jake Weissman and James Geboski and James Kohli and Janice Lam and Japhet Asher and Jean-Baptiste Gaya and Jeff Marcus and Jeff Tang and Jennifer Chan and Jenny Zhen and Jeremy Reizenstein and Jeremy Teboul and Jessica Zhong and Jian Jin and Jingyi Yang and Joe Cummings and Jon Carvill and Jon Shepard and Jonathan McPhie and Jonathan Torres and Josh Ginsburg and Junjie Wang and Kai Wu and Kam Hou U and Karan Saxena and Kartikay Khandelwal and Katayoun Zand and Kathy Matosich and Kaushik Veeraraghavan and Kelly Michelena and Keqian Li and Kiran Jagadeesh and Kun Huang and Kunal Chawla and Kyle Huang and Lailin Chen and Lakshya Garg and Lavender A and Leandro Silva and Lee Bell and Lei Zhang and Liangpeng Guo and Licheng Yu and Liron Moshkovich and Luca Wehrstedt and Madian Khabsa and Manav Avalani and Manish Bhatt and Martynas Mankus and Matan Hasson and Matthew Lennie and Matthias Reso and Maxim Groshev and Maxim Naumov and Maya Lathi and Meghan Keneally and Miao Liu and Michael L. Seltzer and Michal Valko and Michelle Restrepo and Mihir Patel and Mik Vyatskov and Mikayel Samvelyan and Mike Clark and Mike Macey and Mike Wang and Miquel Jubert Hermoso and Mo Metanat and Mohammad Rastegari and Munish Bansal and Nandhini Santhanam and Natascha Parks and Natasha White and Navyata Bawa and Nayan Singhal and Nick Egebo and Nicolas Usunier and Nikhil Mehta and Nikolay Pavlovich Laptev and Ning Dong and Norman Cheng and Oleg Chernoguz and Olivia Hart and Omkar Salpekar and Ozlem Kalinli and Parkin Kent and Parth Parekh and Paul Saab and Pavan Balaji and Pedro Rittner and Philip Bontrager and Pierre Roux and Piotr Dollar and Polina Zvyagina and Prashant Ratanchandani and Pritish Yuvraj and Qian Liang and Rachad Alao and Rachel Rodriguez and Rafi Ayub and Raghotham Murthy and Raghu Nayani and Rahul Mitra and Rangaprabhu Parthasarathy and Raymond Li and Rebekkah Hogan and Robin Battey and Rocky Wang and Russ Howes and Ruty Rinott and Sachin Mehta and Sachin Siby and Sai Jayesh Bondu and Samyak Datta and Sara Chugh and Sara Hunt and Sargun Dhillon and Sasha Sidorov and Satadru Pan and Saurabh Mahajan and Saurabh Verma and Seiji Yamamoto and Sharadh Ramaswamy and Shaun Lindsay and Shaun Lindsay and Sheng Feng and Shenghao Lin and Shengxin Cindy Zha and Shishir Patil and Shiva Shankar and Shuqiang Zhang and Shuqiang Zhang and Sinong Wang and Sneha Agarwal and Soji Sajuyigbe and Soumith Chintala and Stephanie Max and Stephen Chen and Steve Kehoe and Steve Satterfield and Sudarshan Govindaprasad and Sumit Gupta and Summer Deng and Sungmin Cho and Sunny Virk and Suraj Subramanian and Sy Choudhury and Sydney Goldman and Tal Remez and Tamar Glaser and Tamara Best and Thilo Koehler and Thomas Robinson and Tianhe Li and Tianjun Zhang and Tim Matthews and Timothy Chou and Tzook Shaked and Varun Vontimitta and Victoria Ajayi and Victoria Montanez and Vijai Mohan and Vinay Satish Kumar and Vishal Mangla and Vlad Ionescu and Vlad Poenaru and Vlad Tiberiu Mihailescu and Vladimir Ivanov and Wei Li and Wenchen Wang and Wenwen Jiang and Wes Bouaziz and Will Constable and Xiaocheng Tang and Xiaojian Wu and Xiaolan Wang and Xilun Wu and Xinbo Gao and Yaniv Kleinman and Yanjun Chen and Ye Hu and Ye Jia and Ye Qi and Yenda Li and Yilin Zhang and Ying Zhang and Yossi Adi and Youngjin Nam and Yu and Wang and Yu Zhao and Yuchen Hao and Yundi Qian and Yunlu Li and Yuzi He and Zach Rait and Zachary DeVito and Zef Rosnbrick and Zhaoduo Wen and Zhenyu Yang and Zhiwei Zhao and Zhiyu Ma},
      year={2024},
      eprint={2407.21783},
      archivePrefix={arXiv},
      primaryClass={cs.AI},
      url={https://arxiv.org/abs/2407.21783}, 
}

@misc{yang2025qwen3technicalreport,
      title={{Qwen3 Technical Report}}, 
      author={An Yang and Anfeng Li and Baosong Yang and Beichen Zhang and Binyuan Hui and Bo Zheng and Bowen Yu and Chang Gao and Chengen Huang and Chenxu Lv and Chujie Zheng and Dayiheng Liu and Fan Zhou and Fei Huang and Feng Hu and Hao Ge and Haoran Wei and Huan Lin and Jialong Tang and Jian Yang and Jianhong Tu and Jianwei Zhang and Jianxin Yang and Jiaxi Yang and Jing Zhou and Jingren Zhou and Junyang Lin and Kai Dang and Keqin Bao and Kexin Yang and Le Yu and Lianghao Deng and Mei Li and Mingfeng Xue and Mingze Li and Pei Zhang and Peng Wang and Qin Zhu and Rui Men and Ruize Gao and Shixuan Liu and Shuang Luo and Tianhao Li and Tianyi Tang and Wenbiao Yin and Xingzhang Ren and Xinyu Wang and Xinyu Zhang and Xuancheng Ren and Yang Fan and Yang Su and Yichang Zhang and Yinger Zhang and Yu Wan and Yuqiong Liu and Zekun Wang and Zeyu Cui and Zhenru Zhang and Zhipeng Zhou and Zihan Qiu},
      year={2025},
      eprint={2505.09388},
      archivePrefix={arXiv},
      primaryClass={cs.CL},
      url={https://arxiv.org/abs/2505.09388}, 
}

@misc{openai2026gpt55,
  author       = {{OpenAI}},
  title        = {Introducing {GPT-5.5}},
  year         = {2026},
  month        = apr,
  day          = {23},
  howpublished = {\url{https://openai.com/index/introducing-gpt-5-5/}},
  note         = {Accessed: 2026-09-04}
}

@misc{dang2024ayaexpansecombiningresearch,
      title={Aya Expanse: Combining Research Breakthroughs for a New Multilingual Frontier}, 
      author={John Dang and Shivalika Singh and Daniel D'souza and Arash Ahmadian and Alejandro Salamanca and Madeline Smith and Aidan Peppin and Sungjin Hong and Manoj Govindassamy and Terrence Zhao and Sandra Kublik and Meor Amer and Viraat Aryabumi and Jon Ander Campos and Yi-Chern Tan and Tom Kocmi and Florian Strub and Nathan Grinsztajn and Yannis Flet-Berliac and Acyr Locatelli and Hangyu Lin and Dwarak Talupuru and Bharat Venkitesh and David Cairuz and Bowen Yang and Tim Chung and Wei-Yin Ko and Sylvie Shang Shi and Amir Shukayev and Sammie Bae and Aleksandra Piktus and Roman Castagné and Felipe Cruz-Salinas and Eddie Kim and Lucas Crawhall-Stein and Adrien Morisot and Sudip Roy and Phil Blunsom and Ivan Zhang and Aidan Gomez and Nick Frosst and Marzieh Fadaee and Beyza Ermis and Ahmet Üstün and Sara Hooker},
      year={2024},
      eprint={2412.04261},
      archivePrefix={arXiv},
      primaryClass={cs.CL},
      url={https://arxiv.org/abs/2412.04261}, 
}

@misc{deepseekai2026deepseekv4highlyefficientmilliontoken,
      title={DeepSeek-V4: Towards Highly Efficient Million-Token Context Intelligence}, 
      author={DeepSeek-AI and Anyi Xu and Bangcai Lin and Bing Xue and Bingxuan Wang and Bingzheng Xu and Bochao Wu and Bowei Zhang and Chaofan Lin and Chen Dong and Chenchen Ling and Chengda Lu and Chenggang Zhao and Chengqi Deng and Chengyu Hou and Chenhao Xu and Chenze Shao and Chong Ruan and Conner Sun and Damai Dai and Daya Guo and Dejian Yang and Deli Chen and Donghao Li and Dongjie Ji and Erhang Li and Fang Wei and Fangyun Lin and Fangzhou Yuan and Feiyu Xia and Fucong Dai and Guangbo Hao and Guanting Chen and Guoai Cao and Guolai Meng and Guowei Li and Han Yu and Han Zhang and Hanwei Xu and Hao Li and Haofen Liang and Haoling Zhang and Haoming Luo and Haoran Wei and Haotian Yuan and Haowei Zhang and Haowen Luo and Haoyu Chen and Haozhe Ji and Hengqing Zhang and Honghui Ding and Hongxuan Tang and Huanqi Cao and Huazuo Gao and Hui Qu and Hui Zeng and J Yang and JQ Zhu and Jia Luo and Jia Song and Jia Yu and Jialiang Huang and Jialu Cai and Jian Liang and Jiangting Zhou and Jiasheng Ye and Jiashi Li and Jiaxin Xu and Jiewen Hu and Jieyu Yang and Jin Chen and Jin Yan and Jingchang Chen and Jingli Zhou and Jingting Xiang and Jingyang Yuan and Jingyuan Cheng and Jingzi Zhou and Jinhua Zhu and Jiping Yu and Joseph Sun and Jun Ran and Junguang Jiang and Junjie Qiu and Junlong Li and Junmin Zheng and Junxiao Song and Kai Dong and Kaige Gao and Kang Guan and Kexing Zhou and Kezhao Huang and Kuai Yu and Lean Wang and Lecong Zhang and Lei Wang and Leyi Xia and Li Zhang and Liang Zhao and Lihua Guo and Lingxiao Luo and Linwang Ma and Linyan Zhu and Litong Wang and Liyu Cai and Liyue Zhang and Longhao Chen and MS Di and MY Xu and Max Mei and Miaojun Wang and Mingchuan Zhang and Minghua Zhang and Minghui Tang and Mingming Li and Mingxu Zhou and Minmin Han and Ning Wang and Panpan Huang and Panpan Wang and Peixin Cong and Peiyi Wang and Peng Zhang and Qiancheng Wang and Qihao Zhu and Qingyang Li and Qinyu Chen and Qiushi Du and Qiwei Jiang and Rui Tian and Ruifan Xu and Ruijie Lu and Ruiling Xu and Ruiqi Ge and Ruisong Zhang and Ruizhe Pan and Runji Wang and Runqian Chen and Runqiu Yin and Runxin Xu and Ruomeng Shen and Ruoyu Zhang and Ruyi Chen and SH Liu and Shanghao Lu and Shangmian Sun and Shangyan Zhou and Shanhuang Chen and Shaofei Cai and Shaoheng Nie and Shaoqing Wu and Shaoyuan Chen and Shengding Hu and Shengyu Liu and Shiqiang Hu and Shirong Ma and Shiyu Wang and Shuiping Yu and Shunfeng Zhou and Shuting Pan and Shuying Yu and Songyang Zhou and Tao Ni and Tao Yun and Tian Jin and Tian Pei and Tian Ye and Tianle Lin and Tianran Ji and Tianyi Cui and Tianyuan Yue and Tingting Yu and Tun Wang and W Zhang and WL Xiao and Wangding Zeng and Wei An and Weilin Zhao and Wen Liu and Wenfeng Liang and Wenjie Pang and Wenjing Luo and Wenjing Yao and Wenjun Gao and Wenkai Yang and Wenlve Huang and Wenqing Hou and Wentao Zhang and Wenting Ma and Xi Gao and Xiang He and Xiangwen Wang and Xianzu Wang and Xiao Bi and Xiaodong Liu and Xiaohan Wang and Xiaokang Chen and Xiaokang Zhang and Xiaotao Nie and Xiaowen Sun and Xiaoxiang Wang and Xin Cheng and Xin Liu and Xin Xie and Xingchao Liu and Xingchen Liu and Xingkai Yu and Xingyou Li and Xinyu Yang and Xinyu Zhang and Xu Chen and Xuanyu Wang and Xuecheng Su and Xueyin Chen and Xuheng Lin and Xuwei Fu and YC Yan and YQ Wang and YW Ma and Yanfeng Luo and Yang Zhang and Yanhong Xu and Yanru Ma and Yanwen Huang and Yao Li and Yao Li and Yao Xu and Yao Zhao and Yaofeng Sun and Yaohui Wang and Yi Qian and Yi Shao and Yi Yu and Yichao Zhang and Yifan Ding and Yifan Shi and Yijia Wu and Yiliang Xiong and Yiling Ma and Ying He and Ying Tang and Ying Zhou and Yingjia Luo and Yinmin Zhong and Yishi Piao and Yisong Wang and Yixiang Zhang and Yixiao Chen and Yixuan Tan and Yixuan Wei and Yiyang Ma and Yiyuan Liu and Yonglun Yang and Yongqiang Guo and Yongtong Wu and Yu Wu and YuKun Li and Yuan Cheng and Yuan Ou and Yuanfan Xu and Yuanhao Li and Yuduan Wang and Yuehan Yang and Yuer Xu and Yuhan Wu and Yuhao Meng and Yuheng Zou and Yukun Zha and Yunfan Xiong and Yupeng Chen and Yuping Lin and Yuqian Cao and Yuqian Wang and Yushun Zhang and Yuting Yan and Yutong Lin and Yuxian Gu and Yuxiang Luo and Yuxiang You and Yuxuan Liu and Yuxuan Zhou and Yuyang Zhou and Yuzhen Huang and ZF Wu and Zehao Wang and Zehua Zhao and Zehui Ren and Zekai Zhang and Zhangli Sha and Zhe Fu and Zhe Ju and Zhean Xu and Zhenda Xie and Zhengyan Zhang and Zheren Gao and Zhewen Hao and Zhibin Gou and Zhicheng Ma and Zhigang Yan and Zhihong Shao and Zhixian Huang and Zhixuan Chen and Zhiyu Wu and Zhizhou Ren and Zhongyu Wu and Zhuoshu Li and Zhuping Zhang and Zian Xu and Zihao Wang and Zihua Qu and Zihui Gu and Zijia Zhu and Zilin Li and Zipeng Zhang and Ziwei Xie and Ziyi Gao and Ziyi Wan and Zizheng Pan and Zongqing Yao},
      year={2026},
      eprint={2606.19348},
      archivePrefix={arXiv},
      primaryClass={cs.CL},
      url={https://arxiv.org/abs/2606.19348}, 
}

@misc{openai2025gpt41,
  author       = {{OpenAI}},
  title        = {Introducing {GPT-4.1} in the {API}},
  year         = {2025},
  month        = apr,
  day          = {14},
  howpublished = {\url{https://openai.com/index/gpt-4-1/}},
  note         = {Accessed: 2026-09-04}
}

@inproceedings{schuster-etal-2026-whose,
    title = "Whose Facts Win? {LLM} Source Preferences under Knowledge Conflicts",
    author = "Schuster, Jakob  and
      Gautam, Vagrant  and
      Markert, Katja",
    editor = "Liakata, Maria  and
      Moreira, Viviane P.  and
      Zhang, Jiajun  and
      Jurgens, David",
    booktitle = "Proceedings of the 64th Annual Meeting of the {A}ssociation for {C}omputational {L}inguistics (Volume 1: Long Papers)",
    month = jul,
    year = "2026",
    address = "San Diego, California, United States",
    publisher = "Association for Computational Linguistics",
    url = "https://aclanthology.org/2026.acl-long.1357/",
    doi = "10.18653/v1/2026.acl-long.1357",
    pages = "29430--29459",
    ISBN = "979-8-89176-390-6",
}

@misc{cattan2025draggedconflictsdetectingaddressing,
      title={DRAGged into Conflicts: Detecting and Addressing Conflicting Sources in Search-Augmented LLMs}, 
      author={Arie Cattan and Alon Jacovi and Ori Ram and Jonathan Herzig and Roee Aharoni and Sasha Goldshtein and Eran Ofek and Idan Szpektor and Avi Caciularu},
      year={2025},
      eprint={2506.08500},
      archivePrefix={arXiv},
      primaryClass={cs.CL},
      url={https://arxiv.org/abs/2506.08500}, 
}

@inproceedings{information_2026,
author = {Nadalic Sotic, Bruno and Kamps, Jaap},
title = {Information Seeking Behavior in LLM-Based RAG: Mental Models and Missing Information},
year = {2026},
isbn = {9798400725999},
publisher = {Association for Computing Machinery},
address = {New York, NY, USA},
url = {https://doi.org/10.1145/3805712.3808537},
doi = {10.1145/3805712.3808537},
booktitle = {Proceedings of the 49th International ACM SIGIR Conference on Research and Development in Information Retrieval},
pages = {2684–2695},
numpages = {12},
location = {Australia},
series = {SIGIR '26}
}

@inproceedings{recency_2025,
author = {Fang, Hanpei and Tao, Sijie and Chen, Nuo and Chang, Kai-Xin and Sakai, Tetsuya},
title = {Do Large Language Models Favor Recent Content? A Study on Recency Bias in LLM-Based Reranking},
year = {2025},
isbn = {9798400722189},
publisher = {Association for Computing Machinery},
address = {New York, NY, USA},
url = {https://doi.org/10.1145/3767695.3769493},
doi = {10.1145/3767695.3769493},
booktitle = {Proceedings of the 2025 Annual International ACM SIGIR Conference on Research and Development in Information Retrieval in the Asia Pacific Region},
pages = {85–94},
numpages = {10},
location = {China},
series = {SIGIR-AP 2025}
}

@inproceedings{qlora,
author = {Dettmers, Tim and Pagnoni, Artidoro and Holtzman, Ari and Zettlemoyer, Luke},
title = {QLORA: efficient finetuning of quantized LLMs},
year = {2023},
publisher = {Curran Associates Inc.},
address = {Red Hook, NY, USA},
booktitle = {Proceedings of the 37th International Conference on Neural Information Processing Systems},
articleno = {441},
numpages = {28},
location = {New Orleans, LA, USA},
series = {NIPS '23}
}

@misc{openai2024gpt4ocard,
      title={GPT-4o System Card}, 
      author={OpenAI and : and Aaron Hurst and Adam Lerer and Adam P. Goucher and Adam Perelman and Aditya Ramesh and Aidan Clark and AJ Ostrow and Akila Welihinda and Alan Hayes and Alec Radford and Aleksander Mądry and Alex Baker-Whitcomb and Alex Beutel and Alex Borzunov and Alex Carney and Alex Chow and Alex Kirillov and Alex Nichol and Alex Paino and Alex Renzin and Alex Tachard Passos and Alexander Kirillov and Alexi Christakis and Alexis Conneau and Ali Kamali and Allan Jabri and Allison Moyer and Allison Tam and Amadou Crookes and Amin Tootoochian and Amin Tootoonchian and Ananya Kumar and Andrea Vallone and Andrej Karpathy and Andrew Braunstein and Andrew Cann and Andrew Codispoti and Andrew Galu and Andrew Kondrich and Andrew Tulloch and Andrey Mishchenko and Angela Baek and Angela Jiang and Antoine Pelisse and Antonia Woodford and Anuj Gosalia and Arka Dhar and Ashley Pantuliano and Avi Nayak and Avital Oliver and Barret Zoph and Behrooz Ghorbani and Ben Leimberger and Ben Rossen and Ben Sokolowsky and Ben Wang and Benjamin Zweig and Beth Hoover and Blake Samic and Bob McGrew and Bobby Spero and Bogo Giertler and Bowen Cheng and Brad Lightcap and Brandon Walkin and Brendan Quinn and Brian Guarraci and Brian Hsu and Bright Kellogg and Brydon Eastman and Camillo Lugaresi and Carroll Wainwright and Cary Bassin and Cary Hudson and Casey Chu and Chad Nelson and Chak Li and Chan Jun Shern and Channing Conger and Charlotte Barette and Chelsea Voss and Chen Ding and Cheng Lu and Chong Zhang and Chris Beaumont and Chris Hallacy and Chris Koch and Christian Gibson and Christina Kim and Christine Choi and Christine McLeavey and Christopher Hesse and Claudia Fischer and Clemens Winter and Coley Czarnecki and Colin Jarvis and Colin Wei and Constantin Koumouzelis and Dane Sherburn and Daniel Kappler and Daniel Levin and Daniel Levy and David Carr and David Farhi and David Mely and David Robinson and David Sasaki and Denny Jin and Dev Valladares and Dimitris Tsipras and Doug Li and Duc Phong Nguyen and Duncan Findlay and Edede Oiwoh and Edmund Wong and Ehsan Asdar and Elizabeth Proehl and Elizabeth Yang and Eric Antonow and Eric Kramer and Eric Peterson and Eric Sigler and Eric Wallace and Eugene Brevdo and Evan Mays and Farzad Khorasani and Felipe Petroski Such and Filippo Raso and Francis Zhang and Fred von Lohmann and Freddie Sulit and Gabriel Goh and Gene Oden and Geoff Salmon and Giulio Starace and Greg Brockman and Hadi Salman and Haiming Bao and Haitang Hu and Hannah Wong and Haoyu Wang and Heather Schmidt and Heather Whitney and Heewoo Jun and Hendrik Kirchner and Henrique Ponde de Oliveira Pinto and Hongyu Ren and Huiwen Chang and Hyung Won Chung and Ian Kivlichan and Ian O'Connell and Ian O'Connell and Ian Osband and Ian Silber and Ian Sohl and Ibrahim Okuyucu and Ikai Lan and Ilya Kostrikov and Ilya Sutskever and Ingmar Kanitscheider and Ishaan Gulrajani and Jacob Coxon and Jacob Menick and Jakub Pachocki and James Aung and James Betker and James Crooks and James Lennon and Jamie Kiros and Jan Leike and Jane Park and Jason Kwon and Jason Phang and Jason Teplitz and Jason Wei and Jason Wolfe and Jay Chen and Jeff Harris and Jenia Varavva and Jessica Gan Lee and Jessica Shieh and Ji Lin and Jiahui Yu and Jiayi Weng and Jie Tang and Jieqi Yu and Joanne Jang and Joaquin Quinonero Candela and Joe Beutler and Joe Landers and Joel Parish and Johannes Heidecke and John Schulman and Jonathan Lachman and Jonathan McKay and Jonathan Uesato and Jonathan Ward and Jong Wook Kim and Joost Huizinga and Jordan Sitkin and Jos Kraaijeveld and Josh Gross and Josh Kaplan and Josh Snyder and Joshua Achiam and Joy Jiao and Joyce Lee and Juntang Zhuang and Justyn Harriman and Kai Fricke and Kai Hayashi and Karan Singhal and Katy Shi and Kavin Karthik and Kayla Wood and Kendra Rimbach and Kenny Hsu and Kenny Nguyen and Keren Gu-Lemberg and Kevin Button and Kevin Liu and Kiel Howe and Krithika Muthukumar and Kyle Luther and Lama Ahmad and Larry Kai and Lauren Itow and Lauren Workman and Leher Pathak and Leo Chen and Li Jing and Lia Guy and Liam Fedus and Liang Zhou and Lien Mamitsuka and Lilian Weng and Lindsay McCallum and Lindsey Held and Long Ouyang and Louis Feuvrier and Lu Zhang and Lukas Kondraciuk and Lukasz Kaiser and Luke Hewitt and Luke Metz and Lyric Doshi and Mada Aflak and Maddie Simens and Madelaine Boyd and Madeleine Thompson and Marat Dukhan and Mark Chen and Mark Gray and Mark Hudnall and Marvin Zhang and Marwan Aljubeh and Mateusz Litwin and Matthew Zeng and Max Johnson and Maya Shetty and Mayank Gupta and Meghan Shah and Mehmet Yatbaz and Meng Jia Yang and Mengchao Zhong and Mia Glaese and Mianna Chen and Michael Janner and Michael Lampe and Michael Petrov and Michael Wu and Michele Wang and Michelle Fradin and Michelle Pokrass and Miguel Castro and Miguel Oom Temudo de Castro and Mikhail Pavlov and Miles Brundage and Miles Wang and Minal Khan and Mira Murati and Mo Bavarian and Molly Lin and Murat Yesildal and Nacho Soto and Natalia Gimelshein and Natalie Cone and Natalie Staudacher and Natalie Summers and Natan LaFontaine and Neil Chowdhury and Nick Ryder and Nick Stathas and Nick Turley and Nik Tezak and Niko Felix and Nithanth Kudige and Nitish Keskar and Noah Deutsch and Noel Bundick and Nora Puckett and Ofir Nachum and Ola Okelola and Oleg Boiko and Oleg Murk and Oliver Jaffe and Olivia Watkins and Olivier Godement and Owen Campbell-Moore and Patrick Chao and Paul McMillan and Pavel Belov and Peng Su and Peter Bak and Peter Bakkum and Peter Deng and Peter Dolan and Peter Hoeschele and Peter Welinder and Phil Tillet and Philip Pronin and Philippe Tillet and Prafulla Dhariwal and Qiming Yuan and Rachel Dias and Rachel Lim and Rahul Arora and Rajan Troll and Randall Lin and Rapha Gontijo Lopes and Raul Puri and Reah Miyara and Reimar Leike and Renaud Gaubert and Reza Zamani and Ricky Wang and Rob Donnelly and Rob Honsby and Rocky Smith and Rohan Sahai and Rohit Ramchandani and Romain Huet and Rory Carmichael and Rowan Zellers and Roy Chen and Ruby Chen and Ruslan Nigmatullin and Ryan Cheu and Saachi Jain and Sam Altman and Sam Schoenholz and Sam Toizer and Samuel Miserendino and Sandhini Agarwal and Sara Culver and Scott Ethersmith and Scott Gray and Sean Grove and Sean Metzger and Shamez Hermani and Shantanu Jain and Shengjia Zhao and Sherwin Wu and Shino Jomoto and Shirong Wu and Shuaiqi and Xia and Sonia Phene and Spencer Papay and Srinivas Narayanan and Steve Coffey and Steve Lee and Stewart Hall and Suchir Balaji and Tal Broda and Tal Stramer and Tao Xu and Tarun Gogineni and Taya Christianson and Ted Sanders and Tejal Patwardhan and Thomas Cunninghman and Thomas Degry and Thomas Dimson and Thomas Raoux and Thomas Shadwell and Tianhao Zheng and Todd Underwood and Todor Markov and Toki Sherbakov and Tom Rubin and Tom Stasi and Tomer Kaftan and Tristan Heywood and Troy Peterson and Tyce Walters and Tyna Eloundou and Valerie Qi and Veit Moeller and Vinnie Monaco and Vishal Kuo and Vlad Fomenko and Wayne Chang and Weiyi Zheng and Wenda Zhou and Wesam Manassra and Will Sheu and Wojciech Zaremba and Yash Patil and Yilei Qian and Yongjik Kim and Youlong Cheng and Yu Zhang and Yuchen He and Yuchen Zhang and Yujia Jin and Yunxing Dai and Yury Malkov},
      year={2024},
      eprint={2410.21276},
      archivePrefix={arXiv},
      primaryClass={cs.CL},
      url={https://arxiv.org/abs/2410.21276}, 
}

@misc{openai2026gpt56,
  author       = {{OpenAI}},
  title        = {{GPT-5.6: Frontier Intelligence That Scales with Your Ambition}},
  year         = {2026},
  howpublished = {\url{https://openai.com/index/gpt-5-6/}},
  note         = {Accessed: 2026-09-05}
}

@misc{fu2026languagesdisagreeselfevolvingmultilingual,
      title={When Languages Disagree: Self-Evolving Multilingual LLM Judges}, 
      author={Xiyan Fu and Wei Lu},
      year={2026},
      eprint={2606.08092},
      archivePrefix={arXiv},
      primaryClass={cs.CL},
      url={https://arxiv.org/abs/2606.08092}, 
}

@inproceedings{filatova2009directions,
  title={Directions for exploiting asymmetries in multilingual Wikipedia},
  author={Filatova, Elena},
  booktitle={Proceedings of the Third International Workshop on Cross Lingual Information Access: Addressing the Information Need of Multilingual Societies (CLIAWS3)},
  pages={30--37},
  year={2009}
}

@inproceedings{joshi2020state,
  title={The state and fate of linguistic diversity and inclusion in the NLP world},
  author={Joshi, Pratik and Santy, Sebastin and Budhiraja, Amar and Bali, Kalika and Choudhury, Monojit},
  booktitle={Proceedings of the 58th annual meeting of the association for computational linguistics},
  pages={6282--6293},
  year={2020}
}

@inproceedings{petrov2023token_unfairness,
    title = {Language Model Tokenizers Introduce Unfairness Between Languages},
    author = {Petrov, Aleksandar and La Malfa, Emanuele and H. S. Torr, Philip and Bibi, Adel},    
    booktitle = {Advances in Neural Information Processing Systems},
    url = {https://arxiv.org/abs/2305.15425},
    year = {2023}
}

@misc{goldman2025eclekticnovelchallengeset,
      title={ECLeKTic: a Novel Challenge Set for Evaluation of Cross-Lingual Knowledge Transfer}, 
      author={Omer Goldman and Uri Shaham and Dan Malkin and Sivan Eiger and Avinatan Hassidim and Yossi Matias and Joshua Maynez and Adi Mayrav Gilady and Jason Riesa and Shruti Rijhwani and Laura Rimell and Idan Szpektor and Reut Tsarfaty and Matan Eyal},
      year={2025},
      eprint={2502.21228},
      archivePrefix={arXiv},
      primaryClass={cs.CL},
      url={https://arxiv.org/abs/2502.21228}, 
}

@inproceedings{echochamber_2024,
author = {Sharma, Nikhil and Liao, Q. Vera and Xiao, Ziang},
title = {Generative Echo Chamber? Effect of LLM-Powered Search Systems on Diverse Information Seeking},
year = {2024},
isbn = {9798400703300},
publisher = {Association for Computing Machinery},
address = {New York, NY, USA},
url = {https://doi.org/10.1145/3613904.3642459},
doi = {10.1145/3613904.3642459},
booktitle = {Proceedings of the 2024 CHI Conference on Human Factors in Computing Systems},
articleno = {1033},
numpages = {17},
location = {Honolulu, HI, USA},
series = {CHI '24}
}

@misc{jang2026mlairemultilinguallanguageawareinformation,
      title={MLAIRE: Multilingual Language-Aware Information Retrieval Evaluation Protocal}, 
      author={Youngjoon Jang and Seongtae Hong and Hyeonseok Moon and Heuiseok Lim},
      year={2026},
      eprint={2605.07249},
      archivePrefix={arXiv},
      primaryClass={cs.IR},
      url={https://arxiv.org/abs/2605.07249}, 
}

@inproceedings{zhang-etal-2025-longcite,
    title = "{L}ong{C}ite: Enabling {LLM}s to Generate Fine-grained Citations in Long-Context {QA}",
    author = "Zhang, Jiajie  and
      Bai, Yushi  and
      Lv, Xin  and
      Gu, Wanjun  and
      Liu, Danqing  and
      Zou, Minhao  and
      Cao, Shulin  and
      Hou, Lei  and
      Dong, Yuxiao  and
      Feng, Ling  and
      Li, Juanzi",
    editor = "Che, Wanxiang  and
      Nabende, Joyce  and
      Shutova, Ekaterina  and
      Pilehvar, Mohammad Taher",
    booktitle = "Findings of the Association for Computational Linguistics: ACL 2025",
    month = jul,
    year = "2025",
    address = "Vienna, Austria",
    publisher = "Association for Computational Linguistics",
    url = "https://aclanthology.org/2025.findings-acl.264/",
    doi = "10.18653/v1/2025.findings-acl.264",
    pages = "5098--5122",
    ISBN = "979-8-89176-256-5",
}

@article{liu-etal-2024-lost,
    title = "Lost in the Middle: How Language Models Use Long Contexts",
    author = "Liu, Nelson F.  and
      Lin, Kevin  and
      Hewitt, John  and
      Paranjape, Ashwin  and
      Bevilacqua, Michele  and
      Petroni, Fabio  and
      Liang, Percy",
    journal = "Transactions of the Association for Computational Linguistics",
    volume = "12",
    year = "2024",
    address = "Cambridge, MA",
    publisher = "MIT Press",
    url = "https://aclanthology.org/2024.tacl-1.9/",
    doi = "10.1162/tacl_a_00638",
    pages = "157--173",
}

@misc{hong2026lamaropenlanguageawaremultilingual,
      title={LAMAR: An Open Language-Aware Multilingual Alignment Reranker}, 
      author={Seongtae Hong and Youngjoon Jang and Jungseob Lee and Seungyoon Lee and Heuiseok Lim},
      year={2026},
      eprint={2607.22042},
      archivePrefix={arXiv},
      primaryClass={cs.IR},
      url={https://arxiv.org/abs/2607.22042}, 
}

@misc{devries2020ecologicallyvalidresearchlanguage,
      title={Towards Ecologically Valid Research on Language User Interfaces}, 
      author={Harm de Vries and Dzmitry Bahdanau and Christopher Manning},
      year={2020},
      eprint={2007.14435},
      archivePrefix={arXiv},
      primaryClass={cs.CL},
      url={https://arxiv.org/abs/2007.14435}, 
}

@inproceedings{lee-etal-2026-clear,
    title = "{CLEAR}: Cross-Lingual Enhancement in Retrieval via Reverse-training",
    author = "Lee, Seungyoon  and
      Kim, Minhyuk  and
      Hong, Seongtae  and
      Jang, Youngjoon  and
      Oh, Dongsuk  and
      Lim, Heuiseok",
    editor = "Liakata, Maria  and
      Moreira, Viviane P.  and
      Zhang, Jiajun  and
      Jurgens, David",
    booktitle = "Proceedings of the 64th Annual Meeting of the {A}ssociation for {C}omputational {L}inguistics (Volume 1: Long Papers)",
    month = jul,
    year = "2026",
    address = "San Diego, California, United States",
    publisher = "Association for Computational Linguistics",
    url = "https://aclanthology.org/2026.acl-long.13/",
    doi = "10.18653/v1/2026.acl-long.13",
    pages = "347--362",
    ISBN = "979-8-89176-390-6",
}

@inproceedings{amiraz-etal-2025-cross,
    title = "The Cross-Lingual Cost: Retrieval Biases in {RAG} over {A}rabic-{E}nglish Corpora",
    author = "Amiraz, Chen  and
      Fyodorov, Yaroslav  and
      Haramaty, Elad  and
      Karnin, Zohar  and
      Lewin-Eytan, Liane",
    editor = "Darwish, Kareem  and
      Ali, Ahmed  and
      Abu Farha, Ibrahim  and
      Touileb, Samia  and
      Zitouni, Imed  and
      Abdelali, Ahmed  and
      Al-Ghamdi, Sharefah  and
      Alkhereyf, Sakhar  and
      Zaghouani, Wajdi  and
      Khalifa, Salam  and
      AlKhamissi, Badr  and
      Almatham, Rawan  and
      Hamed, Injy  and
      Alyafeai, Zaid  and
      Alowisheq, Areeb  and
      Inoue, Go  and
      Mrini, Khalil  and
      Alshammari, Waad",
    booktitle = "Proceedings of The Third Arabic Natural Language Processing Conference",
    month = nov,
    year = "2025",
    address = "Suzhou, China",
    publisher = "Association for Computational Linguistics",
    url = "https://aclanthology.org/2025.arabicnlp-main.6/",
    doi = "10.18653/v1/2025.arabicnlp-main.6",
    pages = "69--83",
    ISBN = "979-8-89176-352-4",
}

@article{flaxman2016filter,
  title={Filter bubbles, echo chambers, and online news consumption},
  author={Flaxman, Seth and Goel, Sharad and Rao, Justin M},
  journal={Public opinion quarterly},
  volume={80},
  number={S1},
  pages={298--320},
  year={2016},
  publisher={Oxford University Press US}
}

@incollection{sunstein2018republic,
  title={\# Republic: Divided democracy in the age of social media},
  author={Sunstein, Cass R},
  booktitle={\# Republic},
  year={2018},
  publisher={Princeton university press}
}

@inproceedings{yang2024language,
  title={Language fairness in multilingual information retrieval},
  author={Yang, Eugene and J{\"a}nich, Thomas and Mayfield, James and Lawrie, Dawn},
  booktitle={Proceedings of the 47th International ACM SIGIR Conference on Research and Development in Information Retrieval},
  pages={2487--2491},
  year={2024}
}

@inproceedings{li2025multilingual,
  title={Multilingual retrieval augmented generation for culturally-sensitive tasks: A benchmark for cross-lingual robustness},
  author={Li, Bryan and Luo, Fiona and Haider, Samar and Agashe, Adwait and Li, Siyu and Liu, Runqi and Miao, Miranda Muqing and Ramakrishnan, Shriya and Yuan, Yuan and Callison-Burch, Chris},
  booktitle={Findings of the Association for Computational Linguistics: ACL 2025},
  pages={4215--4241},
  year={2025}
}

@inproceedings{qi-etal-2025-consistency,
    title = "On the Consistency of Multilingual Context Utilization in Retrieval-Augmented Generation",
    author = "Qi, Jirui  and
      Fern{\'a}ndez, Raquel  and
      Bisazza, Arianna",
    editor = "Adelani, David Ifeoluwa  and
      Arnett, Catherine  and
      Ataman, Duygu  and
      Chang, Tyler A.  and
      Gonen, Hila  and
      Raja, Rahul  and
      Schmidt, Fabian  and
      Stap, David  and
      Wang, Jiayi",
    booktitle = "Proceedings of the 5th Workshop on Multilingual Representation Learning (MRL 2025)",
    month = nov,
    year = "2025",
    address = "Suzhuo, China",
    publisher = "Association for Computational Linguistics",
    url = "https://aclanthology.org/2025.mrl-main.15/",
    doi = "10.18653/v1/2025.mrl-main.15",
    pages = "199--225",
    ISBN = "979-8-89176-345-6",
}

@inproceedings{samir-etal-2024-locating,
    title = "Locating Information Gaps and Narrative Inconsistencies Across Languages: A Case Study of {LGBT} People Portrayals on {W}ikipedia",
    author = "Samir, Farhan  and
      Park, Chan Young  and
      Field, Anjalie  and
      Shwartz, Vered  and
      Tsvetkov, Yulia",
    editor = "Al-Onaizan, Yaser  and
      Bansal, Mohit  and
      Chen, Yun-Nung",
    booktitle = "Proceedings of the 2024 Conference on Empirical Methods in Natural Language Processing",
    month = nov,
    year = "2024",
    address = "Miami, Florida, USA",
    publisher = "Association for Computational Linguistics",
    url = "https://aclanthology.org/2024.emnlp-main.384/",
    doi = "10.18653/v1/2024.emnlp-main.384",
    pages = "6747--6762",
}

@inproceedings{hecht2010tower,
  title={The tower of Babel meets web 2.0: user-generated content and its applications in a multilingual context},
  author={Hecht, Brent and Gergle, Darren},
  booktitle={Proceedings of the SIGCHI conference on human factors in computing systems},
  pages={291--300},
  year={2010}
}

@misc{zhao2026abundanceconcealsweaknessknowledge,
      title={When Abundance Conceals Weakness: Knowledge Conflict in Multilingual Models}, 
      author={Jiaqi Zhao and Qiang Huang and Haodong Chen and Xiaoxing You and Jun Yu},
      year={2026},
      eprint={2601.07041},
      archivePrefix={arXiv},
      primaryClass={cs.CL},
      url={https://arxiv.org/abs/2601.07041}, 
}

@article{loshchilov2017decoupled,
  title={Decoupled weight decay regularization},
  author={Loshchilov, Ilya and Hutter, Frank},
  journal={arXiv preprint arXiv:1711.05101},
  year={2017}
}

@book{tukey1977exploratory,
  author    = {Tukey, John W.},
  title     = {Exploratory Data Analysis},
  publisher = {Addison-Wesley},
  address   = {Reading, MA},
  year      = {1977}
}

@inproceedings{massa2012manypedia,
  title={Manypedia: Comparing language points of view of Wikipedia communities},
  author={Massa, Paolo and Scrinzi, Federico},
  booktitle={Proceedings of the Eighth Annual International Symposium on Wikis and Open Collaboration},
  pages={1--9},
  year={2012}
}

@inproceedings{miquel2020wikipedia,
  title={The wikipedia diversity observatory: A project to identify and bridge content gaps in wikipedia},
  author={Miquel-Rib{\'e}, Marc and Laniado, David},
  booktitle={Proceedings of the 16th International Symposium on Open Collaboration},
  pages={1--4},
  year={2020}
}

@inproceedings{bao2012omnipedia,
  title={Omnipedia: bridging the wikipedia language gap},
  author={Bao, Patti and Hecht, Brent and Carton, Samuel and Quaderi, Mahmood and Horn, Michael and Gergle, Darren},
  booktitle={Proceedings of the SIGCHI Conference on Human Factors in Computing Systems},
  pages={1075--1084},
  year={2012}
}

@phdthesis{hecht2013mining,
  title={The mining and application of diverse cultural perspectives in user-generated content},
  author={Hecht, Brent Jaron},
  year={2013},
  school={Northwestern University}
}

@book{kumar2021digital,
  title={The digital frontier: Infrastructures of control on the global web},
  author={Kumar, Sangeet},
  year={2021},
  publisher={Indiana University Press}
}

@article{miquel2018wikipedia,
  title={Wikipedia culture gap: quantifying content imbalances across 40 language editions},
  author={Miquel-Rib{\'e}, Marc and Laniado, David},
  journal={Frontiers in physics},
  volume={6},
  pages={54},
  year={2018},
  publisher={Frontiers Media SA}
}

@inproceedings{sen2015barriers,
  title={Barriers to the localness of volunteered geographic information},
  author={Sen, Shilad W and Ford, Heather and Musicant, David R and Graham, Mark and Keyes, OS and Hecht, Brent},
  booktitle={Proceedings of the 33rd Annual ACM Conference on Human Factors in Computing Systems},
  pages={197--206},
  year={2015}
}

@article{haraway1988situated,
  title={Situated knowledges: The science question in feminism and the privilege of partial perspective},
  author={Haraway, Donna},
  journal={Feminist studies},
  volume={14},
  number={3},
  pages={575--599},
  year={1988},
  publisher={JSTOR}
}

@article{johnson2022considerations,
  title={Considerations for multilingual wikipedia research},
  author={Johnson, Isaac and Lescak, Emily},
  journal={arXiv preprint arXiv:2204.02483},
  year={2022}
}

@article{callahan2011cultural,
  title={Cultural bias in Wikipedia content on famous persons},
  author={Callahan, Ewa S and Herring, Susan C},
  journal={Journal of the American society for information science and technology},
  volume={62},
  number={10},
  pages={1899--1915},
  year={2011},
  publisher={Wiley Online Library}
}

@inproceedings{hecht2009measuring,
  title={Measuring self-focus bias in community-maintained knowledge repositories},
  author={Hecht, Brent and Gergle, Darren},
  booktitle={Proceedings of the fourth international conference on communities and technologies},
  pages={11--20},
  year={2009}
}

@inproceedings{hale2014multilinguals,
  title={Multilinguals and Wikipedia editing},
  author={Hale, Scott A},
  booktitle={Proceedings of the 2014 ACM conference on Web science},
  pages={99--108},
  year={2014}
}

@inproceedings{mo-etal-2026-faithfulness,
    title = "Faithfulness vs. Safety: Evaluating {LLM} Behavior Under Counterfactual Medical Evidence",
    author = "Mo, Kaijie  and
      Venkatayogi, Siddhartha  and
      Shaib, Chantal  and
      Kouzy, Ramez  and
      Xu, Wei  and
      Wallace, Byron C  and
      Li, Junyi Jessy",
    editor = "Liakata, Maria  and
      Moreira, Viviane P.  and
      Zhang, Jiajun  and
      Jurgens, David",
    booktitle = "Findings of the {A}ssociation for {C}omputational {L}inguistics: {ACL} 2026",
    month = jul,
    year = "2026",
    address = "San Diego, California, United States",
    publisher = "Association for Computational Linguistics",
    url = "https://aclanthology.org/2026.findings-acl.1847/",
    doi = "10.18653/v1/2026.findings-acl.1847",
    pages = "37053--37081",
    ISBN = "979-8-89176-395-1",
}

@inproceedings{chen-etal-2022-rich,
    title = "Rich Knowledge Sources Bring Complex Knowledge Conflicts: Recalibrating Models to Reflect Conflicting Evidence",
    author = "Chen, Hung-Ting  and
      Zhang, Michael  and
      Choi, Eunsol",
    editor = "Goldberg, Yoav  and
      Kozareva, Zornitsa  and
      Zhang, Yue",
    booktitle = "Proceedings of the 2022 Conference on Empirical Methods in Natural Language Processing",
    month = dec,
    year = "2022",
    address = "Abu Dhabi, United Arab Emirates",
    publisher = "Association for Computational Linguistics",
    url = "https://aclanthology.org/2022.emnlp-main.146/",
    doi = "10.18653/v1/2022.emnlp-main.146",
    pages = "2292--2307",
}

@misc{guerreiro2023xcomettransparentmachinetranslation,
      title={xCOMET: Transparent Machine Translation Evaluation through Fine-grained Error Detection}, 
      author={Nuno M. Guerreiro and Ricardo Rei and Daan van Stigt and Luisa Coheur and Pierre Colombo and André F. T. Martins},
      year={2023},
      eprint={2310.10482},
      archivePrefix={arXiv},
      primaryClass={cs.CL},
      url={https://arxiv.org/abs/2310.10482}, 
}

@inproceedings{sun-etal-2026-task,
    title = "Task Matters: Knowledge Requirements Shape {LLM} Responses to Context{--}Memory Conflict",
    author = "Sun, Kaiser  and
      Bai, Fan  and
      Dredze, Mark",
    editor = "Liakata, Maria  and
      Moreira, Viviane P.  and
      Zhang, Jiajun  and
      Jurgens, David",
    booktitle = "Findings of the {A}ssociation for {C}omputational {L}inguistics: {ACL} 2026",
    month = jul,
    year = "2026",
    address = "San Diego, California, United States",
    publisher = "Association for Computational Linguistics",
    url = "https://aclanthology.org/2026.findings-acl.202/",
    doi = "10.18653/v1/2026.findings-acl.202",
    pages = "4154--4176",
    ISBN = "979-8-89176-395-1",
}

@misc{li2024formalityfavoredunravelinglearning,
      title={Formality is Favored: Unraveling the Learning Preferences of Large Language Models on Data with Conflicting Knowledge}, 
      author={Jiahuan Li and Yiqing Cao and Shujian Huang and Jiajun Chen},
      year={2024},
      eprint={2410.04784},
      archivePrefix={arXiv},
      primaryClass={cs.CL},
      url={https://arxiv.org/abs/2410.04784}, 
}

@misc{kang2025largevisionlanguagemodelneeds,
      title={Your Large Vision-Language Model Only Needs A Few Attention Heads For Visual Grounding}, 
      author={Seil Kang and Jinyeong Kim and Junhyeok Kim and Seong Jae Hwang},
      year={2025},
      eprint={2503.06287},
      archivePrefix={arXiv},
      primaryClass={cs.CV},
      url={https://arxiv.org/abs/2503.06287}, 
}

@inproceedings{yeh-li-2026-retrieved,
    title = "How Retrieved Context Shapes Internal Representations in {RAG}",
    author = "Yeh, Samuel  and
      Li, Sharon",
    editor = "Liakata, Maria  and
      Moreira, Viviane P.  and
      Zhang, Jiajun  and
      Jurgens, David",
    booktitle = "Findings of the {A}ssociation for {C}omputational {L}inguistics: {ACL} 2026",
    month = jul,
    year = "2026",
    address = "San Diego, California, United States",
    publisher = "Association for Computational Linguistics",
    url = "https://aclanthology.org/2026.findings-acl.758/",
    doi = "10.18653/v1/2026.findings-acl.758",
    pages = "15446--15468",
    ISBN = "979-8-89176-395-1",
}

@inproceedings{jin-etal-2024-cutting,
    title = "Cutting Off the Head Ends the Conflict: A Mechanism for Interpreting and Mitigating Knowledge Conflicts in Language Models",
    author = "Jin, Zhuoran  and
      Cao, Pengfei  and
      Yuan, Hongbang  and
      Chen, Yubo  and
      Xu, Jiexin  and
      Li, Huaijun  and
      Jiang, Xiaojian  and
      Liu, Kang  and
      Zhao, Jun",
    editor = "Ku, Lun-Wei  and
      Martins, Andre  and
      Srikumar, Vivek",
    booktitle = "Findings of the Association for Computational Linguistics: ACL 2024",
    month = aug,
    year = "2024",
    address = "Bangkok, Thailand",
    publisher = "Association for Computational Linguistics",
    url = "https://aclanthology.org/2024.findings-acl.70/",
    doi = "10.18653/v1/2024.findings-acl.70",
    pages = "1193--1215",
}

@inproceedings{
li2025taming,
title={Taming Knowledge Conflicts in Language Models},
author={Gaotang Li and Yuzhong Chen and Hanghang Tong},
booktitle={Forty-second International Conference on Machine Learning},
year={2025},
url={https://openreview.net/forum?id=0cEZyhHEks}
}

@inproceedings{jin-etal-2024-tug,
    title = "Tug-of-War between Knowledge: Exploring and Resolving Knowledge Conflicts in Retrieval-Augmented Language Models",
    author = "Jin, Zhuoran  and
      Cao, Pengfei  and
      Chen, Yubo  and
      Liu, Kang  and
      Jiang, Xiaojian  and
      Xu, Jiexin  and
      Qiuxia, Li  and
      Zhao, Jun",
    editor = "Calzolari, Nicoletta  and
      Kan, Min-Yen  and
      Hoste, Veronique  and
      Lenci, Alessandro  and
      Sakti, Sakriani  and
      Xue, Nianwen",
    booktitle = "Proceedings of the 2024 Joint International Conference on Computational Linguistics, Language Resources and Evaluation (LREC-COLING 2024)",
    month = may,
    year = "2024",
    address = "Torino, Italia",
    publisher = "ELRA and ICCL",
    url = "https://aclanthology.org/2024.lrec-main.1466/",
    pages = "16867--16878",
}

@misc{
zhu2026focus,
title={Focus Directions Make Your Language Models Pay More Attention to Relevant Contexts},
author={Youxiang Zhu and Ruochen Li and Danqing Wang and Daniel Haehn and Xiaohui Liang},
year={2026},
url={https://openreview.net/forum?id=h7Tm6dSWWp}
}

@inproceedings{jiang-etal-2025-hicd,
    title = "{HICD}: Hallucination-Inducing via Attention Dispersion for Contrastive Decoding to Mitigate Hallucinations in Large Language Models",
    author = "Jiang, Xinyan  and
      Ye, Hang  and
      Zhu, Yongxin  and
      Zheng, Xiaoying  and
      Chen, Zikang  and
      Gong, Jun",
    editor = "Che, Wanxiang  and
      Nabende, Joyce  and
      Shutova, Ekaterina  and
      Pilehvar, Mohammad Taher",
    booktitle = "Findings of the Association for Computational Linguistics: ACL 2025",
    month = jul,
    year = "2025",
    address = "Vienna, Austria",
    publisher = "Association for Computational Linguistics",
    url = "https://aclanthology.org/2025.findings-acl.405/",
    doi = "10.18653/v1/2025.findings-acl.405",
    pages = "7764--7786",
    ISBN = "979-8-89176-256-5",
}

@inbook{sixteen_2019,
author = {Michel, Paul and Levy, Omer and Neubig, Graham},
title = {Are sixteen heads really better than one?},
year = {2019},
publisher = {Curran Associates Inc.},
address = {Red Hook, NY, USA},
booktitle = {Proceedings of the 33rd International Conference on Neural Information Processing Systems},
articleno = {1257},
numpages = {11}
}

@inproceedings{focus_2026,
author = {Liu, Xin and Song, Qiyang and Zhou, Qihang and Du, Haichao and Xu, Shaowen and Jiang, Wenbo and Zhang, Weijuan and Jia, Xiaoqi},
title = {Focusing on language: revealing and exploiting language attention heads in multilingual large language models},
year = {2026},
isbn = {978-1-57735-906-7},
publisher = {AAAI Press},
url = {https://doi.org/10.1609/aaai.v40i38.40492},
doi = {10.1609/aaai.v40i38.40492},
booktitle = {Proceedings of the Fortieth AAAI Conference on Artificial Intelligence and Thirty-Eighth Conference on Innovative Applications of Artificial Intelligence and Sixteenth Symposium on Educational Advances in Artificial Intelligence},
articleno = {3591},
numpages = {9},
series = {AAAI'26/IAAI'26/EAAI'26}
}

@inproceedings{
wang2023interpretability,
title={Interpretability in the Wild: a Circuit for Indirect Object Identification in {GPT}-2 Small},
author={Kevin Ro Wang and Alexandre Variengien and Arthur Conmy and Buck Shlegeris and Jacob Steinhardt},
booktitle={The Eleventh International Conference on Learning Representations },
year={2023},
url={https://openreview.net/forum?id=NpsVSN6o4ul}
}

@inproceedings{crosbie-shutova-2025-induction,
    title = "Induction Heads as an Essential Mechanism for Pattern Matching in In-context Learning",
    author = "Crosbie, Joy  and
      Shutova, Ekaterina",
    editor = "Chiruzzo, Luis  and
      Ritter, Alan  and
      Wang, Lu",
    booktitle = "Findings of the Association for Computational Linguistics: NAACL 2025",
    month = apr,
    year = "2025",
    address = "Albuquerque, New Mexico",
    publisher = "Association for Computational Linguistics",
    url = "https://aclanthology.org/2025.findings-naacl.283/",
    doi = "10.18653/v1/2025.findings-naacl.283",
    pages = "5049--5111",
    ISBN = "979-8-89176-195-7",
}

@inproceedings{nam2025causal,
  title={Causal Head Gating: A Framework for Interpreting Roles of Attention Heads in Transformers},
  author={Nam, Andrew and Conklin, Henry and Yang, Yukang and Griffiths, Thomas and Cohen, Jonathan and Leslie, Sarah-Jane},
  booktitle={Advances in Neural Information Processing Systems (NeurIPS)},
  year={2025}
}

@inproceedings{feng-etal-2022-language,
    title = "Language-agnostic {BERT} Sentence Embedding",
    author = "Feng, Fangxiaoyu  and
      Yang, Yinfei  and
      Cer, Daniel  and
      Arivazhagan, Naveen  and
      Wang, Wei",
    editor = "Muresan, Smaranda  and
      Nakov, Preslav  and
      Villavicencio, Aline",
    booktitle = "Proceedings of the 60th Annual Meeting of the Association for Computational Linguistics (Volume 1: Long Papers)",
    month = may,
    year = "2022",
    address = "Dublin, Ireland",
    publisher = "Association for Computational Linguistics",
    url = "https://aclanthology.org/2022.acl-long.62/",
    doi = "10.18653/v1/2022.acl-long.62",
    pages = "878--891",
}

@article{zhang-etal-2023-miracl,
    title = {{{MIRACL}: A Multilingual Retrieval Dataset Covering 18 Diverse Languages}},
    author = "Zhang, Xinyu  and
      Thakur, Nandan  and
      Ogundepo, Odunayo  and
      Kamalloo, Ehsan  and
      Alfonso-Hermelo, David  and
      Li, Xiaoguang  and
      Liu, Qun  and
      Rezagholizadeh, Mehdi  and
      Lin, Jimmy",
    journal = "Transactions of the Association for Computational Linguistics",
    volume = "11",
    year = "2023",
    address = "Cambridge, MA",
    publisher = "MIT Press",
    url = "https://aclanthology.org/2023.tacl-1.63/",
    doi = "10.1162/tacl_a_00595",
    pages = "1114--1131",
}

@inproceedings{conflict_2026,
author = {Zhao, Tianzhe and Chen, Jiaoyan and Zhang, Shuxiu and Zhu, Haiping and Lin, Qika and Liu, Jun},
title = {Exploring Knowledge Conflicts for Faithful LLM Reasoning: Benchmark and Method},
year = {2026},
isbn = {9798400725999},
publisher = {Association for Computing Machinery},
address = {New York, NY, USA},
url = {https://doi.org/10.1145/3805712.3809650},
doi = {10.1145/3805712.3809650},
booktitle = {Proceedings of the 49th International ACM SIGIR Conference on Research and Development in Information Retrieval},
pages = {2464–2474},
numpages = {11},
location = {Australia},
series = {SIGIR '26}
}

@inproceedings{longpre-etal-2021-entity,
    title = "Entity-Based Knowledge Conflicts in Question Answering",
    author = "Longpre, Shayne  and
      Perisetla, Kartik  and
      Chen, Anthony  and
      Ramesh, Nikhil  and
      DuBois, Chris  and
      Singh, Sameer",
    editor = "Moens, Marie-Francine  and
      Huang, Xuanjing  and
      Specia, Lucia  and
      Yih, Scott Wen-tau",
    booktitle = "Proceedings of the 2021 Conference on Empirical Methods in Natural Language Processing",
    month = nov,
    year = "2021",
    address = "Online and Punta Cana, Dominican Republic",
    publisher = "Association for Computational Linguistics",
    url = "https://aclanthology.org/2021.emnlp-main.565/",
    doi = "10.18653/v1/2021.emnlp-main.565",
    pages = "7052--7063",
}

@inproceedings{
wang2024resolving,
title={Resolving Knowledge Conflicts in Large Language Models},
author={Yike Wang and Shangbin Feng and Heng Wang and Weijia Shi and Vidhisha Balachandran and Tianxing He and Yulia Tsvetkov},
booktitle={First Conference on Language Modeling},
year={2024},
url={https://openreview.net/forum?id=ptvV5HGTNN}
}
\bibliographystyle{iclr2027_conference}

\appendix
\clearpage
\section*{Appendix}

\section{Details for Creating Waldo}
\label{appendix:waldo_details}

\subsection{Source Languages}
\label{appendix:lang_stats}

We present detailed statistics for each source language in~\autoref{tab:lang_stats}.

\definecolor{midpink}{RGB}{225, 149, 171}
\definecolor{darkpink}{RGB}{186, 67, 101}
\definecolor{darkerpink}{RGB}{99, 36, 54}

\definecolor{midgreen}{RGB}{96, 171, 117}
\definecolor{darkgreen}{RGB}{32, 84, 47}
\definecolor{darkergreen}{RGB}{0, 156, 43}

\definecolor{midblue}{RGB}{66, 135, 245}
\definecolor{darkkblue}{RGB}{35, 77, 145}
\definecolor{darkerblue}{RGB}{96, 117, 150}

\begin{table}[H]
\centering
\resizebox{\linewidth}{!}{%
    \begin{tabular}{llllllrr}
    \toprule
    \textbf{Family} & \textbf{Language} & \textbf{Script} & \textbf{Synthesis} & \textbf{Word Order} & \textbf{Resource Level} & \textbf{\# Speakers} & \textbf{\# Wikipedia Size} \\
    \midrule

    \multirow{3}{*}{Indo-European} & English & Latin & \textcolor{midblue}{analytic} & \textcolor{midpink}{SVO} & \cellcolor{midgreen!40} high & \cellcolor{midgreen!80} 1,130M & \cellcolor{darkgreen!50} 5,758,285 \\
    
    & French & Latin & \textcolor{darkkblue}{fusional} & \textcolor{midpink}{SVO} & \cellcolor{midgreen!40} high & \cellcolor{midgreen!30} 398M & \cellcolor{darkgreen!30} 2,325,608 \\
    
    & Bengali & Bengali & \textcolor{darkkblue}{fusional} & \textcolor{darkpink}{SOV} & \cellcolor{red!20} low & \cellcolor{midgreen!30} 337M & \cellcolor{red!15} 63,762 \\
    \midrule
    
    Sino-Tibetan & Chinese & Chinese & \textcolor{midblue}{analytic} & \textcolor{midpink}{SVO} & \cellcolor{midgreen!40} high & \cellcolor{midgreen!80} 1,350M & \cellcolor{darkgreen!20} 1,246,389 \\
    \midrule
    
    Koreanic & Korean & Hangul & \textcolor{darkerblue}{agglutinative} & \textcolor{darkpink}{SOV} & \cellcolor{orange!30} mid & \cellcolor{red!30} 80M & \cellcolor{orange!15} 437,373 \\
    \midrule

    Japonic & Japanese & Japanese & \textcolor{darkerblue}{agglutinative} & \textcolor{darkpink}{SOV} & \cellcolor{orange!30} mid & \cellcolor{red!20} 128M & \cellcolor{darkgreen!20} 1,133,444 \\

    \bottomrule
    \end{tabular}
}
\caption{\textbf{Characteristics of source languages.} For each language, we show language family, script, linguistic typologies (synthesis and word order), and resource level measured by the number of speakers and Wikipedia articles \citep{zhang-etal-2023-miracl}.}  
\label{tab:lang_stats}
\end{table}

\subsection{Heuristics \& LLM Annotation}
\label{appendix:heuristics} 

\paragraph{Heuristics.}
We conduct a manual audit of a sample of sourced Wikipedia revisions to develop a filtering procedure for our controlled evaluation. Based on this audit, we define a suite of heuristic rules that retain only revisions meeting our criteria. We apply the following rules sequentially:
\begin{itemize}[leftmargin=15pt, itemsep=2pt, parsep=-1pt]
    \item \textbf{Required added text:} Drop a row when the added text content is missing, non-string, or empty after trimming whitespace.
    \item \textbf{Remove bot edits:} Drop a row if either the edit comment or contributor username contains a language-specific bot term.
    \item \textbf{Remove maintenance edits:} Drop a row if its comment matches language-specific maintenance keywords such as typo/spelling fixes, links, references, formatting, categories, templates, cleanup, or wikification.
    \item \textbf{Remove reverts:} Drop a row if its comment contains language-specific revert indicators, including terms like \texttt{revert}, \texttt{undo}, \texttt{rv}, \texttt{vandal}, or their equivalents in other source languages.
    \item \textbf{Remove high-symbol, low-content edits:} Drop a row if either comment or the added text has at least 12 characters, at least 60\% non-content symbols, and no more than 8 alphanumeric or source-script characters.
    \item \textbf{Remove markup-heavy additions:} Drop a row if the added text contains wiki or HTML markup patterns and either at least half of its characters are non-content symbols or it contains four or more pipe characters ($\mid$).
    \item \textbf{Remove repetition:} Drop a row if the added text contains a character repeated at least 8 consecutive times.
    \item \textbf{Require punctuation:} Drop a row if the added text does not contain a language-specific sentence terminator.
    \item \textbf{Require source-language text:} Use \textsc{lingua} language detector tool\footnote{\url{https://pypi.org/project/lingua-language-detector/}} on the added text and drop a row if it is undetectable or detected as a language other than the current source language.
    \item \textbf{Remove length outliers:} Compute character-length quartiles of non-empty added text content, then drop rows outside Tukey's $\mathrm{Q}1-1.5\times\mathrm{IQR}$ to  $\mathrm{Q}3+1.5\times\mathrm{IQR}$ bounds \citep{tukey1977exploratory}.
\end{itemize}

\paragraph{LLM Annotation.}
As shown in Prompts A.1--A.2 (Appendix~\ref{appendix:prompts}), we instruct the model to provide a binary verdict and rationale for whether: (1) the added text introduces at least one factual statement without removing existing information, and (2) the added text is absent from the latest version of the corresponding English Wikipedia document, identified using the same Wikidata QID. 
To determine whether the added text appears in the latest source-language Wikipedia document, we retrieve the document and perform exact string matching.

We report the retention rate for each source language after (1) heuristic filtering, (2) LLM annotation in A.1, and (3) LLM annotation in A.2 in~\autoref{tab:annotation}. This yields approximately 4.8K retained revisions across the five source languages.

\paragraph{Additional LLM Annotation Variants.}
We test five additional annotation variants on 100 randomly sampled revisions and manually audited their outputs. These analyses are used to qualitatively refine the LLM Annotation prompt:
\begin{itemize}[leftmargin=15pt, itemsep=2pt, parsep=-1pt]
    \item \textbf{Adding the first paragraph.}
    We add the first paragraph of each Wikipedia article as additional context, because it often provides a high-level overview of the topic. This variant increases the retention rate from 8\% to 9\%, corresponding to one additional retained revision. However, manual inspection found that this additional example did not satisfy our filtering criteria. We therefore decide not to include the first paragraph in the final prompt.
        \item \textbf{Providing Wikipedia context.}
    We add a brief definition of Wikipedia to provide additional context, which led to no observable change in the retention rate (8\%), so we omit this definition.
        \item \textbf{Removing the constraint on deleted text.}
    We test removing the instruction that the deleted text must not remove meaningful information, to accommodate revisions that replace an incorrect fact with contradictory factual information. However, without this constraint, the LLM fails to retain some revisions that simultaneously add a standalone fact and remove unrelated content. We therefore retain the constraint in the final prompt.
    \item \textbf{Allowing general knowledge.}
    We instruct the LLM to use both the provided context and general commonsense knowledge during the annotation. In the manual audit, this helped filter revisions containing irrelevant jokes or degenerate edits, thus we retain this instruction.
        \item \textbf{Translating the prompt.}
    We translate the full prompt into the source language for revisions in that language. This variant did not yield an observable overall improvement and missed several revisions judged meaningful in the manual audit. We use the English prompt for all languages.
\end{itemize}

\begin{table}[H]
\centering
\resizebox{\linewidth}{!}{%
    \begin{tabular}{lrrrr}
    \toprule
    \textbf{Language} & \textbf{Original} & \textbf{(1) Heuristics} & \textbf{(2) LLM Annotation (A.1)} & \textbf{(3) LLM Annotation (A.2)} \\
    \midrule

    \textbf{French} & 229,438 & 42,446 (18.5\%) & 3,917 (1.7\%) & 504 (0.2\%) \\
    \textbf{Chinese} & 251,136 & 107,486 (42.8\%) & 6,160 (2.5\%) & 1,227 (0.5\%) \\
    \textbf{Japanese} & 257,041 & 112,841 (43.9\%) & 7,778 (3.0\%) & 1,125 (0.4\%) \\
    \textbf{Korean} & 465,500 & 68,894 (14.8\%) & 5,293 (1.1\%) & 943 (0.2\%) \\
    \textbf{Bengali} & 725,835 & 172,023 (23.7\%) & 5,702 (7.9\%) & 1,043 (0.1\%) \\

    \rowcolor{gray!15}
    \textbf{Average} & & (28.7\%) & (3.2\%) & (0.3\%) \\
    
    \bottomrule
    \end{tabular}
}
\caption{\textbf{Retention rates by each source language after heuristic filtering and LLM annotation.} We compute each retention rate relative to the total number of revisions in the original set.}  
\label{tab:annotation}
\end{table}

\paragraph{\textsc{InfoGAP} Framework.}
As an alternative to LLM annotation for determining whether added text appears in the latest English Wikipedia document, we consider \textsc{InfoGap} \citep{samir-etal-2024-locating}. 
\textsc{InfoGap} identifies information overlaps and gaps between documents in different languages through three stages: (1) fact decomposition, which extracts atomic facts from each document; (2) multilingual alignment, which aligns facts across language versions; and (3) verification, which determines whether aligned facts are semantically equivalent.

Given an English document, we decompose its content into a set of atomic facts using the fact-decomposition prompt of \citet{samir-etal-2024-locating} and the same model used for LLM annotation, \textsc{GPT-5.6-Terra}. 
For multilingual alignment, we embed the decomposed English facts and the added text using \textsc{LaBSE} \citep{feng-etal-2022-language}. 
For each English fact, we compute a hubness score, which is the average cosine similarity to its top-$k$ nearest neighbors, and select the highest-scoring facts as alignment candidates. We then use \textsc{GPT-5.6-Terra} to verify each candidate. If at least one candidate fact survives verification, we label the added text as ``match''; otherwise, as ``no match.''

Compared with our LLM annotation approach, \textsc{InfoGap} produces a slightly lower retention rate. 
This difference primarily arises when the English document contains only part of the information in the added text. 
For example, consider the Korean addition \begin{CJK}{UTF8}{mj}``BMW IX는 BMW에서 제조한 크로스오버 SUV이다.''\end{CJK} (\textit{``BMW iX is a crossover SUV manufactured by BMW.''}). The corresponding English document describes the BMW iX as an electric sport utility vehicle but does not state that it is a crossover SUV. 
Our LLM annotation labels this case as ``no match,'' whereas \textsc{InfoGap} labels it as ``match.'' 
After qualitatively examining several such revisions, we decide to use LLM annotation for our final pipeline because it better handles these partial-overlap cases.
The two methods agree on 45.5\% of ``match'' labels, 97.1\% of ``no match'' labels, and 87.9\% of labels overall.

\paragraph{Examples of Excluded Revisions.}

\autoref{tab:examples} presents qualitative examples of revisions excluded during the Heuristics \& LLM Annotation stage. Most revisions are excluded because they do not meet one or more of the rules.
The resulting set is intentionally a highly curated subset rather than a representative sample of Wikipedia revisions, and the low retention rate reflects the rarity of revisions that support our controlled evaluation design.

\begin{table}[H]
\centering
\resizebox{\linewidth}{!}{%
    \begin{tabular}{lp{5cm}p{5cm}p{7cm}}
    \toprule
    \textbf{Stage} & \textbf{Added text} & \textbf{Removed text} & \textbf{Description}  \\
    \midrule

    \multirow{1}{*}{\textbf{Heuristics}} & * 1895 * 1896 * 1914 * 1919 * 1937 * 1939 * 1945 * 1953 * 1954 * 1963 * 1973 * 1979 & *1895 *1896 *1914 *1919 *1937 *1939 *1945 *1953 *1954 *1963 *1973 *1979 & \textbf{Remove bot edits}; Username with ``TedBot'' and comment as ``Bot: Standardizing template names and styles'' \\
    \cmidrule{2-4}
    & bgcolor=``\#ccffcc''|align=left| |bgcolor=\#87CEEB|AFC|-bgcolor=``\#ffffcc''|align=left| & & \textbf{Remove reverts} \\
    \cmidrule{2-4}
    & \texttwemoji{1f347}\texttwemoji{1f347}\texttwemoji{1f347}\texttwemoji{1f347}\texttwemoji{1f347}\texttwemoji{1f34a}\texttwemoji{1f34a}\texttwemoji{1f34a}\texttwemoji{1f34a}\texttwemoji{1f34a}\texttwemoji{1f95d}\texttwemoji{1f95d}\texttwemoji{1f95d}\texttwemoji{1f95d}\texttwemoji{1f95d} [...] & & \textbf{Remove high-symbol, low-content edits} \\
     \cmidrule{2-4}
    & bgcolor=``\#ccffcc''|align=left| |bgcolor=\#87CEEB|AFC|-bgcolor=``\#ffffcc''|align=left| & & \textbf{Remove markup-heavy additions} \\
     \cmidrule{2-4}
    & \# \# \# \# \# \# \# \# [...] & & \textbf{Remove repetition} \\
    \midrule
    
    \textbf{LLM Annotation} & \begin{CJK}{UTF8}{mj}Yanggu County(楊口郡)is\end{CJK} & \begin{CJK}{UTF8}{mj}Yanggu County(楊口郡,) is,\end{CJK} & Removed two commas without adding any meaningful information \\
    \cmidrule{2-4}

    & \begin{CJK}{UTF8}{mj}사용한다. 하지만 러시아가 2030년까지 벨라루스를 합병한다고 하니 오래 가지는 못할 것 같다.\end{CJK} & \begin{CJK}{UTF8}{mj}사용한다.\end{CJK} & Added text translates to ``\textit{However, given reports that Russia plans to annex Belarus by 2030, it seems unlikely to last long,}'' which expresses speculation rather than a verifiable fact \\
    
    \bottomrule
    \end{tabular}
}
\caption{\textbf{Examples of excluded revisions during the Heuristics \& LLM Annotation stage.} }  
\label{tab:examples}
\end{table}

\subsection{QA Generation}
\label{appendix:qa_translation}

We assess LLM-generated QA pairs with \textsc{xCOMET-XL} \citep{guerreiro2023xcomettransparentmachinetranslation} between source-language and English questions \(\mathcal{Q}\) and answers \(\mathcal{A}\). 
The average scores are 0.960 for \(\mathcal{Q}\) and 0.983 for \(\mathcal{A}\), indicating that the content is well-preserved across languages (\autoref{tab:xcomet}).

\begin{table}[H]
\centering
\resizebox{0.6\linewidth}{!}{%
    \begin{tabular}{lrr}
    \toprule
    \textbf{Language} & \textbf{\textsc{xCOMET}($\mathcal{Q}_\mathrm{src}$, $\mathcal{Q}_\mathrm{en}$)} & \textbf{\textsc{xCOMET}($\mathcal{A}_\mathrm{src}$, $\mathcal{A}_\mathrm{src}$)} \\
    \midrule

    \textbf{French} & 0.971 & 0.990 \\
    \textbf{Chinese} & 0.956 & 0.983 \\
    \textbf{Japanese} & 0.959 & 0.982 \\
    \textbf{Korean} & 0.953 & 0.979 \\
    \textbf{Bengali} & 0.970 & 0.985 \\
    \rowcolor{gray!15}
    \textbf{Average} & 0.960 & 0.983 \\

    \bottomrule
    \end{tabular}
}
\caption{\textbf{xCOMET-XL scores by source language.} We evaluate the translation quality of source-language questions ($\mathcal{Q}_\mathrm{src}$) and answers ($\mathcal{A}_\mathrm{src}$), against their English counterparts, $\mathcal{Q}_\mathrm{en}$ and $\mathcal{A}_\mathrm{en}$. Higher scores indicate higher translation quality.}
\label{tab:xcomet}
\end{table}

\subsection{Qualitative Examples of Knowledge Conflicts}
\label{appendix:conflict_examples}

\autoref{tab:conflict_examples} presents representative QA pairs for each conflict type, contrasting the original target fact with the inconsistent version injected into the English document. The injected alternatives are designed to remain plausible, within a reasonable range of possible values (e.g., ``\textit{TV Tokyo}'' $\rightarrow$ ``\textit{Nippon TV}'' in one of the Japanese examples).

\begin{table*}[t]
\centering
\scriptsize
\renewcommand{\arraystretch}{1.2}
\setlength{\tabcolsep}{2pt}
\begin{tabularx}{\textwidth}{
  >{\raggedright\arraybackslash}p{0.9cm}
  >{\raggedright\arraybackslash}p{1.0cm}
  >{\raggedright\arraybackslash}p{1.2cm}
  >{\raggedright\arraybackslash}X
  >{\raggedright\arraybackslash}p{2.0cm}
  >{\raggedright\arraybackslash}p{1.2cm}
  >{\raggedright\arraybackslash}p{1.2cm}
  >{\raggedright\arraybackslash}X}
\toprule
\textbf{Lang.} & \textbf{Type} & \textbf{Title} & \textbf{Target fact} & \textbf{$\mathcal{Q}_\mathrm{en}$} & \textbf{$\mathcal{A}_\mathrm{en}$} & \textbf{$\mathcal{A}^{\prime}_\mathrm{en}$} & \textbf{Conflicting fact} \\
\midrule

\textbf{French} & Number & Higashi-ku, Fukuoka &
* Kane Tanaka (1903-), supercentenaire japonaise et doyenne de l'humanité depuis 2018 et troisième personne vérifiée la plus âgée de tous les temps. &
What was Kane Tanaka's all-time rank among verified oldest people? &
Third & Fourth &
In 2018, Kane Tanaka was recognized as the fourth verified oldest person of all time. \\
\cmidrule{2-8}
 & Misc & NOAAS Okeanos Explorer &
Fichier:Hyocrinida Okeanos 2021.jpg|Un crinoïde (Hyocrinida) observé dans l'Atlantique nord en 2021. &
What kind of organism was observed in the North Atlantic in 2021 by Okeanos Explorer? &
A crinoid (Hyocrinida) & A sea cucumber (Holo\-thuroidea) &
A sea cucumber (Holothuroidea) was observed in the North Atlantic in 2021 by Okeanos Explorer. \\

\midrule

\textbf{Chinese} & Date & Ruud van Nistelrooy &
\begin{CJK*}{UTF8}{gbsn}*2024年10月，曼联主教练坦哈格被解雇，云尼斯达莱以助教身份临危受命成暂代主帅，首次领军曼联于联赛杯大胜李斯特城5:2。\end{CJK*} &
When did Ruud van Nistelrooy take over as Manchester United's interim manager after Ten Hag was fired? &
October 2024 & September 2024 &
He was originally tasked to work alongside Erik ten Hag, however following the sacking of the latter on 28 September 2024, he was appointed as interim head coach. \\
\cmidrule{2-8}
 & Location & Lists of rail accidents &
\begin{CJK*}{UTF8}{gbsn}*8月12日，阿贝里奥苏格兰铁路一辆InterCity 125型列车于苏格兰阿伯丁郡斯冬希文出轨，并导致机车起火。事故造成至少3人死亡。\end{CJK*} &
Where in Scotland did the Abellio ScotRail InterCity 125 derail on August 12? &
Stonehaven, Aberdeenshire & Inverness, Highland &
On August 12, an Abellio ScotRail InterCity 125 derailed near Inverness, Highland, Scotland, and its locomotive caught fire, killing at least three people. \\
\midrule

\textbf{Japanese} & Date & Cinnamoroll &
\begin{CJK*}{UTF8}{min}2023年3月3日、公式YouTubeチャンネルが開設。と同時にYouTube内にてシナモンの誕生日を記念してショートアニメ「シナモンアニメだもん」が開始された。\end{CJK*} &
What date did Cinnamoroll's official YouTube channel launch? &
March 3, 2023 & April 7, 2023 &
On April 7, 2023, Sanrio launched an official YouTube channel for Cinnamon to stream a short anime. \\
\cmidrule{2-8}
 & Organi\-zation & Shintaro Tsuji &
\begin{CJK*}{UTF8}{min}* 2020年8月27日 - 「普通はテレビ受けないの。今回は絶対に言い残したい事があって最後だから」と約30年ぶりのテレビ取材に応じ、テレビ東京で反戦を語る。\end{CJK*} &
Which TV station aired Shintaro Tsuji talking about anti-war views in a rare TV interview? &
TV Tokyo & Nippon TV &
In a rare television interview, Tsuji spoke about his anti-war views on Nippon TV. \\

\midrule

\textbf{Korean} & Person & Soo Jung Lee &
\begin{CJK}{UTF8}{mj}* 이수정·이다혜. 《이수정, 이다혜의 범죄 영화 프로파일》. 민음사. 2020년 3월 31일.\end{CJK} &
Who co-wrote the book about crime film profiling with Lee Soo-jung? &
Lee Da-hye & Kim Da-hye &
In 2020, Lee co-wrote Lee Soo-jung and Kim Da-hye's Crime Film Profile, a book about crime film profiling, with Kim Da-hye. \\
\cmidrule{2-8}
 & Number & Economy of Germany &
\begin{CJK}{UTF8}{mj}앞으로 동독과 서독의 경제적 상황이 균등해지기 위해 걸리는 시간에 대해서 예측한 결과 할레경제연구소는 2030년까지는 기다려야 한다고 전망했다. 지난 30년간 동독의 노동생산성과 소득은 서독의 85\% 수준에 이르렀고, 실업률 또한 양 지역의 격차가 크지 않은 수준에 도달했다.\end{CJK} &
Over the past 30 years, what percentage of West Germany's level did East Germany's labor productivity and income reach? &
85\% of West Germany's level & 72\% of West Germany's level &
Over the past 30 years, East Germany's labour productivity and income reached 72\% of West Germany's level, and the unemployment-rate gap between the two regions also narrowed substantially. \\

\bottomrule
\end{tabularx}
\caption{\textbf{Examples of original and conflicting facts across languages and conflict types.} $\mathcal{Q}_\mathrm{en}$: English question; $\mathcal{A}_\mathrm{en}$: Original English answer; $\mathcal{A}^{\prime}_\mathrm{en}$: Conflict-injected English answer.}
\label{tab:conflict_examples}
\end{table*}

\subsection{Answerability Check}
\label{appendix:answerability}

After applying the answerability check in the knowledge conflict setting, we retain 11,888 QA pairs across five source languages, which serve as our final evaluation set. The set contains 957 French, 3,787 Chinese, 3,074 Japanese, 1,937 Korean, and 2,133 Bengali QA pairs.

\begin{table}[H]
\centering
\resizebox{\linewidth}{!}{%
    \begin{tabular}{llrrr}
    \toprule
    \textbf{Type} & \textbf{Model} & \textbf{Knowledge Cutoff} & \textbf{Context Window} & \textbf{Post-cutoff QA pairs (/12K)} \\
    \toprule

    \multirow{3}{*}{\raisebox{-0.2em}{\includegraphics[height=1em]{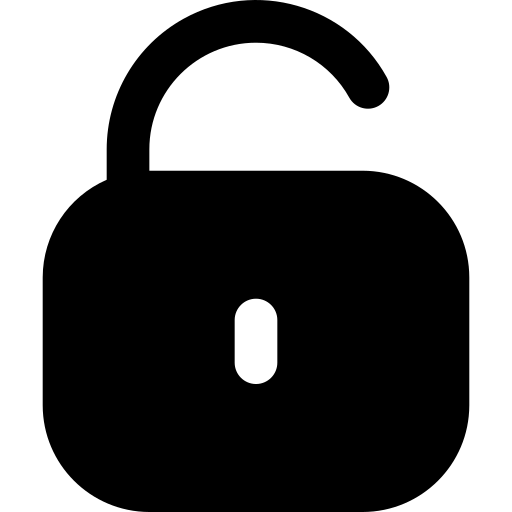}} \textbf{Open-weight}} & \textsc{LLaMA-3.1} 8B & 2023-12 & 128K & 3,551 (30\%) \\
    & \textsc{Qwen-3} 8B & 2023-12 & 128K & 3,551 (30\%) \\
    & \textsc{Aya-Expanse} 8B & 2024-06 & 128K & 2,667 (22\%) \\
    \midrule

    \multirow{5}{*}{\raisebox{-0.2em}{\includegraphics[height=1em]{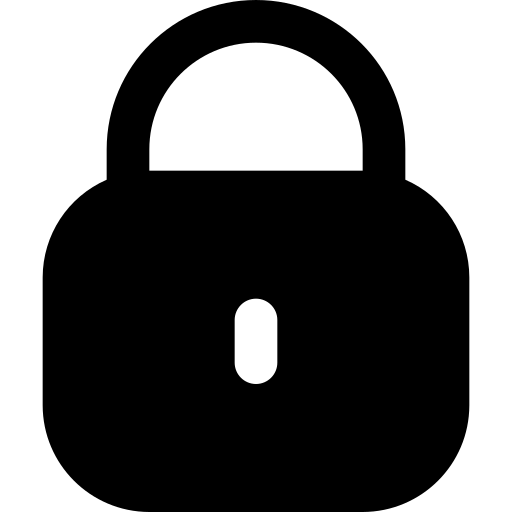}} \textbf{Closed-source}} & \textsc{GPT-4o} & 2023-09 & 128K & 5,129 (43\%) \\
    & \textsc{GPT-4.1} & 2024-05 & 1M & 3,708 (31\%) \\
    & \textsc{DeepSeek-V4-Pro} & 2025-05 & 1M & 1,821 (15\%) \\
    & \textsc{Grok-4.3} & 2025-09 & 1M & 1,283 (11\%) \\
    & \textsc{GPT-5.5} & 2025-12 & 1M & 968 (8\%) \\

    \bottomrule
    \end{tabular}
}
\caption{\textbf{List of evaluated models.} We report each model's knowledge-cutoff date, context-window size, and number of post-cutoff QA pairs. A post-cutoff QA pair is one whose  \texttt{timestamp} postdates the model's reported knowledge cutoff and the one which the model answers incorrectly when only given the question in both the source language and English. When available, we use non-reasoning mode and the default effort setting for closed-source models.}  
\label{tab:model_statistics}
\end{table}

\section{Additional Diagnosis with \textsc{Waldo}}
\label{appendix:more_res}

\subsection{Retrieval-Augmented Robustness Results}
\label{appendix:endtoend}

Our primary experiments provide paired multilingual Wikipedia documents directly to models in order to isolate the effect of query language on evidence use. 
To test whether the pattern persists with additional retrieved evidence, we evaluate 100 randomly sampled QA pairs in a retrieval-augmented setting. For each question, we use OpenAI's web-search tool to retrieve up to five documents from external web sources.\footnote{\url{https://developers.openai.com/api/docs/guides/tools-web-search}} 
We then sample two retrieved documents in the source language and two in English, and provide these four documents together with the original source-language and English Wikipedia documents. Each model therefore receives six documents in total.

As shown in~\autoref{tab:endtoend}, adding retrieved documents generally lowers absolute answer agreement and probability for both source-language and English queries. 
Manual inspection suggests that it introduces additional candidate sources, which increases the difficulty of identifying the target fact. 
Despite this overall reduction, we observe the same pattern as in the controlled two-document setting: answer agreement and probability gaps are small under knowledge gaps but substantially larger under knowledge conflicts. 
These results suggest that query-language preference is not solely an artifact of the controlled two-document context. 

\begin{table}[htbp]
\centering
\resizebox{0.75\linewidth}{!}{%
\begin{tabular}{c l
                S[table-format=2.1]
                S[table-format=2.1]
                S[table-format=2.1, table-space-text-post={\sig{***}}]
                S[table-format=2.1]
                S[table-format=2.1]
                S[table-format=2.1, table-space-text-post={\sig{***}}]}
    \toprule
    \multirow{2}{*}{\textbf{Type}} & \multirow{2}{*}{\textbf{Model}}
      & \multicolumn{3}{c}{\textbf{Agreement (\%)}}
      & \multicolumn{3}{c}{\textbf{Probability (\%)}} \\
    \cmidrule(lr){3-5} \cmidrule(lr){6-8}
    & & {src} & {en} & {$\Delta_\mathrm{agree}$} & {src} & {en} & {$\Delta_\mathrm{prob}$} \\
    \midrule
    \rowcolor{gray!20}
    \multicolumn{8}{c}{\textbf{Knowledge Gap}} \\
    \midrule
    \multirow{3}{*}{\includegraphics[height=1em]{figures/logo/unlock.png}}
      & \textsc{LLaMA-3.1}       & 86.3 & 68.7 & 17.6{\sig{***}} & 36.8 & 20.2 & 16.6{\sig{***}} \\
      & \textsc{Qwen-3}          & 81.2 & 66.5 & 14.7{\sig{***}} & 14.6 &  7.1 &  7.5{\sig{**}}  \\
      & \textsc{Aya-Expanse}     & 59.8 & 41.3 & 18.5{\sig{***}} & 42.1 & 29.4 & 12.7{\sig{***}} \\
    \cmidrule(lr){1-8}
    \multirow{5}{*}{\includegraphics[height=1em]{figures/logo/lock.png}}
      & \textsc{GPT-4o}          & 93.1 & 84.7 &  8.4{\sig{**}}  & \na & \na & \na \\
      & \textsc{GPT-4.1}         & 95.2 & 90.6 &  4.6{\sig{*}}   & \na & \na & \na \\
      & \textsc{DeepSeek-V4-Pro} & 91.4 & 88.3 &  3.1{\sig{*}}   & \na & \na & \na \\
      & \textsc{Grok-4.3}        & 95.8 & 93.7 &  2.1{\sig{*}}   & \na & \na & \na \\
      & \textsc{GPT-5.5}         & 96.4 & 95.1 &  1.3            & \na & \na & \na \\
    \midrule
    \rowcolor{gray!20}
    \multicolumn{8}{c}{\textbf{Knowledge Conflict}} \\
    \midrule
    \multirow{3}{*}{\includegraphics[height=1em]{figures/logo/unlock.png}}
      & \textsc{LLaMA-3.1}       & 76.1 & 10.5 & 65.6{\sig{***}} & 34.2 & 11.8 & 22.4{\sig{***}} \\
      & \textsc{Qwen-3}          & 59.3 & 14.2 & 45.1{\sig{***}} & 12.8 &  4.1 &  8.7{\sig{**}}  \\
      & \textsc{Aya-Expanse}     & 51.7 &  7.4 & 44.3{\sig{***}} & 38.6 & 15.3 & 23.3{\sig{***}} \\
    \cmidrule(lr){1-8}
    \multirow{5}{*}{\includegraphics[height=1em]{figures/logo/lock.png}}
      & \textsc{GPT-4o}          & 61.8 & 14.3 & 47.5{\sig{***}} & \na & \na & \na \\
      & \textsc{GPT-4.1}         & 75.5 & 23.8 & 51.7{\sig{***}} & \na & \na & \na \\
      & \textsc{DeepSeek-V4-Pro} & 67.4 & 21.5 & 45.9{\sig{***}} & \na & \na & \na \\
      & \textsc{Grok-4.3}        & 78.2 & 26.1 & 52.1{\sig{***}} & \na & \na & \na \\
      & \textsc{GPT-5.5}         & 84.9 & 16.8 & 68.1{\sig{***}} & \na & \na & \na \\
    \bottomrule
\end{tabular}%
}
\caption{\textbf{Retrieval-augmented robustness results.} We perform the same statistical tests as in~\autoref{fig:main_results}. Closed-source models do not expose answer probabilities (--). }
\label{tab:endtoend}
\end{table}

\subsection{Per-language Results}
\label{appendix:per_lang}

We report the per-language results for each knowledge gap and conflict setting in Tables~\ref{tab:perlang_gap} and \ref{tab:perlang_conflict}.
Across all five source languages, we observe that the general trend holds: query-language preference is minimal under knowledge gaps but pronounced under conflicts. 
While answer agreement shows no clear pattern by language resource level (\autoref{tab:model_statistics}), answer probability gap is consistently larger for lower-resource languages: under knowledge gaps, 18.3\% (Bengali), 15.2\% (Korean), 14.1\% (Japanese), 10.7\% (Chinese), and 8.4\% (French); under conflicts, 28.0\% (Bengali), 25.8\% (Japanese), 23.5\% (Korean), 17.8\% (Chinese), and 12.6\% (French).

\subsection{Answer Probability for Knowledge Conflicts}
\label{appendix:conflict}

For knowledge conflicts, we additionally measure the probability assigned to the \textit{conflicting} English-supported answer under each query language. This complements the main metric (\S\ref{sec:metrics}), which measures the probability assigned to the source-supported answer.

Let \(\mathcal{A}_{\mathrm{en}}'=(t_1,\ldots,t_k)\) denote the conflicting English-supported answer. For a query in language \(\ell\), we compute its length-normalized likelihood as:
\begin{equation}
P_{\theta}\!\left(
\mathcal{A}_{\mathrm{en}}'
\mid
\mathcal{Q}_{\ell}, \mathcal{C}
\right)
=
\left[
\prod_{m=1}^{k}
p_{\theta}\!\left(
t_m
\mid
\mathcal{Q}_{\ell}, \mathcal{C}, t_{<m}
\right)
\right]^{1/k},
\end{equation}
where \(p_{\theta}\) is the model's next-token distribution and \(t_{<m}\) denotes the preceding answer tokens. We average this score across evaluated QA pairs for each query language and report the conflicting English-supported answer likelihood shift:
\begin{equation}
\Delta_{\mathrm{prob}}^{\mathrm{en}}
=
\mathrm{Prob}(\mathrm{en})
-
\mathrm{Prob}(\mathrm{src}).
\end{equation}
A positive \(\Delta_{\mathrm{prob}}^{\mathrm{en}}\) indicates that the model assigns greater likelihood to the conflicting English answer when queried in English than when queried in the source language; larger values indicate stronger English-directed query-language preference.

\begin{figure*}
    \centering
    \vspace{-2.5em}
    \includegraphics[width=0.5\linewidth]{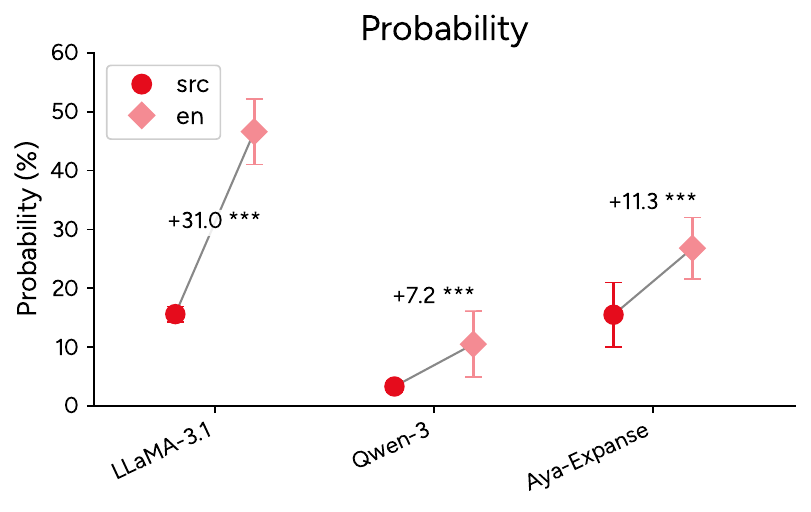}
    \caption{\textbf{Answer probability ($\Delta_\mathrm{prob}^{\mathrm{en}}$) for conflicting English-supported answer.}
    A larger positive $\Delta$(en-src) indicates stronger query-language preference (i.e., English preference). We use Wilcoxon signed-rank test for differences, with Bonferroni correction for multiple comparisons. ***: statistically significant with $p$ $<$ .001.}
    \label{fig:conflict}
\end{figure*}

As shown in~\autoref{fig:conflict}, \(\Delta_{\mathrm{prob}}^{\mathrm{en}}\) is positive for all three open-weight models. This result indicates that query-language preference operates in both directions: models favor the source-language document under source-language queries and the English document under English queries.

\subsection{Additional Evaluation Metrics}
\label{appendix:more_metrics}

We report three additional metrics: (1) \textbf{F1 score gap} complements the LLM-as-judge-based answer agreement metric by measuring token-level overlap with the reference answer through precision and recall. 
We do not report exact match since model responses are often verbose (e.g., ``\textit{It is 43 years.}'' compared to the reference answer ``\textit{43 years}'').
(2) \textbf{Citation attribution} measures whether the model cites the source-language or the English document. Models are instructed to append the cited document as \texttt{[Document ID]} at the end of their response (see Prompt A.7 in Appendix~\ref{appendix:prompts}) \citep{zhang-etal-2025-longcite, ki2026linguistic}, and we extract this ID .
(3) \textbf{Response language} is the language of the model response, detected with the \textsc{lingua} language detection tool.\footnote{\url{https://pypi.org/project/lingua-language-detector/}} We assign 1 if the top-1 detected language matches the query language and 0 otherwise.

As shown in~\autoref{tab:metrics}, F1 score gap mirrors the answer agreement gap. Under knowledge gaps, F1 differences across query languages are small, particularly for closed-source models, whereas under knowledge conflicts, the gap becomes substantially larger.
Citation attribution also shows the same pattern: models more frequently cite the source-language document under source-language queries and more English document under English queries, corroborating the preference through their citation behavior.

Response language stays faithful to the query language in both settings (gap: 93.4\% source-language, 92.9\% English; conflict: 92.1\% source-language, 95.1\% English).
Notably, under knowledge gaps, although the model must retrieve the fact from the source-language document (since it is absent from English), it still mostly responds in the query language. 
This rules out that low answer agreement under English queries stems from an inability to retrieve information cross-lingually, confirming that our answer agreement metric reflects genuine query-language preference rather than a cross-lingual competence gap.

\begin{table}
\centering
\small
\setlength{\tabcolsep}{6pt}
\renewcommand{\arraystretch}{1.6}
\begin{tabular}{
    >{\raggedright\arraybackslash}p{0.18\linewidth}
    |
    >{\raggedright\arraybackslash}p{0.34\linewidth}
    | 
    >{\raggedright\arraybackslash}p{0.34\linewidth}
}
& 
\cellcolor{gray!15}
\centering \textbf{Source-language \ding{51}} \arraybackslash
& \cellcolor{gray!15}
\centering \textbf{Source-language \ding{55}} \arraybackslash \\
\hline
\cellcolor{gray!15}
\textbf{English \ding{51}}
&
Knowledge parity (Appendix~\ref{appendix:truth})
&
Swapped roles (\autoref{tab:swapped})
\\
\hline
\cellcolor{gray!15}
\textbf{English \ding{55}}
&
Main experiments with \textsc{Waldo} (\S\ref{sec:diagnosis})
&
Not dealt in our paper
\\
\end{tabular}
\caption{\textbf{A 2$\times$2 view of tested variants.}
\ding{51}: contains the original target fact; \ding{55}: does not contain the target fact. For completeness, we test additional variants in~\autoref{tab:swapped} and Appendix~\ref{appendix:truth}.}
\label{tab:quadrant}
\end{table}

\begin{table}[htbp]
\centering
\resizebox{0.7\linewidth}{!}{%
\begin{tabular}{c l
                S[table-format=2.1]
                S[table-format=3.1]
                S[table-format=-2.1]
                S[table-format=2.1]
                S[table-format=2.1]
                S[table-format=1.1, table-space-text-post={\sig{**}}]}
    \toprule
    \multirow{2}{*}{\textbf{Type}} & \multirow{2}{*}{\textbf{Model}}
      & \multicolumn{3}{c}{\textbf{Agreement (\%)}}
      & \multicolumn{3}{c}{\textbf{Probability (\%)}} \\
    \cmidrule(lr){3-5} \cmidrule(lr){6-8}
    & & {src} & {en} & {$\Delta_\mathrm{agree}$} & {src} & {en} & {$\Delta_\mathrm{prob}$} \\
    \midrule
    \multirow{3}{*}{\raisebox{-0.2em}{\includegraphics[height=1em]{figures/logo/unlock.png}}}
      & \textsc{LLaMA-3.1}       & 96.3 &  98.4 &  -2.1 & 34.7 & 26.5 & 8.3{\sig{**}} \\
      & \textsc{Qwen-3}          & 95.4 &  95.9 &  -0.4 & 11.7 & 10.1 & 1.6           \\
      & \textsc{Aya-Expanse}     & 71.6 &  84.0 & -12.4 & 43.7 & 34.5 & 9.2{\sig{**}} \\
    \midrule
    \multirow{5}{*}{\raisebox{-0.2em}{\includegraphics[height=1em]{figures/logo/lock.png}}}
      & \textsc{GPT-4o}          & 99.0 &  99.2 &  -0.1 & \na & \na & \na \\
      & \textsc{GPT-4.1}         & 99.8 & 100.0 &  -0.2 & \na & \na & \na \\
      & \textsc{DeepSeek-V4-Pro} & 98.6 &  99.2 &  -0.6 & \na & \na & \na \\
      & \textsc{Grok-4.3}        & 99.2 &  99.6 &  -0.4 & \na & \na & \na \\
      & \textsc{GPT-5.5}         & 98.9 &  99.3 &  -0.4 & \na & \na & \na \\
    \bottomrule
\end{tabular}%
}

\caption{\textbf{Answer agreement ($\Delta_\mathrm{agree}$) and probability ($\Delta_\mathrm{prob}$) for knowledge parity experiment.} \textbf{Agreement} ($\mathrm{src}$, $\mathrm{en}$): answer agreement; \textbf{Probability} ($\mathrm{src}$, $\mathrm{en}$): probability assigned to the source-supported answer, under source-language and English queries, respectively. We perform the same statistical tests as in~\autoref{fig:main_results}. Closed-source models do not expose answer probabilities (--).}
\label{tab:truth}
\end{table}

\begin{table}[htbp]
\centering
\resizebox{0.8\linewidth}{!}{%
\begin{tabular}{clrrrrrrrrr}
    \toprule
    \multirow{2}{*}{\textbf{Type}} & \multirow{2}{*}{\textbf{Model}}
      & \multicolumn{3}{c}{\textbf{F1 score (\%)}}
      & \multicolumn{2}{c}{\textbf{Cite.($\mathcal{Q}_\mathrm{src}$) (\%)}}
      & \multicolumn{2}{c}{\textbf{Cite.($\mathcal{Q}_\mathrm{en}$) (\%)}}
      & \multicolumn{2}{c}{\textbf{Lang. (\%)}} \\
    \cmidrule(lr){3-5} \cmidrule(lr){6-7} \cmidrule(lr){8-9} \cmidrule(lr){10-11}
    & & $\mathrm{src}$ & $\mathrm{en}$ & $\Delta_\mathrm{F1}$ & $\mathrm{src}$ & $\mathrm{en}$ & $\mathrm{src}$ & $\mathrm{en}$ & $\mathrm{src}$ & $\mathrm{en}$ \\
    \midrule
    \rowcolor{gray!20}
    \multicolumn{11}{c}{\textbf{Knowledge Gap}} \\
    \midrule
      \multirow{3}{*}{\includegraphics[height=1em]{figures/logo/unlock.png}}
        & \textsc{LLaMA-3.1}       & 32.1 & 25.4 & 6.7  & 78.6 & 18.2 & 59.0 & 34.6 & 93.9 & 97.1 \\
      & \textsc{Qwen-3}            & 29.2 & 22.8 & 6.4  & 83.7 & 14.7 & 66.7 & 33.0 & 87.0 & 97.1 \\
      & \textsc{Aya-Expanse}       & 33.2 & 24.4 & 8.8  & 48.2 & 27.2 & 38.2 & 42.3 & 92.0 & 69.3 \\
    \cmidrule(lr){1-11}
      \multirow{5}{*}{\includegraphics[height=1em]{figures/logo/lock.png}}
        & \textsc{GPT-4o}          & 33.9 & 34.2 & -0.3 & 94.3 & 5.7  & 89.9 & 10.1 & 99.9  & 94.8 \\
      & \textsc{GPT-4.1}           & 37.6 & 35.4 & 2.2  & 98.6 & 1.4  & 94.2 & 5.8  & 80.8  & 98.3 \\
      & \textsc{DeepSeek-V4-Pro}   & 37.2 & 31.0 & 6.2  & 97.2 & 2.7  & 94.2 & 5.8  & 97.7  & 91.5 \\
      & \textsc{Grok-4.3}          & 39.5 & 30.5 & 9.0  & 97.3 & 1.9  & 94.5 & 5.2  & 100.0 & 97.8 \\
      & \textsc{GPT-5.5}           & 44.3 & 43.0 & 1.3  & 99.6 & 0.4  & 97.7 & 2.3  & 95.9  & 97.6 \\
    \midrule
    \rowcolor{gray!20}
    \multicolumn{11}{c}{\textbf{Knowledge Conflict}} \\
    \midrule
      \multirow{3}{*}{\includegraphics[height=1em]{figures/logo/unlock.png}}
        & \textsc{LLaMA-3.1}       & 32.7 & 14.1 & 18.6 & 65.6 & 30.4 & 14.4 & 81.3 & 93.5 & 98.3 \\
      & \textsc{Qwen-3}            & 29.1 & 14.3 & 14.8 & 64.5 & 33.4 & 24.9 & 72.9 & 88.7 & 97.9 \\
      & \textsc{Aya-Expanse}       & 33.4 & 14.1 & 19.3 & 46.3 & 32.2 & 25.2 & 60.7 & 86.1 & 78.5 \\
    \cmidrule(lr){1-11}
      \multirow{5}{*}{\includegraphics[height=1em]{figures/logo/lock.png}}
        & \textsc{GPT-4o}          & 34.7 & 16.7 & 18.0 & 62.3 & 37.7 & 17.1 & 82.9 & 99.4  & 96.6 \\
      & \textsc{GPT-4.1}           & 38.7 & 18.6 & 20.1 & 78.3 & 21.7 & 27.3 & 72.7 & 78.9  & 97.8 \\
      & \textsc{DeepSeek-V4-Pro}   & 39.4 & 18.9 & 20.5 & 77.4 & 22.5 & 28.0 & 72.0 & 94.1  & 95.2 \\
      & \textsc{Grok-4.3}          & 37.9 & 17.0 & 21.0 & 80.2 & 17.9 & 27.9 & 70.7 & 100.0 & 98.5 \\
      & \textsc{GPT-5.5}           & 44.6 & 20.1 & 24.5 & 89.9 & 10.1 & 19.7 & 80.3 & 95.8  & 98.4 \\
    \bottomrule
\end{tabular}
}
\caption{\textbf{F1 score gap, citation attribution (Cite.($\mathcal{Q}_\mathrm{src}$), ($\mathcal{Q}_\mathrm{en}$)), and response language for each model and setting.}}
\label{tab:metrics}
\end{table}

\subsection{Knowledge Parity}
\label{appendix:truth}

To verify that the observed query-language preference is significantly reduced when equivalent knowledge is available across languages, we introduce a control condition with cross-lingual knowledge parity, which corresponds to another cell in~\autoref{tab:quadrant}. 
In this condition, the source-language and English documents support the same target fact and answer: we inject an English translation of the target fact into the English document.
As shown in \autoref{tab:truth}, the query-language preference gaps measured by answer agreement and answer probability are generally close to zero. 


\section{Hypotheses}
\label{appendix:hypotheses}

To identify potential causes of query-language preference and rule out plausible confounders, we formulate and test four hypotheses. We introduce each hypothesis below.

\subsection{Pre-training Effect}

\begin{hypothesisbox}{}
For a fixed answer, models assign higher conditional probability when the query and document are in the same language than when they are in different languages.
\end{hypothesisbox}

\noindent\textbf{Rationale.}
Multilingual language models are pre-trained primarily on large, language-specific web corpora, whereas explicitly aligned cross-lingual text, such as parallel or bilingual documents, is comparatively limited \citep{burchell-etal-2025-expanded}. 
Consequently, semantically aligned same-language query--document pairs may be more common in pre-training data than equivalent cross-language pairs, potentially inducing a higher prior for same-language query--document combinations.

\noindent\textbf{Test.}
Using three open-weight models, we compare the perplexity of a fixed candidate answer when the query language either matches or differs from the language of the supporting document. For each supported answer \(\mathcal{A}_{\ell}\) and its document \(\mathcal{D}_{\ell}\), where \(\ell \in \{\mathrm{src},\mathrm{en}\}\), we compute:
\begin{equation}
\Delta_{\mathrm{PPL}}^{\ell}
=
\mathrm{PPL}\!\left(
\mathcal{A}_{\ell}
\mid
\mathcal{Q}_{\neg\ell}, \mathcal{D}_{\ell}
\right)
-
\mathrm{PPL}\!\left(
\mathcal{A}_{\ell}
\mid
\mathcal{Q}_{\ell}, \mathcal{D}_{\ell}
\right),
\end{equation}
where \(\neg\ell\) denotes the other query language. For \(\ell=\mathrm{en}\), \(\mathcal{D}_{\mathrm{en}}\) and \(\mathcal{A}_{\mathrm{en}}\) refer to the conflict-injected English document and its supported answer, respectively. A positive gap indicates that the fixed answer has higher perplexity, and therefore lower likelihood, when the query and document languages differ. A gap near zero indicates insensitivity to query--document language alignment.

\noindent\textbf{Result.}
As shown in~\autoref{tab:hypothesis1}, the perplexity gap is generally positive for conflict-supported English answers, whereas results are mixed for source-supported answers. Among the three models, \textsc{Aya-Expanse} 8B exhibits the largest effects and the most statistically significant positive or negative gaps. However, most gaps, including those above and below zero, are not statistically significant. Thus, this analysis provides limited evidence that query--document language alignment, as reflected in answer perplexity, is a substantial driver of query-language preference.

\begin{table}[H]
    \centering
    \resizebox{0.7\linewidth}{!}{%
    \begin{tabular}{llrrrl}
    \toprule
    \textbf{Language} & 
    \textbf{Model} & 
    \textbf{$\mathrm{PPL}(\cdot|\mathcal{Q}_\neg\ell,\mathcal{D}_\ell)$} &
    \textbf{$\mathrm{PPL}(\cdot|\mathcal{Q}_\ell,\mathcal{D}_\ell)$} &
    \textbf{\(\Delta_{\mathrm{PPL}}\)} & $p$ \\

    \midrule
    \rowcolor{gray!20}
\multicolumn{6}{c}{\(\mathcal{A}_{\mathrm{en}}'\)} \\
    \midrule

    \textbf{French} & \textsc{LLaMA-3.1}    & 4.22  & 3.42  & 0.81 & \\
     & \textsc{Qwen-3}       & 8.25  & 8.09  & 0.16 & \\
     & \textsc{Aya-Expanse}  & 21.76 & 10.04 & 11.72 & * \\
    \textbf{Chinese} & \textsc{LLaMA-3.1}    & 3.80  & 2.93  & 0.88 & \\
     & \textsc{Qwen-3}       & 8.78  & 5.43  & 3.36 & \\
     & \textsc{Aya-Expanse}  & 11.57 & 5.77  & 5.80 & \\
    \textbf{Japanese} & \textsc{LLaMA-3.1}    & 2.94  & 2.89  & 0.05 & \\
     & \textsc{Qwen-3}       & 8.84  & 6.16  & 2.68 & \\
     & \textsc{Aya-Expanse}  & 15.07 & 5.83  & 9.24 & \\
    \textbf{Korean} & \textsc{LLaMA-3.1}    & 2.58  & 2.46  & 0.12 & \\
     & \textsc{Qwen-3}       & 7.49  & 5.82  & 1.67 & \\
     & \textsc{Aya-Expanse}  & 11.21 & 5.90  & 5.31 & \\
    \textbf{Bengali} & \textsc{LLaMA-3.1}    & 3.71  & 2.62  & 1.09 & \\
     & \textsc{Qwen-3}       & 6.83  & 4.78  & 2.05 & \\
     & \textsc{Aya-Expanse}  & 12.80 & 5.24  & 7.52 & ** \\

    \midrule
    \rowcolor{gray!20}
    \multicolumn{6}{c}{\textbf{$\mathcal{A}_\mathrm{src}$}} \\
    \midrule

    \textbf{French} & \textsc{LLaMA-3.1}    & 3.92  & 2.97  & 0.95 & \\
     & \textsc{Qwen-3}       & 6.99  & 5.69  & 1.31 &  \\
     & \textsc{Aya-Expanse}  & 18.48 & 18.74 & $-$0.26 & \\
    \textbf{Chinese} & \textsc{LLaMA-3.1}    & 6.15  & 4.26  & 1.88 &  \\
     & \textsc{Qwen-3}       & 8.15  & 8.50  & $-$0.35 & \\
     & \textsc{Aya-Expanse}  & 22.41 & 23.75 & $-$1.34 & \\
    \textbf{Japanese} & \textsc{LLaMA-3.1}    & 4.40  & 2.88  & 1.51 &  \\
     & \textsc{Qwen-3}       & 5.71  & 5.18  & 0.53 &  \\
     & \textsc{Aya-Expanse}  & 10.76 & 11.73 & $-$0.96 & * \\
    \textbf{Korean} & \textsc{LLaMA-3.1}    & 4.95  & 3.81  & 1.14 &  \\
     & \textsc{Qwen-3}       & 6.83  & 6.19  & 0.65 & * \\
     & \textsc{Aya-Expanse}  & 13.24 & 14.24 & $-$0.99 & \\
    \textbf{Bengali} & \textsc{LLaMA-3.1}    & 1.40  & 1.30  & 0.10 & * \\
     & \textsc{Qwen-3}       & 1.79  & 1.68  & 0.11 &  \\
     & \textsc{Aya-Expanse}  & 1.95  & 1.85  & 0.10 & * \\

    \bottomrule
\end{tabular}
}
\caption{\textbf{Results for B.1 Pre-training Effect hypothesis.} Wilcoxon signed-rank test performed against zero; *: statistically significant with $p$ $<$ .05; **: $p$ $<$ .01; unmarked: not significant.}
\label{tab:hypothesis1}
\end{table}

\subsection{Embedding Similarity}

\begin{hypothesisbox}{}
Models are more likely to attribute an answer to the document matching the query language because same-language query--document pairs have higher embedding similarity than cross-language pairs.
\end{hypothesisbox}

\noindent\textbf{Rationale.}
Retrieval-augmented models may favor documents that are more similar to the query in embedding space. If multilingual representations systematically assign higher similarity to same-language than cross-language query--document pairs, this bias could lead models to preferentially cite the document written in the query language.

\noindent\textbf{Test.}
For each of the three open-weight models, we compute cosine similarity between each query and the source-language and conflict-injected English documents using the model's own embeddings. For a query in language \(\ell \in \{\mathrm{src}, \mathrm{en}\}\), let \(\mathcal{D}_{\ell}\) denote the document in the query language and \(\mathcal{D}_{\neg\ell}\) the document in the other language. We compute the same-language similarity advantage:
\begin{equation}
\Delta_{\mathrm{Sim}}^{\ell}
=
\cos\!\left(
\mathrm{embed}(\mathcal{Q}_{\ell}),
\mathrm{embed}(\mathcal{D}_{\ell})
\right)
-
\cos\!\left(
\mathrm{embed}(\mathcal{Q}_{\ell}),
\mathrm{embed}(\mathcal{D}_{\neg\ell})
\right).
\end{equation}
For English queries, \(\mathcal{D}_{\mathrm{en}}\) denotes the conflict-injected English document. A positive value indicates that the query is more similar to the same-language document, while a value near zero indicates little sensitivity to query--document language alignment.

We then compute the point-biserial correlation between \(\Delta_{\mathrm{Sim}}^{\ell}\) and citation attribution, a binary indicator of whether the model cites the document in the query language (Appendix~\ref{appendix:more_metrics}). A significant positive correlation indicates that models are more likely to cite the same-language document when it has a larger embedding-similarity advantage, supporting the hypothesis.

\noindent\textbf{Result.}
As shown in~\autoref{tab:hypothesis2}, correlations between \(\Delta_{\mathrm{Sim}}^{\ell}\) and citation attribution are mixed across models and languages, with only a small number of model--language combinations showing a statistically significant positive correlation. Thus, greater query--document embedding similarity does not reliably predict query-language preference from an attribution perspective.

\begin{table}[H]
\centering
\resizebox{0.55\linewidth}{!}{%
\begin{tabular}{l
    S[table-format=-1.3]
    S[table-format=-1.3]
    S[table-format=-1.3]}
\toprule
{\textbf{Language}} & {\textbf{\textsc{LLaMA-3.1}}} & {\textbf{\textsc{Qwen-3}}} & {\textbf{\textsc{Aya-Expanse}}} \\
\midrule
\rowcolor{gray!20}
\multicolumn{4}{c}{\textbf{$\mathcal{Q}_\mathrm{en}$}} \\
\midrule
\textbf{French}   & 0.137  & 0.189  & 0.171 \\
\textbf{Chinese}  & 0.055  & 0.008  & 0.048    \\
\textbf{Japanese} & 0.042  & 0.274**& 0.073    \\
\textbf{Korean}   & -0.028 & -0.115 & 0.138    \\
\textbf{Bengali}  & 0.108  & -0.080 & 0.132    \\
\midrule
\rowcolor{gray!20}
\multicolumn{4}{c}{\textbf{$\mathcal{Q}_\mathrm{src}$}} \\
\midrule
\textbf{French}   & 0.066  & 0.071  & 0.220*   \\
\textbf{Chinese}  & -0.030 & 0.183  & 0.237*   \\
\textbf{Japanese} & -0.014 & 0.026  & 0.022    \\
\textbf{Korean}   & 0.201* & 0.182  & 0.053    \\
\textbf{Bengali}  & 0.231* & -0.319 & 0.173    \\
\bottomrule
\end{tabular}%
}
\caption{\textbf{Results for B.2 Embedding Similarity hypothesis.} Point-biserial correlation is performed; *: statistically significant with $p$ $<$ .05; **: $p$ $<$ .01; unmarked: not significant.}
\label{tab:hypothesis2}
\end{table}

\subsection{Surface Cues}

\begin{hypothesisbox}{}
Query-language preference is driven by surface-level document properties, including relative document length and input order.
\end{hypothesisbox}

\noindent\textbf{Rationale.}
Language models can be sensitive to superficial properties of evidence that affect their predictions \citep{wan-etal-2024-evidence}. 
Thus, observed query-language preference may instead reflect differences in the source-language and English Wikipedia documents, rather than a preference for the query language itself. 
We examine two potential confounds: relative document length and document order in the input context. Although we randomize document order, residual position effects could still influence which document a model follows \citep{liu-etal-2024-lost}.

\noindent\textbf{Test.}
Using the three open-weight models, we measure document length in tokens using each model's tokenizer and compute the length difference:
\begin{equation}
\Delta_{\mathrm{length}}
=
\mathrm{Length}(\mathcal{D}_{\mathrm{src}})
-
\mathrm{Length}(\mathcal{D}_{\mathrm{en}}').
\end{equation}
Positive values indicate that the source-language document is longer. We also record whether the source-language document appears first in the input context. For each surface cue, we compute its Pearson correlation with the instance-level answer agreement difference between source-language and English queries. A significant correlation would indicate that the cue is associated with observed query-language preference.

\noindent\textbf{Result.}
\autoref{tab:hypothesis3} shows that correlations between relative document length and the answer agreement difference are generally negative across models and languages: as the source-language document becomes longer relative to the English document, answer agreement gap tends to decrease. However, most correlations are not statistically significant, providing little evidence that document length reliably drives query-language preference.

Correlations between document order and the answer agreement difference are mixed in direction, consistently near zero, and not statistically significant. Thus, document position in the input context also does not appear to confound query-language preference.

\begin{table}[H]
\centering
\resizebox{0.55\linewidth}{!}{%
\begin{tabular}{l
    S[table-format=-1.3]
    S[table-format=-1.3]
    S[table-format=-1.3]}
\toprule
{\textbf{Language}} & {\textbf{\textsc{LLaMA-3.1}}} & {\textbf{\textsc{Qwen-3}}} & {\textbf{\textsc{Aya-Expanse}}} \\
\midrule
\rowcolor{gray!20}
\multicolumn{4}{c}{\textbf{$\Delta_\mathrm{length}$}} \\
\midrule
\textbf{French}   & -0.117 & -0.107 & -0.118 \\
\textbf{Chinese}  & 0.046  & -0.227* & -0.255* \\
\textbf{Japanese} & -0.098 & -0.112 & -0.066 \\
\textbf{Korean}   & -0.105 & -0.142 & -0.124 \\
\textbf{Bengali}  & -0.082 & 0.109  & -0.032 \\
\midrule
\rowcolor{gray!20}
\multicolumn{4}{c}{\textbf{$\mathrm{Order}$}} \\
\midrule
\textbf{French}   & 0.082  & 0.033  & -0.029 \\
\textbf{Chinese}  & 0.093  & -0.075 & 0.058 \\
\textbf{Japanese} & -0.062 & 0.106  & -0.042 \\
\textbf{Korean}   & 0.031  & -0.088 & 0.021 \\
\textbf{Bengali}  & -0.131 & 0.039  & 0.081 \\
\bottomrule
\end{tabular}%
}
\caption{\textbf{Results for B.3 Surface Cues hypothesis.} Pearson correlation is performed; *: statistically significant with $p$ $<$ .05; unmarked: not significant.}
\label{tab:hypothesis3}
\end{table}

\subsection{Topic Locality}
\label{appendix:hypothesis4}

\begin{hypothesisbox}{}
Query-language preference may reflect a query's cultural specificity rather than the language in which it is asked.
\end{hypothesisbox}

\noindent\textbf{Rationale.}
Native-language Wikipedia editions often cover regionally relevant topics that are absent from their English counterparts, a pattern associated with self-focus bias \citep{hecht2009measuring, callahan2011cultural} and situated knowledge \citep{haraway1988situated}. 
Consequently, culturally specific queries may be better supported by documents in the associated language. Models might therefore follow the source-language document because of topic locality rather than query language itself. We test this possibility by comparing query-language preference for culturally specific and culturally agnostic queries.

\noindent\textbf{Test.}
Following \citet{park-etal-2026-enhancing}, we use \textsc{GPT-5.6-Terra} to classify each QA pair as culturally \textit{specific} or \textit{agnostic} using Prompt A.10 in Appendix~\ref{appendix:prompts}. Specific queries concern topics culturally associated with the source language or its related region; agnostic queries do not require language- or culture-specific knowledge.

We compare the instance-level answer agreement difference between source-language and English queries across these two groups using Welch's two-sample \(t\)-test. A statistically significant group difference would indicate that topic locality contributes to observed query-language preference.

\noindent\textbf{Result.}
As shown in~\autoref{tab:hypothesis4}, differences in the answer-agreement difference between culturally specific and agnostic queries are not statistically significant. Thus, this analysis provides little evidence that topic locality confounds query-language preference.

\begin{table}[H]
\centering
\resizebox{0.7\linewidth}{!}{%
\begin{tabular}{llrrr}
\toprule
{\textbf{Language}} & {\textbf{Model}} & {\textbf{$\Delta_\mathrm{agree}$(Specific)}} & {\textbf{$\Delta_\mathrm{agree}$(Agnostic)}} & {\textbf{Difference}} \\
\midrule
\textbf{French}   & \textsc{LLaMA-3.1}    & 58.3 & 45.3 & 0.130  \\
                  & \textsc{Qwen-3}       & 48.8 & 40.0 & 0.088  \\
                  & \textsc{Aya-Expanse}  & 16.3 & 21.8 & $-$0.055 \\
\midrule
\textbf{Chinese}  & \textsc{LLaMA-3.1}    & 90.0 & 55.0 & 0.150 \\
                  & \textsc{Qwen-3}       & 57.5 & 70.0 & $-$0.125 \\
                  & \textsc{Aya-Expanse}  & 33.8 & 25.0 & 0.088  \\
\midrule
\textbf{Japanese} & \textsc{LLaMA-3.1}    & 71.6 & 66.7 & 0.049  \\
                  & \textsc{Qwen-3}       & 72.2 & 52.4 & 0.198  \\
                  & \textsc{Aya-Expanse}  & 38.0 & 38.1 & $-$0.001 \\
\midrule
\textbf{Korean}   & \textsc{LLaMA-3.1}    & 50.7 & 50.0 & 0.007  \\
                  & \textsc{Qwen-3}       & 60.6 & 55.6 & 0.050  \\
                  & \textsc{Aya-Expanse}  & 29.6 & 55.6 & $-$0.260 \\
\midrule
\textbf{Bengali}  & \textsc{LLaMA-3.1}    & 61.1 & 75.0 & $-$0.139 \\
                  & \textsc{Qwen-3}       & 55.6 & 65.0 & $-$0.094 \\
                  & \textsc{Aya-Expanse}  & 44.4 & 51.3 & $-$0.068 \\
\bottomrule
\end{tabular}%
}
\caption{\textbf{Results for B.4 Topic Locality hypothesis.} We perform Welch's two-sample $t$-test, where unmarked: not statistically significant.}
\label{tab:hypothesis4}
\end{table}

\section{Mitigation Details}

\subsection{\textsc{QLoRA} Implementation Details}
\label{appendix:implementation}

We use 4-bit quantization with \texttt{bf16} compute. We set LoRA rank $r=$ 16, alpha $a =$ 32, dropout 0.05, and use AdamW optimizer \citep{loshchilov2017decoupled} with learning rate 2e-4 and weight decay 0. 
We use a linear scheduler with 3\% warmup and set the gradient clipping at max norm 1.0.
Training uses batch size 1 with gradient accumulation of 8, and runs for 3 epochs. 
Each model and objective is fine-tuned on a single \texttt{L40s} (48GB) GPU.
We set the clipping bound $c$ to 20 since the mean raw Forget-Retain score difference ranged from $-$15.2 to 56.4, with an average of 14.6, allowing us to leave the typical signal unchanged while limiting the effect of extreme likelihood gaps.

\subsection{Per-query-language Results}
\label{appendix:mitigation_res}

\autoref{tab:mitigation_full} reports the full mitigation results for each model, with answer agreement and probability gaps under source-language and English queries, and all $k$ values. We observe that \textsc{QLoRA}-based training reduces both gaps the most, while \textsc{QPHA} also reduces them, though to a smaller extent.

\begin{table}[htbp]
\centering
\resizebox{\linewidth}{!}{%
\begin{tabular}{l *{3}{S[table-format=2.1] S[table-format=2.1] S[table-format=2.1, table-space-text-post={***}] S[table-format=2.1] S[table-format=2.1] S[table-format=2.1, table-space-text-post={***}]}}
    \toprule
    \multirow{3}{*}{\textbf{Approach}}
      & \multicolumn{6}{c}{\textbf{\textsc{LLaMA-3.1}}}
      & \multicolumn{6}{c}{\textbf{\textsc{Qwen-3}}}
      & \multicolumn{6}{c}{\textbf{\textsc{Aya-Expanse}}} \\
    \cmidrule(lr){2-7} \cmidrule(lr){8-13} \cmidrule(lr){14-19}
      & \multicolumn{3}{c}{\textbf{Agreement}} & \multicolumn{3}{c}{\textbf{Probability}}
      & \multicolumn{3}{c}{\textbf{Agreement}} & \multicolumn{3}{c}{\textbf{Probability}}
      & \multicolumn{3}{c}{\textbf{Agreement}} & \multicolumn{3}{c}{\textbf{Probability}} \\
    \cmidrule(lr){2-4} \cmidrule(lr){5-7} \cmidrule(lr){8-10} \cmidrule(lr){11-13} \cmidrule(lr){14-16} \cmidrule(lr){17-19}
      & {$\mathrm{src}$} & {$\mathrm{en}$} & {$\Delta$} & {$\mathrm{src}$} & {$\mathrm{en}$} & {$\Delta$}
      & {$\mathrm{src}$} & {$\mathrm{en}$} & {$\Delta$} & {$\mathrm{src}$} & {$\mathrm{en}$} & {$\Delta$}
      & {$\mathrm{src}$} & {$\mathrm{en}$} & {$\Delta$} & {$\mathrm{src}$} & {$\mathrm{en}$} & {$\Delta$} \\
    \midrule
    Vanilla         & 78.7 & 15.7 & 63.0 & 47.6 & 17.7 & 29.9 & 76.9 & 17.9 & 59.0 & 17.8 & 5.7 & 12.1 & 72.4 & 13.6 & 58.8 & 58.0 & 24.5 & 33.5 \\
    \midrule
    \rowcolor{gray!20}
    \multicolumn{19}{c}{\textsc{\textbf{QPHA}} ($k=$ 2)} \\
    \midrule
    Random (Zero)   & 78.6 & 15.9 & 62.7 & 47.5 & 18.2 & 29.3 & 76.4 & 19.6 & 56.8 & 17.8 & 6.8 & 11.0 & 72.4 & 13.5 & 58.9 & 57.4 & 24.3 & 33.1 \\
    Random (Mean)   & 78.4 & 15.3 & 63.1 & 47.8 & 17.8 & 30.0 & 76.5 & 18.4 & 58.1 & 18.1 & 6.7 & 11.4 & 72.7 & 13.3 & 59.4 & 57.1 & 24.0 & 33.1 \\
    Zero            & 76.4 & 15.0 & 61.4 & 47.7 & 21.0 & 26.7* & 76.9 & 20.7 & 56.2 & 18.4 & 7.4 & 11.0 & 72.2 & 16.1 & 56.1* & 57.3 & 27.2 & 30.1* \\
    Mean            & 78.5 & 18.8 & 59.7 & 46.6 & 21.9 & 24.7** & 76.8 & 20.4 & 56.4 & 18.2 & 7.4 & 10.8 & 72.9 & 14.4 & 58.5 & 57.8 & 25.1 & 32.7 \\
    \midrule
    \rowcolor{gray!20}
    \multicolumn{19}{c}{\textsc{\textbf{QPHA}} ($k=$ 4)} \\
    \midrule
    Random (Zero)   & 77.5 & 14.6 & 62.9 & 48.7 & 21.8 & 26.9* & 76.0 & 19.3 & 56.7 & 18.2 & 6.4 & 11.8 & 73.5 & 13.7 & 59.8 & 55.7 & 23.9 & 31.8 \\
    Random (Mean)   & 78.8 & 15.6 & 63.2 & 50.0 & 21.0 & 29.0 & 76.8 & 18.5 & 58.3 & 18.3 & 8.2 & 10.1* & 72.7 & 14.2 & 58.5 & 57.1 & 23.8 & 33.3 \\
    Zero            & 75.5 & 15.3 & 60.2 & 46.8 & 24.1 & 22.7** & 76.7 & 21.5 & 55.2* & 18.4 & 7.5 & 10.9 & 72.7 & 16.2 & 56.5 & 57.4 & 26.6 & 30.8* \\
    Mean            & 77.1 & 19.9 & 57.2** & 48.8 & 33.0 & 15.8*** & 76.3 & 20.1 & 56.2 & 17.4 & 6.8 & 10.6 & 72.9 & 16.8 & 56.1* & 58.4 & 26.0 & 32.4 \\
    \midrule
    \rowcolor{gray!20}
    \multicolumn{19}{c}{\textsc{\textbf{QPHA}} ($k=$ 8)} \\
    \midrule
    Random (Zero)   & 77.8 & 16.6 & 61.2 & 49.0 & 22.7 & 26.3* & 75.4 & 18.4 & 57.0 & 18.3 & 6.2 & 12.1 & 72.8 & 14.2 & 58.6 & 56.1 & 25.3 & 30.8* \\
    Random (Mean)   & 77.2 & 17.8 & 59.4* & 50.0 & 21.8 & 28.2 & 75.5 & 18.0 & 57.5 & 18.2 & 6.4 & 11.8 & 72.6 & 15.1 & 57.5 & 56.7 & 24.9 & 31.8 \\
    Zero            & 75.1 & 18.8 & 56.3** & 46.4 & 27.0 & 19.4*** & 76.0 & 21.7 & 54.3* & 18.1 & 7.6 & 10.5 & 72.1 & 16.4 & 55.7* & 57.2 & 27.5 & 29.7* \\
    Mean            & 78.3 & 22.4 & 55.9** & 47.3 & 38.2 & 9.1*** & 76.7 & 21.3 & 55.4* & 17.7 & 7.6 & 10.1* & 72.8 & 17.8 & 55.0* & 57.1 & 28.4 & 28.7** \\
    \midrule
    \rowcolor{gray!20}
    \multicolumn{19}{c}{\textbf{\textsc{QLoRA}}} \\
    \midrule
    Promote         & 89.3 & 87.0 & 2.3*** & 83.5 & 79.2 & 4.3*** & 86.4 & 82.1 & 4.3*** & 77.8 & 70.1 & 7.7** & 78.5 & 78.2 & 0.3*** & 74.4 & 69.5 & 4.9*** \\
    Contrastive     & 85.5 & 84.0 & 1.5*** & 82.1 & 75.7 & 6.4*** & 82.1 & 80.7 & 1.4*** & 76.3 & 71.0 & 5.3*** & 77.4 & 76.5 & 0.9*** & 72.3 & 70.6 & 1.7*** \\
    \bottomrule
\end{tabular}%
}
\caption{\textbf{Effect of intervention approaches on agreement ($\Delta_\mathrm{agree}$) and probability ($\Delta_\mathrm{prob}$) gaps.} 
For both metrics, lower is better (↓), which indicate reduced query-language preference. 
We perform Wilcoxon signed-rank test against Vanilla, with Bonferroni correction. 
*: significant with $p$ $<$ .05; **: $p$ $<$ .01; ***: $p$ $<$ .001; unmarked: not significant.}
\label{tab:mitigation_full}
\end{table}

\subsection{Swapped Roles Mitigation}
\label{appendix:mitigation_swapped}

To test whether our mitigation approaches generalize, we re-run the mitigation experiments (\S\ref{sec:mitigation}) under the swapped-roles setup (\autoref{tab:swapped}), where the conflicting fact is in the source-language document and the original target fact in the English document. 

As shown in~\autoref{tab:mitigation_swapped}, both \textsc{QPHA} and \textsc{QLoRA} remain effective: all models show reduced agreement and probability gaps, with \textsc{QLoRA} again yielding the largest reduction. This confirms that our mitigations generalize to the reversed direction, not only the original \textsc{Waldo} configuration.

\begin{table}[htbp]
\centering
\resizebox{\linewidth}{!}{%
\begin{tabular}{l *{3}{S[table-format=2.1] S[table-format=2.1] S[table-format=-2.1, table-space-text-post={***}] S[table-format=2.1] S[table-format=2.1] S[table-format=-2.1, table-space-text-post={***}]}}
    \toprule
    \multirow{3}{*}{\textbf{Approach}}
      & \multicolumn{6}{c}{\textbf{\textsc{LLaMA-3.1}}}
      & \multicolumn{6}{c}{\textbf{\textsc{Qwen-3}}}
      & \multicolumn{6}{c}{\textbf{\textsc{Aya-Expanse}}} \\
    \cmidrule(lr){2-7} \cmidrule(lr){8-13} \cmidrule(lr){14-19}
      & \multicolumn{3}{c}{\textbf{Agreement}} & \multicolumn{3}{c}{\textbf{Probability}}
      & \multicolumn{3}{c}{\textbf{Agreement}} & \multicolumn{3}{c}{\textbf{Probability}}
      & \multicolumn{3}{c}{\textbf{Agreement}} & \multicolumn{3}{c}{\textbf{Probability}} \\
    \cmidrule(lr){2-4} \cmidrule(lr){5-7} \cmidrule(lr){8-10} \cmidrule(lr){11-13} \cmidrule(lr){14-16} \cmidrule(lr){17-19}
      & {$\mathrm{src}$} & {$\mathrm{en}$} & {$\Delta$} & {$\mathrm{src}$} & {$\mathrm{en}$} & {$\Delta$}
      & {$\mathrm{src}$} & {$\mathrm{en}$} & {$\Delta$} & {$\mathrm{src}$} & {$\mathrm{en}$} & {$\Delta$}
      & {$\mathrm{src}$} & {$\mathrm{en}$} & {$\Delta$} & {$\mathrm{src}$} & {$\mathrm{en}$} & {$\Delta$} \\
    \midrule
    Vanilla         & 22.5 & 86.3 & -63.8 & 30.4 & 18.5 & 11.9 & 23.6 & 77.9 & -54.3 & 11.9 & 5.6 & 6.3 & 27.8 & 70.3 & -42.5 & 20.9 & 17.2 & 3.7 \\
    \midrule
    \rowcolor{gray!20}
    \multicolumn{19}{c}{\textsc{\textbf{QPHA}} ($k=$ 8)} \\
    \midrule
    Zero            & 24.1 & 82.1 & -58.0** & 28.8 & 21.1 & 7.7*  & 27.4 & 77.0 & -49.6* & 11.9 & 7.2 & 4.7* & 30.6 & 65.0 & -34.4*** & 20.8 & 19.5 & 1.3* \\
    Mean            & 27.7 & 85.4 & -57.7** & 30.1 & 26.5 & 3.6**  & 27.0 & 76.1 & -49.1* & 12.6 & 7.3 & 5.3 & 32.0 & 69.4 & -37.4** & 20.8 & 18.6 & 2.2* \\
    Rand-Zero       & 21.9 & 85.4 & -63.5 & 31.0 & 20.5 & 10.5 & 24.1 & 76.4 & -52.3 & 13.1 & 5.7 & 7.4 & 28.4 & 70.7 & -42.3 & 20.7 & 17.3 & 3.4 \\
    Mean-Zero       & 23.1 & 84.8 & -61.7 & 31.4 & 20.2 & 11.2 & 23.7 & 76.5 & -52.8 & 14.5 & 7.2 & 7.3 & 29.3 & 70.5 & -41.2 & 20.8 & 17.2 & 3.6 \\
    \midrule
    \rowcolor{gray!20}
    \multicolumn{19}{c}{\textbf{\textsc{QLoRA}}} \\
    \midrule
    Promote         & 92.3 & 96.9 & -4.6*** & 37.2 & 37.5 & -0.3*** & 87.8 & 87.4 & 0.4*** & 34.0 & 34.0 & 0.0*** & 70.0 & 70.3 & -0.3*** & 31.4 & 31.2 & 0.2** \\
    Contrastive     & 89.3 & 93.1 & -3.8*** & 38.2 & 35.1 & 3.1***  & 86.4 & 83.1 & 3.3*** & 29.9 & 30.1 & -0.2*** & 69.5 & 70.4 & -0.9*** & 24.9 & 26.9 & -2.0** \\
    \bottomrule
\end{tabular}%
}
\caption{\textbf{Effect of intervention approaches on agreement ($\Delta_\mathrm{agree}$) and probability ($\Delta_\mathrm{prob}$) gaps for swapped roles.} 
For both metrics, $\Delta$ values closer to zero indicate reduced query-language preference. 
We perform Wilcoxon signed-rank test against Vanilla, with Bonferroni correction. 
*: significant with $p$ $<$ .05; **: $p$ $<$ .01; ***: $p$ $<$ .001; unmarked: not significant.}
\label{tab:mitigation_swapped}
\end{table}

\subsection{Increasing the Number of Ablated Heads}
\label{appendix:other_models}

\begin{figure*}
    \centering
    \includegraphics[width=0.92\linewidth]{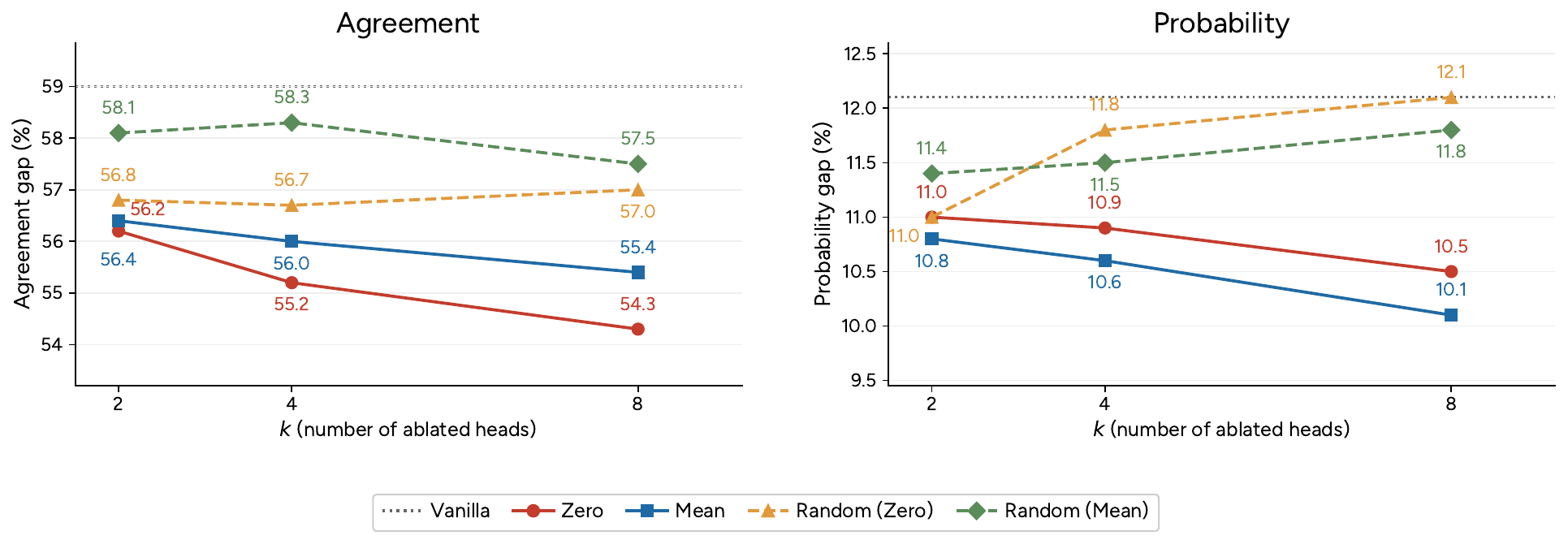}
    \includegraphics[width=0.92\linewidth]{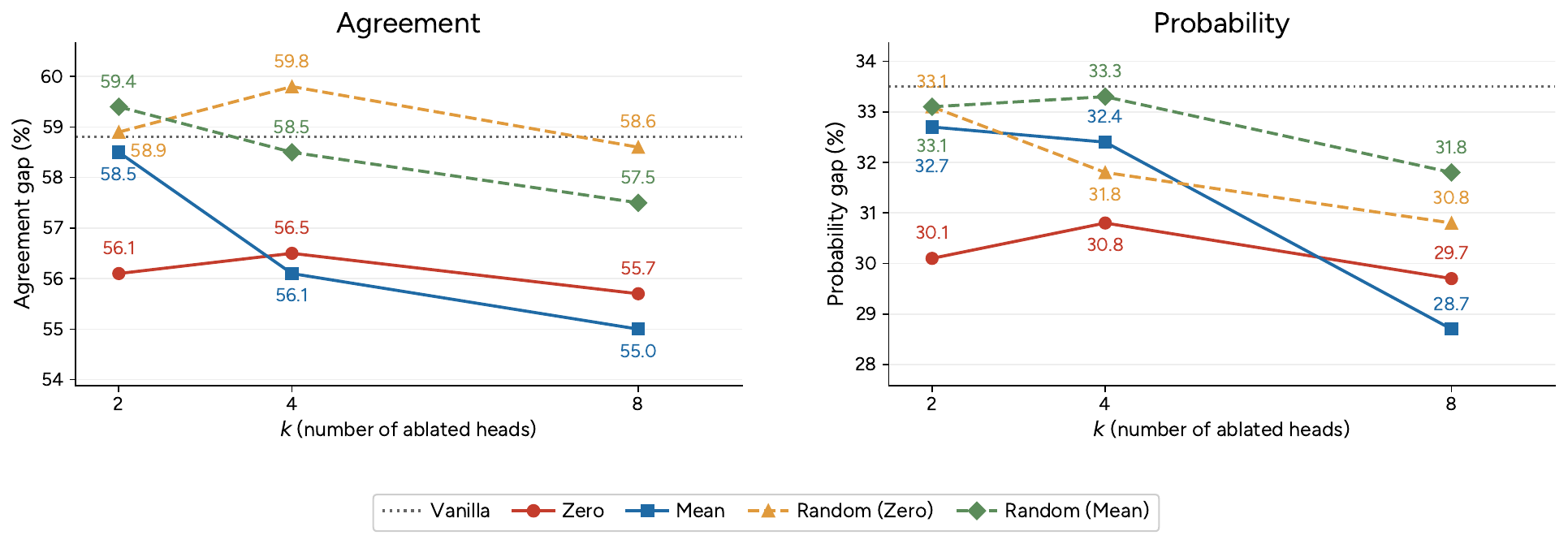}
    \vspace{-1em}
    \caption{\textbf{Answer agreement and probability gap with increasing $k$.} Both metrics decrease as the number of ablated attention heads ($k$) increases, with \textsc{QPHA} showing larger gap than random control. Top: \textsc{Qwen-3}; Bottom: \textsc{Aya-Expanse}.}
    \label{fig:increasing_k_other}
\end{figure*}

We show the effect of increasing the number of ablated attention heads in~\autoref{fig:increasing_k_other}. The trend matches~\autoref{fig:increasing_k}: as $k$ grows, both the agreement and probability gap shrink, with \textsc{QPHA} producing a larger reduction than the random ablation control relative to the vanilla baseline.

\subsection{Ablated Attention Heads}
\label{appendix:ablated_heads}

\begin{figure*}
    \centering
    \includegraphics[width=0.33\linewidth]{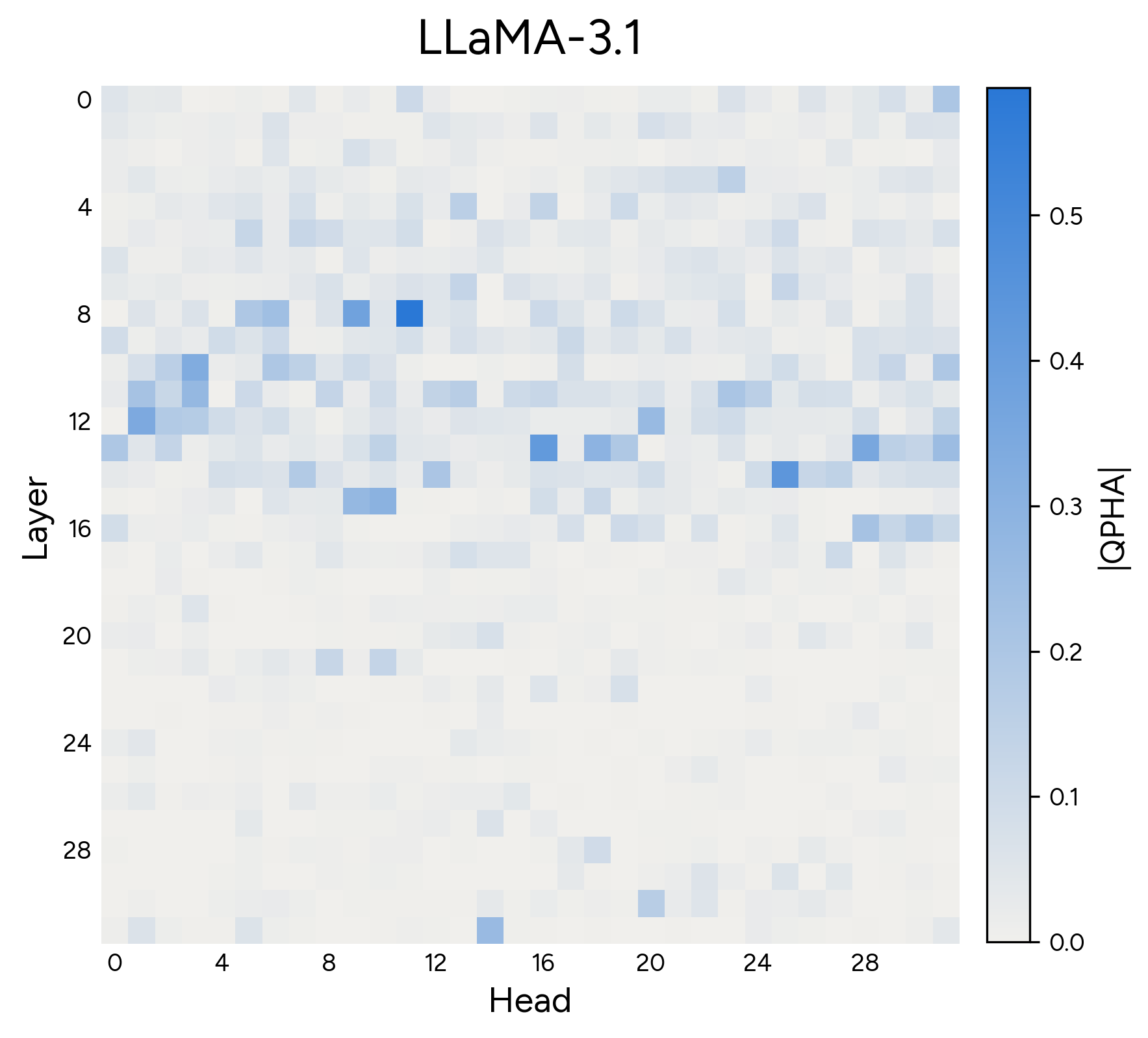}
    \includegraphics[width=0.3\linewidth]{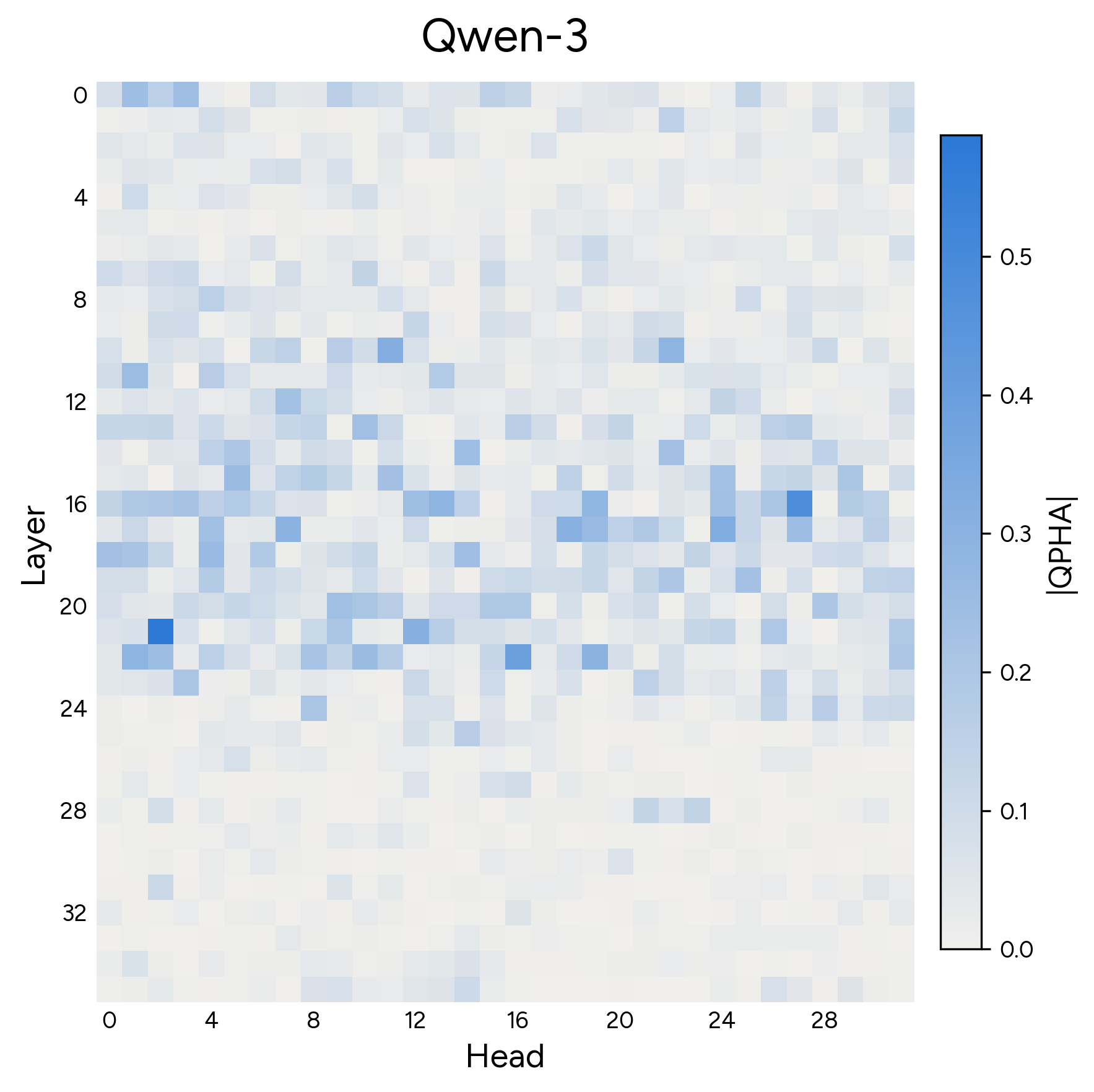}
    \includegraphics[width=0.33\linewidth]{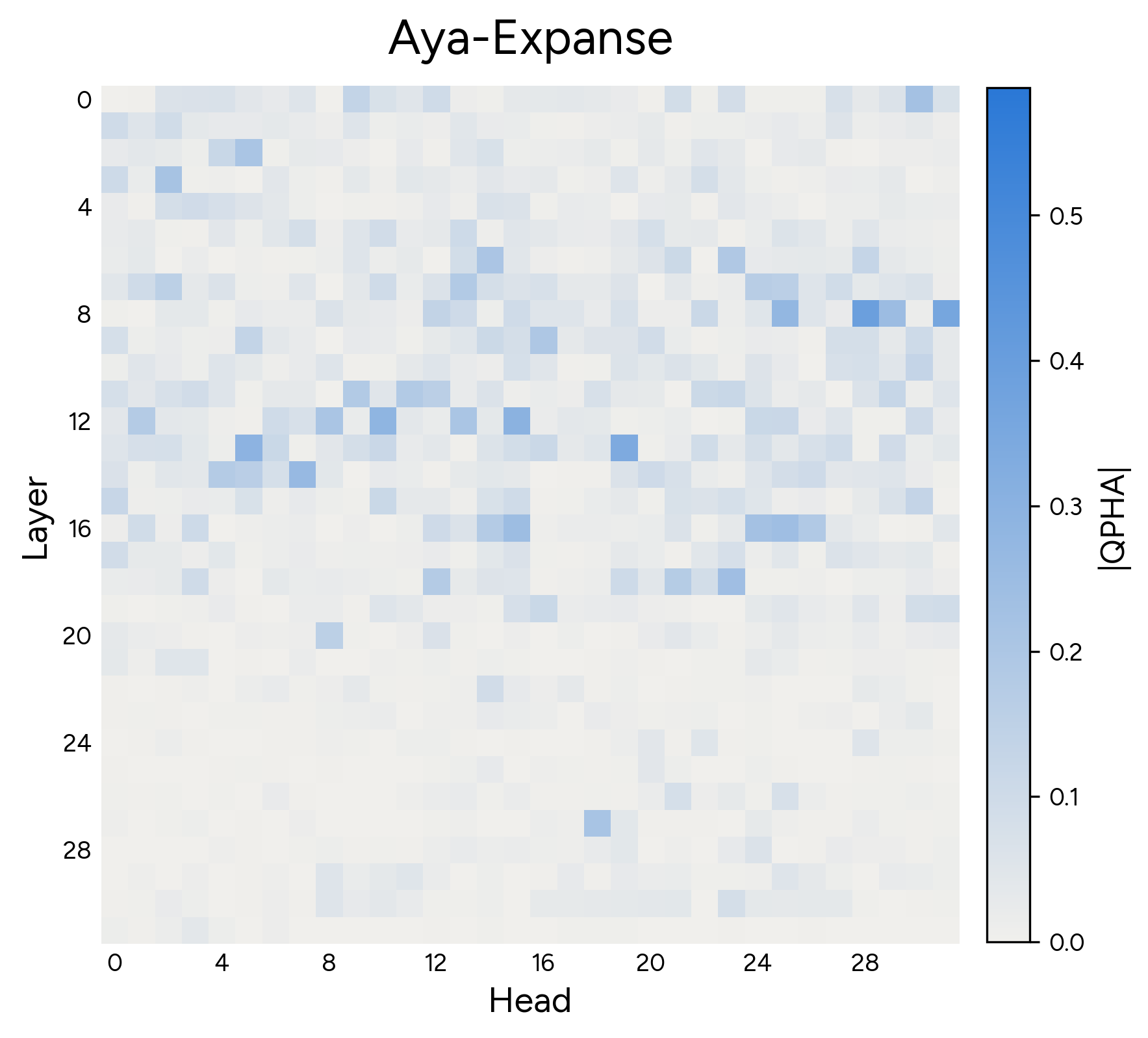}
    \caption{\textbf{Full attention-head score map for each model.} Darker-colored cells indicate higher \textsc{QPHA} scores (i.e., heads more strongly associated with query-language preference).}
    \label{fig:attention}
\end{figure*}

\autoref{tab:ablated_heads} lists the attention heads selected for ablation for each model and value of \(k\). \autoref{fig:attention} visualizes the full attention-head score map for each model, where darker colors indicate higher \textsc{QPHA} scores. The highest-scoring heads are concentrated in early-to-middle layers: layers 12--15 for \textsc{LLaMA-3.1}, 17--22 for \textsc{Qwen-3}, and 8--12 for \textsc{Aya-Expanse}.

\begin{table}[htbp]
\centering
\resizebox{0.8\linewidth}{!}{%
\begin{tabular}{llp{0.65\linewidth}}
    \toprule
    \textbf{Model} & \textbf{$k$} & \textbf{Ablated pairs (\textit{layer}, \textit{head})} \\
    \midrule
    \multirow{3}{*}{\textbf{\textsc{LLaMA-3.1}}}
      & 2 & (12,1), (15,9) \\
      & 4 & (12,1), (15,9), (31,14), (13,31) \\
      & 8 & (12,1), (15,9), (31,14), (13,31), (8,6), (16,28), (11,1), (13,19) \\
    \midrule
    \multirow{3}{*}{\textbf{\textsc{Qwen-3}}}
      & 2 & (17,18), (22,1) \\
      & 4 & (17,18), (22,1), (16,19), (20,10) \\
      & 8 & (17,18), (22,1), (16,19), (20,10), (22,31), (17,30), (10,7), (17,20) \\
    \midrule
    \multirow{3}{*}{\textbf{\textsc{Aya-Expanse}}}
      & 2 & (8,28), (12,13) \\
      & 4 & (8,28), (12,13), (11,29), (12,6) \\
      & 8 & (8,28), (12,13), (11,29), (12,6), (10,27), (29,8), (30,8), (29,11) \\
    \bottomrule
\end{tabular}
}
\caption{\textbf{Ablated attention heads per model and $k$.} \textsc{LLaMA-3.1} and \textsc{Aya-Expanse} have a total of 32 layers, and \textsc{Qwen-3} has 36 layers.}  
\label{tab:ablated_heads}
\end{table}

\begin{table}[htbp]
\centering
\resizebox{0.8\linewidth}{!}{%
\begin{tabular}{llrrrrrr}
    \toprule
       \multirow{2}{*}{\textbf{Language}} & \multirow{2}{*}{\textbf{Model}}
      & \multicolumn{3}{c}{\textbf{Agreement (\%)}}
      & \multicolumn{3}{c}{\textbf{Probability (\%)}} \\
    \cmidrule(lr){3-5} \cmidrule(lr){6-8}
    & & $\mathrm{src}$ & $\mathrm{en}$ & $\Delta_\mathrm{agree}$ & $\mathrm{src}$ & $\mathrm{en}$ & $\Delta_\mathrm{prob}$ \\
    \midrule
    \multirow{8}{*}{\textbf{French}}
      & \textsc{LLaMA-3.1}       & 90.4 & 84.3 & 6.0 & 27.9 & 17.6 & 10.3 \\
      & \textsc{Qwen-3}          & 80.7 & 74.7 & 6.0 & 10.9 & 5.1 & 5.8 \\
      & \textsc{Aya-Expanse}     & 67.5 & 60.2 & 7.2 & 42.7 & 33.5 & 9.2 \\
      & \textsc{GPT-4o}          & 95.5 & 93.6 & 1.8 & \na & \na & \na \\
      & \textsc{GPT-4.1}         & 100.0 & 94.0 & 6.0 & \na & \na & \na \\
      & \textsc{DeepSeek-V4-Pro} & 98.9 & 94.5 & 4.4 & \na & \na & \na \\
      & \textsc{Grok-4.3}        & 99.0 & 97.5 & 1.4 & \na & \na & \na \\
      & \textsc{GPT-5.5}         & 100.0 & 96.2 & 3.8 & \na & \na & \na \\
    \midrule
    \multirow{8}{*}{\textbf{Chinese}}
      & \textsc{LLaMA-3.1}       & 96.4 & 67.3 & 29.1 & 31.5 & 16.3 & 15.2 \\
      & \textsc{Qwen-3}          & 92.7 & 80.0 & 12.7 & 9.5 & 5.2 & 4.3 \\
      & \textsc{Aya-Expanse}     & 69.1 & 45.5 & 23.6 & 43.9 & 31.2 & 12.7 \\
      & \textsc{GPT-4o}          & 99.2 & 88.2 & 11.0 & \na & \na & \na \\
      & \textsc{GPT-4.1}         & 98.0 & 95.0 & 3.0 & \na & \na & \na \\
      & \textsc{DeepSeek-V4-Pro} & 98.4 & 97.3 & 1.1 & \na & \na & \na \\
      & \textsc{Grok-4.3}        & 99.4 & 99.4 & 0.0 & \na & \na & \na \\
      & \textsc{GPT-5.5}         & 99.3 & 99.3 & 0.0 & \na & \na & \na \\
    \midrule
    \multirow{8}{*}{\textbf{Japanese}}
      & \textsc{LLaMA-3.1}       & 93.8 & 71.9 & 21.9 & 51.1 & 31.6 & 19.4 \\
      & \textsc{Qwen-3}          & 79.2 & 55.7 & 23.5 & 13.4 & 7.1 & 6.3 \\
      & \textsc{Aya-Expanse}     & 75.0 & 56.2 & 18.8 & 51.0 & 34.5 & 16.6 \\
      & \textsc{GPT-4o}          & 96.5 & 86.8 & 9.6 & \na & \na & \na \\
      & \textsc{GPT-4.1}         & 98.0 & 93.0 & 5.0 & \na & \na & \na \\
      & \textsc{DeepSeek-V4-Pro} & 98.6 & 96.1 & 2.4 & \na & \na & \na \\
      & \textsc{Grok-4.3}        & 98.8 & 94.7 & 4.1 & \na & \na & \na \\
      & \textsc{GPT-5.5}         & 98.7 & 97.4 & 1.3 & \na & \na & \na \\
    \midrule
    \multirow{8}{*}{\textbf{Korean}}
      & \textsc{LLaMA-3.1}       & 98.1 & 88.7 & 9.4 & 40.3 & 14.9 & 25.4 \\
      & \textsc{Qwen-3}          & 92.5 & 84.9 & 7.4 & 18.8 & 6.2 & 12.6 \\
      & \textsc{Aya-Expanse}     & 86.8 & 66.0 & 20.8 & 45.9 & 38.4 & 7.5 \\
      & \textsc{GPT-4o}          & 100.0 & 96.4 & 3.6 & \na & \na & \na \\
      & \textsc{GPT-4.1}         & 100.0 & 97.6 & 2.4 & \na & \na & \na \\
      & \textsc{DeepSeek-V4-Pro} & 100.0 & 98.2 & 1.8 & \na & \na & \na \\
      & \textsc{Grok-4.3}        & 97.9 & 98.1 & $-$0.2 & \na & \na & \na \\
      & \textsc{GPT-5.5}         & 100.0 & 100.0 & 0.0 & \na & \na & \na \\
    \midrule
    \multirow{8}{*}{\textbf{Bengali}}
      & \textsc{LLaMA-3.1}       & 91.4 & 78.5 & 12.9 & 70.7 & 48.6 & 22.1 \\
      & \textsc{Qwen-3}          & 87.1 & 76.3 & 10.8 & 39.2 & 25.1 & 14.1 \\
      & \textsc{Aya-Expanse}     & 52.7 & 32.3 & 20.4 & 69.2 & 50.5 & 18.7 \\
      & \textsc{GPT-4o}          & 98.0 & 96.7 & 1.3  & \na & \na & \na \\
      & \textsc{GPT-4.1}         & 97.9 & 96.9 & 1.0  & \na & \na & \na \\
      & \textsc{DeepSeek-V4-Pro} & 94.3 & 93.8 & 0.5  & \na & \na & \na \\
      & \textsc{Grok-4.3}        & 97.5 & 96.3 & 1.2  & \na & \na & \na \\
      & \textsc{GPT-5.5}         & 95.6 & 99.3 & $-$3.7 & \na & \na & \na \\
    \bottomrule
\end{tabular}
}
\caption{\textbf{Per-language answer agreement ($\Delta_\mathrm{agree}$) and probabilities ($\Delta_\mathrm{prob}$) under \textcolor{first}{knowledge gap} setting.} \textbf{Agreement} ($\mathrm{src}$, $\mathrm{en}$): answer agreement; \textbf{Probability} $\mathrm{src}$, $\mathrm{en}$): probability assigned to the source-supported answer, under source-language and English queries, respectively. Closed-source models do not expose answer probabilities (--).
A larger positive $\Delta$ value indicates stronger query-language preference.}
\label{tab:perlang_gap}
\end{table}

\begin{table}[htbp]
\centering
\resizebox{0.8\linewidth}{!}{%
\begin{tabular}{llrrrrrr}
    \toprule
       \multirow{2}{*}{\textbf{Lang.}} & \multirow{2}{*}{\textbf{Model}}
      & \multicolumn{3}{c}{\textbf{Agreement (\%)}}
      & \multicolumn{3}{c}{\textbf{Probability (\%)}} \\
    \cmidrule(lr){3-5} \cmidrule(lr){6-8}
    & & $\mathrm{src}$ & $\mathrm{en}$ & $\Delta_\mathrm{agree}$ & $\mathrm{src}$ & $\mathrm{en}$ & $\Delta_\mathrm{prob}$ \\
    \midrule
    \multirow{8}{*}{\textbf{French}}
      & \textsc{LLaMA-3.1}       & 78.3 & 19.3 & 59.0 & 25.9 & 10.8 & 15.1 \\
      & \textsc{Qwen-3}          & 53.0 & 19.3 & 33.7 & 9.0 & 3.3 & 5.7 \\
      & \textsc{Aya-Expanse}     & 54.2 & 28.9 & 25.3 & 37.5 & 20.4 & 17.1 \\
      & \textsc{GPT-4o}          & 51.1 & 25.0 & 26.1 & \na & \na & \na \\
      & \textsc{GPT-4.1}         & 72.5 & 20.9 & 51.5 & \na & \na & \na \\
      & \textsc{DeepSeek-V4-Pro} & 53.1 & 16.9 & 36.2 & \na & \na & \na \\
      & \textsc{Grok-4.3}        & 79.7 & 24.3 & 55.4 & \na & \na & \na \\
      & \textsc{GPT-5.5}         & 81.5 & 20.4 & 61.1 & \na & \na & \na \\
    \midrule
    \multirow{8}{*}{\textbf{Chinese}}
      & \textsc{LLaMA-3.1}       & 85.5 & 7.3 & 78.2 & 29.7 & 7.6 & 22.0 \\
      & \textsc{Qwen-3}          & 78.2 & 18.2 & 60.0 & 7.9 & 4.0 & 3.9 \\
      & \textsc{Aya-Expanse}     & 63.6 & 7.2 & 56.4 & 44.1 & 16.7 & 27.4 \\
      & \textsc{GPT-4o}          & 81.6 & 10.6 & 70.9 & \na & \na & \na \\
      & \textsc{GPT-4.1}         & 90.9 & 21.6 & 69.2 & \na & \na & \na \\
      & \textsc{DeepSeek-V4-Pro} & 91.8 & 29.9 & 61.9 & \na & \na & \na \\
      & \textsc{Grok-4.3}        & 83.0 & 26.0 & 57.0 & \na & \na & \na \\
      & \textsc{GPT-5.5}         & 87.5 & 23.9 & 63.6 & \na & \na & \na \\
    \midrule
    \multirow{8}{*}{\textbf{Japanese}}
      & \textsc{LLaMA-3.1}       & 84.4 & 10.9 & 73.4 & 46.5 & 10.1 & 36.4 \\
      & \textsc{Qwen-3}          & 71.9 & 14.1 & 57.8 & 10.1 & 4.0 & 6.1 \\
      & \textsc{Aya-Expanse}     & 59.4 & 7.8 & 51.6 & 47.5 & 12.6 & 34.9 \\
      & \textsc{GPT-4o}          & 75.4 & 12.7 & 62.7 & \na & \na & \na \\
      & \textsc{GPT-4.1}         & 79.7 & 35.6 & 44.1 & \na & \na & \na \\
      & \textsc{DeepSeek-V4-Pro} & 74.6 & 23.2 & 51.4 & \na & \na & \na \\
      & \textsc{Grok-4.3}        & 84.8 & 35.2 & 49.5 & \na & \na & \na \\
      & \textsc{GPT-5.5}         & 97.0 & 25.0 & 72.0 & \na & \na & \na \\
    \midrule
    \multirow{8}{*}{\textbf{Korean}}
      & \textsc{LLaMA-3.1}       & 83.0 & 15.1 & 67.9 & 35.6 & 7.7 & 27.9 \\
      & \textsc{Qwen-3}          & 77.4 & 20.8 & 56.6 & 16.8 & 4.3 & 12.5 \\
      & \textsc{Aya-Expanse}     & 86.8 & 5.7 & 81.1 & 44.6 & 14.6 & 30.0 \\
      & \textsc{GPT-4o}          & 60.9 & 17.8 & 43.1 & \na & \na & \na \\
      & \textsc{GPT-4.1}         & 80.3 & 36.0 & 44.4 & \na & \na & \na \\
      & \textsc{DeepSeek-V4-Pro} & 71.3 & 39.1 & 32.2 & \na & \na & \na \\
      & \textsc{Grok-4.3}        & 73.0 & 32.0 & 41.0 & \na & \na & \na \\
      & \textsc{GPT-5.5}         & 93.0 & 17.0 & 76.0 & \na & \na & \na \\
    \midrule
    \multirow{8}{*}{\textbf{Bengali}}
      & \textsc{LLaMA-3.1}       & 80.6 & 14.0 & 66.7 & 69.2 & 40.6 & 28.6 \\
      & \textsc{Qwen-3}          & 63.4 & 7.5 & 55.9 & 37.4 & 14.0 & 23.4 \\
      & \textsc{Aya-Expanse}     & 52.7 & 5.4 & 47.3 & 70.8 & 38.9 & 31.9 \\
      & \textsc{GPT-4o}          & 78.6 & 28.0 & 50.6 & \na & \na & \na \\
      & \textsc{GPT-4.1}         & 76.5 & 28.5 & 48.0 & \na & \na & \na \\
      & \textsc{DeepSeek-V4-Pro} & 87.3 & 27.6 & 59.7 & \na & \na & \na \\
      & \textsc{Grok-4.3}        & 88.0 & 29.1 & 59.0 & \na & \na & \na \\
      & \textsc{GPT-5.5}         & 88.5 & 19.8 & 68.8 & \na & \na & \na \\
    \bottomrule
\end{tabular}
}
\caption{\textbf{Per-language answer agreement ($\Delta_\mathrm{agree}$) and probabilities ($\Delta_\mathrm{prob}$) under \textcolor{second}{knowledge conflict} setting.} \textbf{Agreement} ($\mathrm{src}$, $\mathrm{en}$): answer agreement; \textbf{Probability} $\mathrm{src}$, $\mathrm{en}$): probability assigned to the source-supported answer, under source-language and English queries, respectively. Closed-source models do not expose answer probabilities (--).
A larger positive $\Delta$ value indicates stronger query-language preference.}
\label{tab:perlang_conflict}
\end{table}

\section{Prompts}
\label{appendix:prompts}

\begin{figure*}
    \centering
    \includegraphics[width=0.95\linewidth]{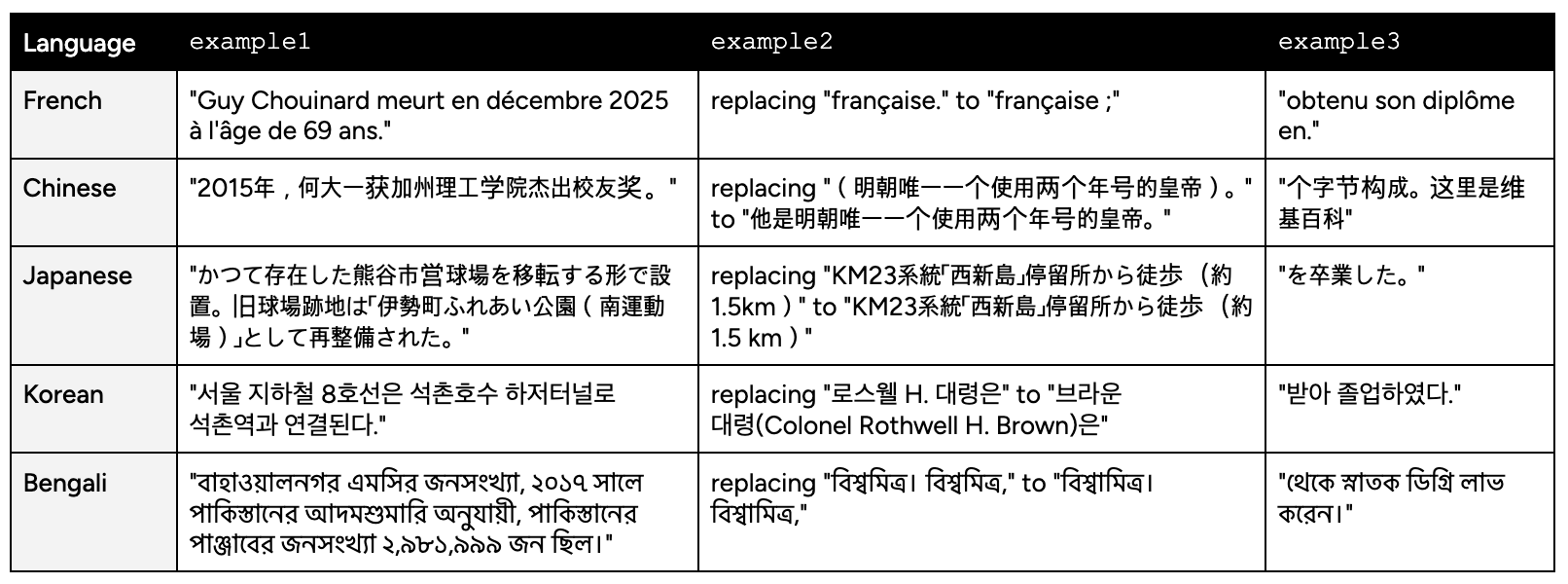}
    \caption{\textbf{Per-language examples propagated into Prompt A.1.} Native speakers manually collected these examples from sourced Wikipedia revisions.} 
    \label{fig:prompt_example}
\end{figure*}

We present the prompts used in the data construction pipeline for \textsc{Waldo} (\autoref{fig:main_fig}).
For the knowledge gap setting, we provide the prompts for LLM Annotation (A.1--A.2), and QA Generation (A.3).
The in-context examples in A.1 differ by the source language, as illustrated in~\autoref{fig:prompt_example}.
For the knowledge conflict setting, we provide the prompts for Conflict Injection \& QA Generation (A.4) and Answerability Check (A.5).
We also include the prompts used to qualitatively analyze the injected conflict types in~\autoref{tab:conflict_type} (A.6), the prompt used to query models, adapted from GPTResearcher\footnote{\url{https://github.com/assafelovic/gpt-researcher}} (A.7), and the prompt used to compare model responses with document-supported answers when computing answer agreement (\S\ref{sec:metrics}) (A.8).

We also provide the prompt used to swap roles of  documents (\autoref{tab:swapped}; A.9) and to classify queries as culturally specific or agnostic for the Topic Locality hypothesis (Appendix~\ref{appendix:hypothesis4}; A.10).

\begin{prompt}[title={Prompt A.1. LLM Annotation 1}]
\textbf{Instruction:} You are an annotation assistant for Wikipedia revision analysis. Your task is to evaluate a single revision to a French-language Wikipedia article. \\
You will be given: \\
- Article title: The title of the Wikipedia article \\
- Context: The surrounding Wikipedia article text \\
- Added information: The text newly introduced by the revision \\
- Removed information: The text deleted by the revision \\ \\
Your task is to evaluate the revision on two dimensions. \\
(1) ``\textbf{is fact}'': Is the added information a meaningful, self-contained factual addition? \\
Mark ``true'' if the added text:\\
- Introduces at least one new, standalone fact not present in the removed text (e.g., \{\texttt{example1}\}) \\
- AND the removed text does not delete meaningful information \\ \\
Mark ``false'' if the added text: \\
- Is only a stylistic, formatting, or phrasing change to existing content (e.g., \{\texttt{example2}\}) \\
- OR is an incomplete sentence fragment that cannot stand alone as a fact (e.g., \{\texttt{example3}\}) \\
- If the addition contains multiple facts, or one fact plus a minor fragment, mark ``true.'' \\ \\ 
(2) ``\textbf{date}'': When did the event described in the added information occur? \\
- Provide your best estimate in YYYY-MM format, use not only the provided context but also your general knowledge \\
- If only the year is determinable, use YYYY-01 as a convention \\
- If the date is genuinely indeterminate, use None \\ \\
Return only this JSON object with no other text: \\
\{\{ \\
  ``is fact'': true or false, \\
  ``date'': YYYY-MM or None, \\
  ``rationale is fact'': One-line explanation of your ``is fact'' decision, \\
  ``rationale date'': One-line explanation of your date \\
\}\} \\ \\
\textbf{Article title:} \{\texttt{title}\} \\
\textbf{Context:} \{\texttt{context}\} \\
\textbf{Added information:} \{\texttt{added}\} \\
\textbf{Removed information:} \{\texttt{removed}\}
\end{prompt}

\begin{prompt}[title={Prompt A.2. LLM Annotation 2}]
\textbf{Instruction:} You are an expert fact-checker. Your task is to determine whether a piece of information from a Wikipedia article is present in the English Wikipedia article on the same topic. \\
You will be given: \\
- Source information: A snippet of text from the source-language Wikipedia article \\
- Source context: The surrounding context of the source information in the article \\
- English Wikipedia title: The title of the corresponding English article \\
- English Wikipedia content: The content of the English article \\ \\
Decision rules: \\
- Match on meaning, not exact wording. Translations and paraphrases count as matches. \\
- Return true only if the key claim in the source information is clearly supported by the English content. \\
- If the claim is only partially present or vague, return false. \\
- If a number, date, name, or other specific detail in the source information conflicts with the English content, return false. \\ \\
Return only this JSON object with no other text: \\
\{\{ \\
  ``is present'': true or false, \\
  ``match'': The supporting sentence(s) quoted from the English Wikipedia content, or an empty string if is present is false \\
  ``rationale'': One-line explanation of your ``is present'' decision, \\
\}\} \\ \\
\textbf{Source information:} \{\texttt{added}\} \\
\textbf{Source context:} \{\texttt{context}\} \\ \\
\textbf{English Wikipedia title:} \{\texttt{English title}\} \\
\textbf{English Wikipedia content:} \\
\{\texttt{English article}\}
\end{prompt}
\begin{prompt}[title={Prompt A.3. QA Generation}]
\textbf{Instruction:} You are a query generator that targets information found in \texttt{\{lang\}}-language Wikipedia. Your task is to generate realistic user queries that target information present in a \texttt{\{lang\}} Wikipedia article but absent from (or only partially or vaguely covered by) the corresponding English article. For each query, you also produce the correct answer. \\ \\
You will be given: \\
- \texttt{\{lang\}} article title: The title of the \texttt{\{lang\}} Wikipedia article \\
- \texttt{\{lang\}} context: The surrounding context of the target information within the article \\
- English article title: The title of the corresponding English Wikipedia article, use this to understand how the topic is rendered in English. \\
- \texttt{\{lang\}} information: The text containing the target information the query should ask about \\
- Rationale: An explanation of why this information is missing or incomplete in the English article, use this only as a guide for choosing which part of the \texttt{\{lang\}} information to target when the English article already covers part of it \\ \\
Requirements for each query: \\
1. Produce the same query in both natural English and natural \texttt{\{lang\}}. The i-th English query and the i-th \texttt{\{lang\}} query must express the same question. \\
2. The query must be answerable using only the \texttt{\{lang\}} information provided. \\
3. Keep each query self-contained and focused on one specific fact. \\
4. Phrase it the way a real person would when typing into a search engine or asking an assistant, using conversational speech. \\
5. Frame it so it requires a factual answer, not a yes/no response. \\
6. Do not mention Wikipedia, the article, or the existence of \texttt{\{lang\}} or English versions. \\
7. Do not copy long phrases verbatim from the \texttt{\{lang\}} information; rephrase it in the user's own words. \\
8. Ensure ``en query,'' ``src query,'' ``en answer,'' and ``src answer'' are all lists of strings of equal length. \\ \\
Requirements for each answer: \\
9. Provide the answer to each query in both natural English (``en answer'') and natural \texttt{\{lang\}} (``src answer''). The i-th answer must correspond to the i-th query. \\
10. Each answer must be grounded solely in the \texttt{\{lang\}} information provided; do not add facts that are not present in it. \\
11. Give only the specific fact that answers the query, such as the value, name, date, or phrase itself. Do not restate or echo the question, and do not wrap the fact in a full sentence. \\ \\
Return only this JSON object with no other text: \\
\{\{ \\
  ``en query'': [List of English-language user queries targeting the \texttt{\{lang\}} information], \\
  ``src query'': [List of \texttt{\{lang\}}-language user queries targeting the \texttt{\{lang\}} information], \\
  ``en answer'': [List of English-language answers, aligned by index with en query], \\
  ``src answer'': [List of \texttt{\{lang\}}-language answers, aligned by index with src query] \\
\}\} \\ \\
\textbf{\texttt{\{lang\}} article title:} \texttt{\{title\}} \\
\textbf{\texttt{\{lang\}} context:} \texttt{\{context\}} \\
\textbf{English article title:} \texttt{\{English title\}} \\ \\
\textbf{\texttt{\{lang\}} information:} \texttt{\{added\}} \\
\textbf{Rationale:} \texttt{\{rationale\}}
\end{prompt}

\begin{prompt}[title={Prompt A.4. Conflict Injection+QA Generation}]
\textbf{Instruction:} Your task is to insert an altered fact into an English Wikipedia article, based on a fact stated in a source text. \\
You are given: \\
- The original English Wikipedia article \\
- A source text stating a fact \\
- A question about the article and its current (correct) answer\\ \\
Produce: \\
1. ``new sent'': exactly one English sentence that states a changed version of the key fact from the source text (e.g., a different name, number, date, or place than the correct one). \\
2. ``new article'': the original article with ``new sent'' inserted verbatim, replacing the original statement of that fact so the article still reads naturally.  \\
3. ``new answer'': the answer to the question if it were answered using ``new article'' instead of the original article. \\ \\
Rules: \\
- ``new sent'' must change the answer to the given question; i.e., if a reader relied on ``new sent'' instead of the original fact, they would answer the question differently than the correct answer provided. \\
- ``new sent'' must alter the most important fact in the source text (name, quantity, date, location, etc.). If several facts are entangled (e.g., a date tied to a name), alter all of them consistently.  \\
- Do not create the change by simple negation or polarity flip; change the underlying fact.  \\
- ``new article'' must contain ``new sent'' verbatim.  \\
- If the original fact appears more than once in the article, update every occurrence so the article remains internally consistent; do not leave the old and new facts contradicting each other.  \\
- Leave all unrelated content unchanged. Do not shorten or skip any unrelated content.  \\
- Maintain a neutral, encyclopedic tone; the insertion must not be stylistically detectable as an edit.  \\
- ``new answer'' must differ from the correct answer given below; it must be the specific fact, value, name, or date itself, not a full sentence, and must not restate the question.  \\
- Return only a valid JSON object with exactly these three keys: ``new sent'', ``new article'', and ``new answer'' \\ \\
\textbf{English Wikipedia title:} \{\texttt{English title}\}  \\
\textbf{English Wikipedia article:} \{\texttt{English article}\} \\
\textbf{Question:} \{\texttt{question}\} \\
\textbf{Correct answer:} \{\texttt{answer}\} \\
\textbf{Source information:} \{\texttt{added}\} \\
\textbf{Response:}
\end{prompt}

\begin{prompt}[title={Prompt A.5. Answerability Check}]
\textbf{Instruction:} You are given a reference article and a question. Answer the question using only the information in the article. \\ \\
\textbf{Article:} \{\texttt{context}\} \\
\textbf{Question:} \{\texttt{question}\} \\
\textbf{Answer:} \\ \\ \\
\textbf{Instruction:} Does the Model Answer convey the same fact as the Target Answer for the given question? Reply with exactly one word: ``match'' or ``no match.'' \\ \\
\textbf{Question:} \{\texttt{question}\} \\
\textbf{Target Answer:} \{\texttt{conflict answer}\} \\
\textbf{Model Response:} \{\texttt{model response}\} \\
\textbf{Verdict:}
\end{prompt}

\begin{prompt}[title={Prompt A.6. Conflict Type Analysis}]
\textbf{Instruction:} You are given a question, its original correct answer, and a conflicting answer that was substituted in its place (an altered version of the same fact). Classify what kind of fact was changed, using exactly one of the following categories: \\
- date: dates, years, months, or spans/durations of time (e.g. 2021 $\rightarrow$ 2019, about 30 years $\rightarrow$ about 25 years) \\
- number: quantities, counts, statistics, or measurements that are not dates (e.g., vote counts, ages, scores) \\
- person: names of real people or fictional characters  \\
- location: places, geographic regions, cities, or countries \\
- organization: companies, media outlets, institutions, teams, or other groups \\
- misc: anything that doesn't fit the above categories (e.g. roles/occupations, topics/themes, object or species names, reasons)  \\ \\
Respond with exactly one word from: date, number, person, location, organization, misc. \\
\textbf{Question:} \{\texttt{question}\} \\
\textbf{Original answer:} \{\texttt{original answer}\} \\
\textbf{Conflicting answer:} \{\texttt{conflict answer}\} \\
\textbf{Altered sentence:} \{\texttt{conflict sentence}\} \\
\textbf{Category:}
\end{prompt}

\begin{prompt}[title={Prompt A.7. Querying Models}]
\textbf{Information} \\
\textbf{Document ID:} 1 \\
\textbf{Title:} \{\texttt{title1}\} \\
\textbf{Article:} \{\texttt{article1}\} \\ \\
\textbf{Document ID:} 2 \\
\textbf{Title:} \{\texttt{title2}\} \\
\textbf{Article:} \{\texttt{article2}\} \\ \\
\textbf{Query:} \{\texttt{query}\} \\ \\
Using the above information, the response is the answer to the query in a single sentence. You must cite the most relevant document by including only its Document ID in brackets at the end of the sentence (e.g., [Document ID]). \\ \\
\textbf{Response:}
\end{prompt}

\begin{prompt}[title={Prompt A.8. Answer Agreement Judge}]
\textbf{Instruction:} Does the Model Answer convey the same fact as the Gold Answer for the given question? Ignore language, phrasing, formatting, and any trailing citation markers like [1] or [2]. Reply with exactly one word: ``match'' or ``no match.'' \\ \\
\textbf{Question:} \{\texttt{question}\} \\
\textbf{Gold Answer:} \{\texttt{gold answer}\} \\
\textbf{Model Response:} \{\texttt{model response}\} \\
\textbf{Verdict:}
\end{prompt}

\begin{prompt}[title={Prompt A.9. Swapping Roles of Documents}]
\textbf{Instruction:} Your task is to swap the factual information between two Wikipedia passages while keeping each replacement in the language of its destination article. \\ \\
Produce exactly two strings: \\
1. ``swapped text'': express the complete meaning of conflict fact in the language of the source article. It will replace the source fact exactly. \\
2. ``swapped conflict sent'': express the complete meaning of source fact in the language of conflict article. It will replace conflict fact exactly.\\ \\
Rules: \\
- Preserve the facts exactly when translating: names, dates, numbers, events, relations, and polarity must not change. \\
- Match the destination's language, grammar, capitalization, punctuation, and encyclopedic style. \\
- A target may be a sentence fragment. Use its surrounding context to make the replacement fit the same grammatical position; do not force it into a standalone sentence. \\
- Do not combine the old and new facts, explain the edit, or add citations. \\
- Return only one JSON object with exactly the keys ``swapped text'' and ``swapped conflict sent''.\\ \\
\textbf{Source fact:} \{\texttt{source fact}\} \\
\textbf{Source article :} \{\texttt{source article}\} \\
\textbf{Conflict fact:} \{\texttt{conflict fact}\} \\
\textbf{Conflict article :} \{\texttt{conflict article}\}
\end{prompt}
\begin{prompt}[title={Prompt A.10. Topic Locality Classification}]
\textbf{Instruction:} You are given a list of English questions (English query) with their English answers (English answer), and the corresponding original \texttt{\{lang\}} questions (Source-language query) and answers (Source-language answer) they were translated from. The i-th items across all four lists express the same question/answer pair. \\ \\
Your task is to classify each question/answer pair as exactly one of: \\
- ``\textbf{specific}'': the fact is tied to \texttt{\{lang\}} language, culture, geography, history, institutions, or audience; the kind of fact that is unlikely to be covered by a source that isn't specifically about that context.
- ``\textbf{agnostic}'': the fact is general knowledge that isn't inherently tied to that \texttt{\{lang\}} context, even though it happens to appear in this \texttt{\{lang\}} Wikipedia article (e.g., a fact about a person, place, or event outside that context, or a universal/technical fact). \\ \\
Requirements: \\
1. Produce exactly one label per question/answer pair, aligned by index. \\
2. The label must be exactly ``specific'' or ``agnostic'' (lowercase, no other text). \\
3. Ensure the output list has the same length as English query. \\ \\
Return only this JSON object with no other text: \\
\{\{ \\
  ``classification'': [List of ``specific'' or ``agnostic'' labels, aligned by index with the English query] \\
\}\} \\ \\
\textbf{English query:} \texttt{\{en query\}} \\
\textbf{English answer:} \texttt{\{en answer\}} \\
\textbf{Source-language query:} \texttt{\{src query\}} \\
\textbf{Source-language answer:} \texttt{\{src answer\}}
\end{prompt}


\end{document}